\documentclass{article}

\PassOptionsToPackage{numbers, compress}{natbib}
\usepackage[preprint]{neurips_2024}

\usepackage[utf8]{inputenc} 
\usepackage[T1]{fontenc}    
\usepackage{url}            
\usepackage{booktabs}       
\usepackage{amsfonts}       
\usepackage{nicefrac}       
\usepackage{microtype}      
\usepackage{xcolor}         
\usepackage{graphicx}       
\usepackage[font=normal]{subfig} 
\usepackage{comment} 
\usepackage{longtable} 
\usepackage{tabularx} 
\usepackage{amsmath} 
\usepackage{wrapfig}
\usepackage{placeins} 
\usepackage{listings} 
\usepackage{courier} 
\usepackage{multirow}

\definecolor{olive}{HTML}{B5D7A8}
\definecolor{gold}{HTML}{FFE599}
\definecolor{mikan}{HTML}{F9CB9C}

\definecolor{deepblue}{rgb}{0,0,0.5}
\definecolor{deepred}{rgb}{0.6,0,0}
\definecolor{deepgreen}{rgb}{0,0.5,0}

\makeatletter
\AtBeginDocument{%
\renewcommand{\sectionautorefname}{\S\@gobble}

\renewcommand{\subsectionautorefname}{\S\@gobble}  
}
\makeatother

\newcommand\rurl[1]{%
\href{https://#1}{\nolinkurl{#1}}%
}
\definecolor{mydarkblue}{rgb}{0,0.08,0.45}
\usepackage[colorlinks,citecolor=mydarkblue,urlcolor=mydarkblue,linkcolor=mydarkblue]{hyperref}
\usepackage{cleveref}

\usepackage{minitoc} 

\noptcrule

\newcommand{\model}{\textsc{GritLM}}
\newcommand{\modelbase}{\textsc{GritLM~7B}}
\newcommand{\modelbig}{\textsc{GritLM~8x7B}}
\newcommand{\method}{\textsc{GRIT}}
\newcommand{\tulu}{T\"ulu}
\newcommand{\llama}{Llama}

\DeclareSymbolFont{extraup}{U}{zavm}{m}{n}
\DeclareMathSymbol{\varheart}{\mathalpha}{extraup}{86}
\DeclareMathSymbol{\vardiamond}{\mathalpha}{extraup}{87}

\title{Generative Representational Instruction Tuning}

\author{
{\bf
Niklas Muennighoff
$^{\hspace{.1em}{\color{purple}\boldsymbol{c}}}$
\quad
Hongjin Su
$^{\hspace{.1em}{\color{purple}\boldsymbol{h}}}$
\quad
Liang Wang
$^{\hspace{.1em}{\color{purple}\boldsymbol{m}}}$
\quad
Nan Yang
$^{\hspace{.1em}{\color{purple}\boldsymbol{m}}}$
\vspace{4pt}
} \\
{\bf
Furu Wei
$^{{\color{purple}\boldsymbol{m}}}$
\enskip
Tao Yu
$^{{\color{purple}\boldsymbol{h}}}$
\enskip
Amanpreet Singh
$^{{\color{purple}\boldsymbol{c}}}$
\enskip
Douwe Kiela
$^{{\color{purple}\boldsymbol{c}}}$
\vspace{4pt}
} \\
{
$^{\color{purple}\boldsymbol{c}}$ Contextual AI \quad
$^{\color{purple}\boldsymbol{h}}$ The University of Hong Kong \quad
$^{\color{purple}\boldsymbol{m}}$ Microsoft Corporation 
\vspace{4pt}
}\\
\texttt{niklas@contextual.ai}
}

\begin{document}

\doparttoc
\faketableofcontents

\maketitle

\begin{abstract}

All text-based language problems can be reduced to either generation or embedding. Current models only perform well at one or the other. We introduce generative representational instruction tuning (\method{}) whereby a large language model is trained to handle both generative and embedding tasks by distinguishing between them through instructions. Compared to other open models, our resulting \modelbase{} sets a new state of the art on the Massive Text Embedding Benchmark~(MTEB) and outperforms all models up to its size on a range of generative tasks. By scaling up further, \modelbig{} outperforms all open generative language models that we tried while still being among the best embedding models. Notably, we find that \method{} matches training on only generative or embedding data, thus we can unify both at no performance loss. Among other benefits, the unification via \method{} speeds up Retrieval-Augmented Generation (RAG) by >~60\% for long documents, by no longer requiring separate retrieval and generation models. Models, code, etc. are freely available at \url{https://github.com/ContextualAI/gritlm}.


\begin{figure*}[htbp]
\centering
\begin{center}
\includegraphics[width=0.99\textwidth]{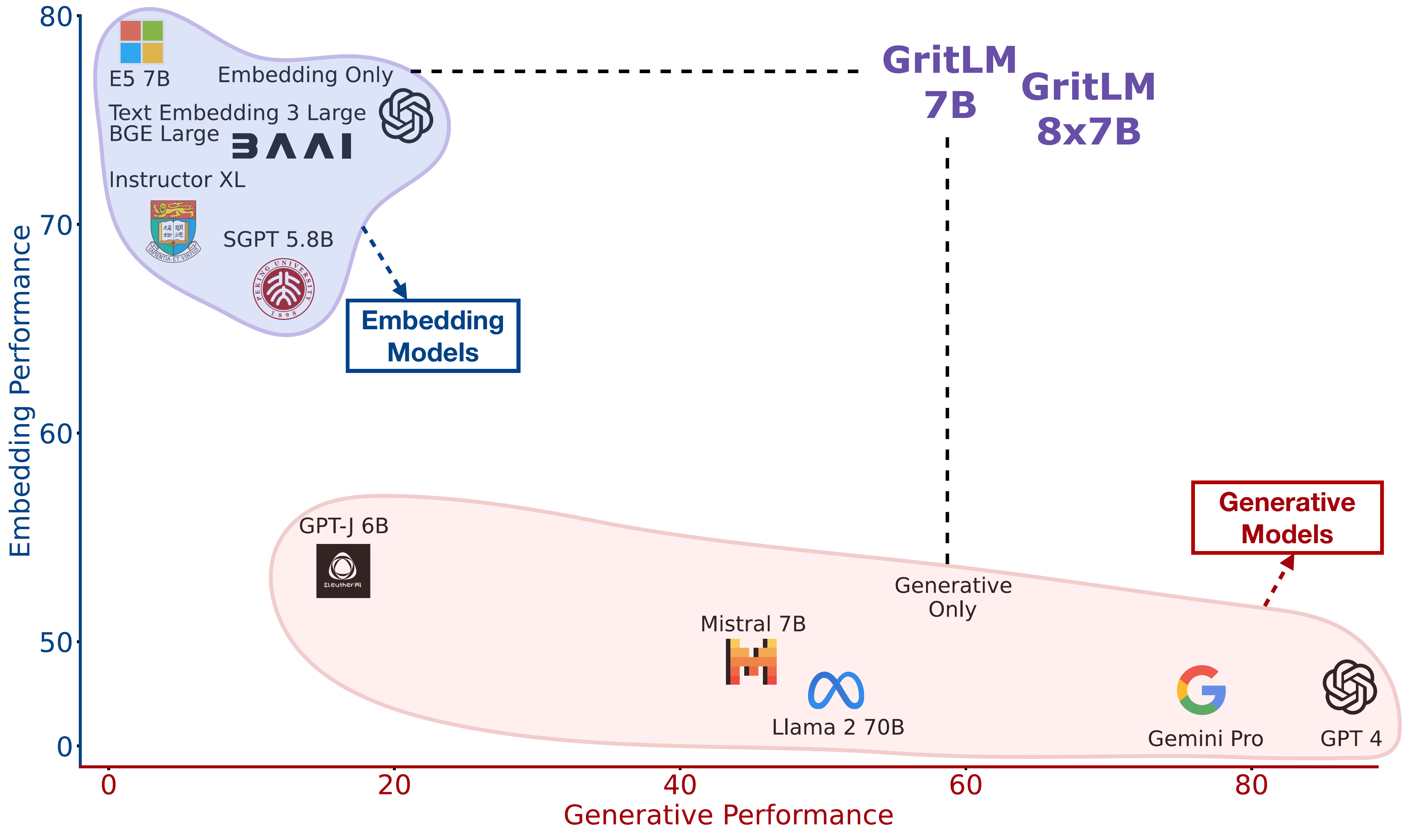}
\caption{\textbf{Performance of various models on text representation (embedding) and generation tasks.} \model{} is the first model to perform best-in-class at both types of tasks simultaneously.}
\label{fig:performance}
\end{center}
\end{figure*}

\begin{figure*}[htbp]
\centering
\begin{center}
\includegraphics[width=\textwidth]{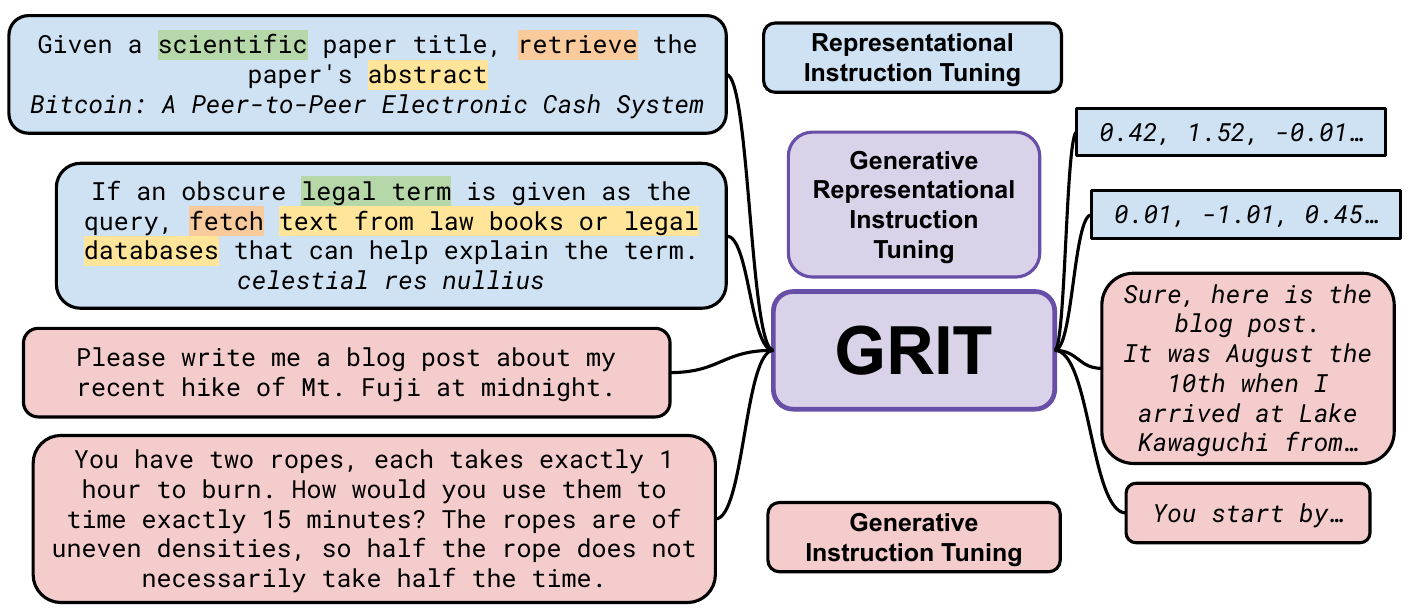}
\caption{\textbf{\method{}.} The same model handles both text representation and generation tasks based on the given instruction. For representation tasks, instructions ideally contain the target \colorbox{olive}{domain}, \colorbox{mikan}{intent}, and \colorbox{gold}{unit}~\cite{asai2022taskaware}. The representation is a tensor of numbers, while the generative output is text.} 
\label{fig:octopus}
\end{center}
\end{figure*}

\end{abstract}

\section{Introduction}
\label{sec:intro}

Creating a single general model that performs well at a wide range of tasks has been a long-standing goal of the field of artificial intelligence~\cite{kaiser2017model,jaegle2021perceiver,cho2021unifying,reed2022generalist,singh2022flava}. Recently, large language models (LLMs) have emerged as a promising direction for a single multi-task model~\cite{radford2019language,brown2020language}. Prior work has argued that all text-based language problems can be reduced to generation and thus handled by a single LLM~\cite{raffel2023exploring,du2021all}.

However, tasks that use embeddings, such as clustering or retrieval~\cite{muennighoff2023mteb}, have largely been ignored from this perspective. Today, text embeddings power many critical real-world applications ranging from search engines to user-facing chatbots~\cite{huang2020embedding,su2017chatbot}. While integrating text embeddings into the generative paradigm is possible by generating a sequence of numbers to form the embedding tensor, it becomes impractical due to the high dimensionality and precision requirements of embeddings. Thus, it is more common and much easier to use the hidden state of the model as the embedding representation, which is already a numeric tensor~\cite{muennighoff2022sgpt,wang2020sbertwk,morris2023text}. However, doing so for current generative models leads to poor performance. For example, while the T5 model~\cite{raffel2023exploring,sanh2022multitask} can handle any generative task in a sequence-to-sequence fashion, it requires finetuning to make its hidden state useful for text embedding~\cite{ni2021large,ni2021sentencet5} during which it loses its generative capabilities.


We introduce \method{} (generative representational instruction tuning) which unifies embedding and generative tasks, leading to a model that excels at both tasks as shown in \autoref{fig:performance}. \autoref{fig:octopus} depicts how \method{} combines two previously disjoint training paradigms: (1) \textit{Generative instruction tuning}, whereby the model is trained to respond to instructions by generating an answer~\cite{wei2022finetuned,sanh2022multitask}; and (2) \textit{Representational instruction tuning}, whereby the model is trained to represent a provided input according to an instruction~\cite{su2023embedder,asai2022taskaware}. Via the instructions and separate loss functions the model learns to differentiate the two streams. We test our approach on models with up to 47B parameters and, due to its simplicity, we expect the method to generalize to any LLM, even non-transformers. This unification via \method{} leads to three advantages:\\
\textbf{a) Performance:} Our unified model matches the performance of embedding-only and generative-only variants, even outperforming them on some tasks. At 7B parameters, \model{} sets a new state of the art on the Massive Text Embedding Benchmark~\cite{muennighoff2023mteb} among open models and at the same time outperforms much larger models on generative tasks, such as \llama{} 2 70B. By scaling further, \model{} \textsc{8x7B} is the best open generative language model on our task average, while only using 13B parameters at inference. Further, as our models use sliding window attention~\cite{child2019generating,beltagy2020longformer} they can handle generative and embedding inputs of arbitrary length.\\
\textbf{b) Efficiency:} Generative and embedding models are commonly used together to make up for each other's deficiencies~\cite{guu2020realm,lewis2021retrievalaugmented}. One such scenario is Retrieval-Augmented Generation (RAG)~\cite{lewis2021retrievalaugmented}, where an embedding model is used to retrieve context that is provided to the generative model to answer a user query. This requires passing the user query and the context into both the generative and the embedding model for a total of four forward passes. With \model{}, the embedding and generative model are equivalent, allowing us to cache computations and halve the necessary number of forward passes. We find that this can lead to >~60\% faster RAG at inference with long documents.\\
\textbf{c) Simplicity:} Currently, API providers such as OpenAI provide separate generative and embedding endpoints. This requires separate load balancing, additional storage, and more complex serving software. A single model that handles both use cases significantly simplifies infrastructure needs.

Compared to generative instruction tuning, the main downside of \method{} is that it requires more finetuning compute due to training with two objectives. However, finetuning is generally cheap compared to pretraining, thus we think the benefits vastly outstrip this problem. Further, when considering training a separate generative and embedding model from scratch (e.g. for RAG), GritLM is generally cheaper when incorporating the pretraining compute, as there is only one pretraining and finetuning for GritLM, not separate ones for both a generative and an embedding model. Thus we recommend practitioners building instruction-following language models to adopt \method{}.

Alongside GRIT, we introduce novel performance improvements for embedding models including the use of bidirectional attention with mean pooling for LLM embeddings and making in-batch negatives stem from the same dataset rather than any dataset, as well as novelties for generative models such as mixing sample- and token-level loss aggregation. We ablate these in detail in \autoref{sec:ablations}. We also propose new ways to reduce memory requirements during embedding model training in \autoref{sec:embmem}.

\section{\method{}}
\label{sec:method}

\begin{figure*}[htbp]
\centering
\begin{center}
\includegraphics[width=\textwidth]{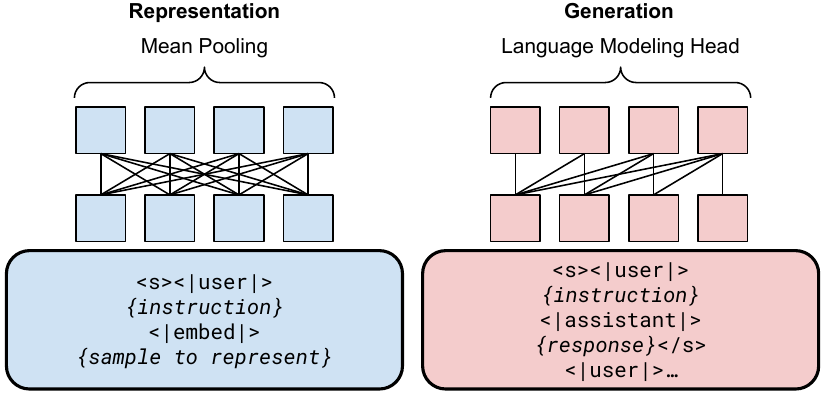}
\caption{\textbf{\model{} architecture and format.} \emph{Left:} \model{} uses bidirectional attention over the input for embedding tasks. Mean pooling is applied over the final hidden state to yield the final representation. \emph{Right:} \model{} uses causal attention over the input for generative tasks. A language modeling head on top of the hidden states predicts the next tokens. The format supports conversations with multiple turns (indicated with ``...'').}
\label{fig:format}
\end{center}
\end{figure*}

\method{} unifies representational instruction tuning~\cite{su2023embedder,asai2022taskaware,wang2024improving} and generative instruction tuning~\cite{wei2022finetuned,sanh2022multitask,muennighoff2023crosslingual} into a single model. We finetune a pretrained large language model~\cite{brown2020language} with embedding and generative instruction data in a consistent format as depicted in \autoref{fig:format}. For embedding data, we follow prior work and compute the loss using a contrastive objective with in-batch negatives~\cite{chen2020simple,gao2022simcse}:
\begin{equation}\label{eq:rep}
\mathcal{L}_{\text{Rep}} = -\frac{1}{M}\sum_{i=1}^M\log\frac{\exp(\tau \cdot \sigma(f_{\theta}(q^{(i)}), f_{\theta}(d^{(i)})))}{\sum_{j=1}^M\exp(\tau \cdot \sigma(f_{\theta}(q^{(i)}), f_{\theta}(d^{(j)})))}
\end{equation}
where $f$ is \model{} parametrized by the model $\theta$, $\tau$ is a temperature hyperparameter and $\sigma$ corresponds to pooling applied to each output followed by cosine similarity. $q$ and $d$ are query and document samples. As depicted in \autoref{fig:format}, we use bidirectional attention followed by mean pooling, which corresponds to averaging the hidden states across the sequence length. During pooling, we only average the final hidden states of the input sample, ignoring the instruction and format tokens. However, the instruction and format tokens still influence the final representation through the self-attention mechanism~\cite{vaswani2023attention}. 

To compute the loss on generative data, we use the language modeling objective whereby the model needs to predict the next token~\cite{radford2018improving,radford2019language}:
\begin{equation}\label{eq:gen}
\mathcal{L}_{\text{Gen}} = -\frac{1}{N}\sum_{i=1}^{N} \log P(f_{\theta, \eta}(x^{(i)}) | f_{\theta, \eta}(x^{(<i)}))
\end{equation}
where $f$ is \model{} parametrized by the model $\theta$ and the language modeling head $\eta$, which is only used for generation. $x$ are generative training samples. We only compute loss over predicted tokens i.e. ``\verb|{response}</s>|'' in \autoref{fig:format}. A key consideration is whether the generative loss is aggregated at the sample or token level. Aggregating at the sample level corresponds to giving each sample the same weight within a batch regardless of its token count. Such aggregation is commonly used for instruction tuning, as it can boost performance on discriminative tasks~\cite{muennighoff2023crosslingual}. However, \citet{muennighoff2023crosslingual} also show how this in turn leads to a model biased toward short generations. Meanwhile, aggregation at the token level corresponds to giving each token the same weight, thus samples with many tokens become more important. This usually leads to a model producing longer generations, which can be important for performance on generative tasks. Especially, human or machine-evaluated generative tasks, such as AlpacaEval~\cite{alpaca_eval}, are known to be biased toward preferring longer generations~\cite{wang2023far}. Note that when every sample has the same sequence length such as during pretraining or when the batch size is 1, token and sample level generative loss are equal to each other. One can also mix the two to balance their trade-offs, for example doing token level loss across a subset of the batch and then giving each subset the same weight. We explore the trade-offs in our ablations in \autoref{sec:ablations}. We sum the objectives with optional loss weights $\lambda_{\text{Rep}}$ and $\lambda_{\text{Gen}}$:
\begin{equation}\label{eq:grit}
\mathcal{L}_{\method{}} = \lambda_{\text{Rep}}\mathcal{L}_{\text{Rep}} + \lambda_{\text{Gen}}\mathcal{L}_{\text{Gen}}
\end{equation}
Notably, our formulation supports differing numbers of embedding samples ($M$) and generative samples/tokens ($N$). This allows for significantly increasing the embedding batch size while keeping the generative batch size fixed. A large embedding batch size is often key to well-performing text embedding models~\cite{xiao2023cpack}. However, it comes at the cost of requiring more compute at each step.


\section{Experiments}
\label{sec:exp}

In this section, we first outline our experimental setup in \autoref{sec:setup}. In \autoref{sec:main}, we discuss and benchmark the embedding and generative performance of our models. Finally, in \autoref{sec:ablations}, we ablate the settings that led to our final models, including training data, precision, pooling, sequence length, and loss weights.

\subsection{Setup}
\label{sec:setup}

We finetune our final models from Mistral 7B~\cite{jiang2023mistral} and Mixtral 8x7B~\cite{jiang2024mixtral} using adaptations of E5~\cite{wang2024improving} and the \tulu{} 2 data~\cite{ivison2023camels}. For E5, we adapt it by adding S2ORC~\cite{lo2020s2orc} to increase its scientific data (``E5S''), while for \tulu{} 2 we filter out their custom prompts that contain answers related to the origin of their model. For \modelbase{}, we use a batch size of 2048 for embedding data and 256 for generative data and we train the model for a total of 1253 steps corresponding to one epoch on the generative data and 1.36 epochs on the embedding data. For \modelbig{}, the embedding batch size is 256 due to compute limitations. We use several strategies to reduce the memory required during training including a novel technique to split the embedding triplet into separate forward and backward passes detailed in \autoref{sec:embmem}. Other hyperparameters are detailed in the ablation experiments in \autoref{sec:ablations} and \autoref{sec:hps}. 

For embedding performance we evaluate using the 56 main datasets from MTEB~\cite{muennighoff2023mteb}. For generative performance, we largely follow the evaluation setup of \citet{ivison2023camels} except that we use the HumanEvalSynthesize~\cite{muennighoff2023octopack} variant of HumanEval, as it is more adequate for instruction-following models. We explain each task in more detail in \autoref{sec:eval}.

\subsection{Main Results}
\label{sec:main}

\begin{table}[htbp]
\centering
\caption{\textbf{Embedding performance of \model{} and others}. We indicate parameter counts where available (B=billions). See \autoref{sec:eval} for task, metric, and dataset details. \autoref{sec:mtebresults} contains per-dataset results of \model{} models. LLMs not finetuned for embedding (\llama{} 2 70B, Mistral 7B (Instruct), GPT-J 6B, Gen.-only) are evaluated with weighted-mean pooling~\cite{muennighoff2022sgpt}. $^\varheart$Results from the MTEB leaderboard \scriptsize{(\url{https://hf.co/spaces/mteb/leaderboard})}}
\resizebox{\textwidth}{!}{
\begin{tabular}{l|ccccccc|c}
\toprule
Task ($\rightarrow$) & CLF & Clust. & PairCLF & Rerank & Retrieval & STS & Summ. & Avg. \\
Metric ($\rightarrow$) & Acc. & V-Meas. & AP & MAP & nDCG & Spear. & Spear. & \\
Dataset \# ($\rightarrow$) & 12 & 11 & 3 & 4 & 15 & 10 & 1 & 56 \\
\midrule
\multicolumn{9}{c}{Proprietary models$^\varheart$} \\
\midrule
OpenAI v3 & 75.5 & 49.0 & 85.7 & 59.2 & 55.4 & 81.7 & 29.9 & 64.6 \\
\midrule
\multicolumn{9}{c}{Other Open Models$^\varheart$} \\
\midrule
\llama{} 2 70B & 60.4 & 29.0 & 47.1 & 38.5 & 9.0 & 49.1 & 26.1 & 35.6 \\
Mistral 7B & 63.5 & 34.6 & 53.5 & 43.2 & 13.2 & 57.4 & 19.7 & 40.5 \\
Mistral 7B Instruct & 67.1 & 34.6 & 59.6 & 44.8 & 16.3 & 63.4 & 25.9 & 43.7 \\
GPT-J 6B & 66.2 & 39.0 & 60.6 & 48.9 & 19.8 & 60.9 & 26.3 & 45.2 \\
SGPT BE 5.8B & 68.1 & 40.3 & 82.0 & 56.6 & 50.3 & 78.1 & 31.5 & 58.9 \\
Instructor XL 1.5B & 73.1 & 44.7 & 86.6 & 57.3 & 49.3 & 83.1 & \textbf{32.3} & 61.8 \\
BGE Large 0.34B & 76.0 & 46.1 & 87.1 & 60.0 & 54.3 & 83.1 & \underline{31.6} & 64.2 \\
E5 Mistral 7B & 78.5 & 50.3 & \textbf{88.3} & 60.2 & 56.9 & \textbf{84.6} & 31.4 & 66.6 \\
\midrule
\multicolumn{9}{c}{\textbf{\model{}}} \\
\midrule
Gen.-only 7B & 65.4 & 32.7 & 54.2 & 43.0 & 13.7 & 60.2 & 21.1 & 41.2 \\
Emb.-only 7B & \underline{78.8} & \textbf{51.1} & 87.1 & \textbf{60.7} & \textbf{57.5} & \underline{83.8} & 30.2 & \underline{\textbf{66.8}}\\
\model{} \textsc{7B} & \textbf{79.5} & \underline{50.6} & \underline{87.2} & \underline{60.5} & \underline{57.4} & 83.4 & 30.4 & \underline{\textbf{66.8}} \\
\model{} \textsc{8x7B} & 78.5 & 50.1 & 85.0 & 59.8 & 55.1 & 83.3 & 29.8 & 65.7 \\
\bottomrule
\end{tabular}
}
\label{tab:emb}
\end{table}

\begin{table}[htbp]
\centering
\caption{\textbf{Generative performance of \model{} and others}. We indicate parameter counts where available (B=billions). See \autoref{sec:eval} for dataset, setup, metric details, and abbreviations. $^\varheart$Results from \citet{ivison2023camels} except for numbers marked with $^\vardiamond$ which are from~\citet{touvron2023llama} and $^\dagger$ which are from us. For models that cannot be easily used as chat models, we set Alpaca to 0.}
\resizebox{\textwidth}{!}{
\begin{tabular}{l|llllll|l}
\toprule
Dataset ($\rightarrow$) & MMLU & GSM8K & BBH & TyDi QA & HumanEval & Alpaca & Avg. \\
Setup ($\rightarrow$) & 0 FS & 8 FS, CoT & 3 FS, CoT & 1 FS, GP & 0 FS & 0 FS, 1.0 & \\
Metric ($\rightarrow$) & EM & EM & EM & F1 & pass@1 & \% Win & \\
\midrule
\multicolumn{8}{c}{Proprietary models$^\varheart$} \\
\midrule
GPT-4-0613 & 81.4 & 95.0 & 89.1 & 65.2 & 86.6$^{\dagger}$ & 91.2 & 84.8 \\
\midrule
\multicolumn{8}{c}{Other Open Models$^\varheart$} \\
\midrule
GPT-J 6B & 27.7 & 2.5 & 30.2 & 9.4 & 9.8 & 0.0 & 13.3 \\
SGPT BE 5.8B & 24.4 & 1.0 & 0.0 & 22.8 & 0.0 & 0.0 & 8.0 \\
Zephyr 7B $\beta$ & 58.6 & 28.0 & 44.9 & 23.7 & 28.5 & 85.8 & 44.9 \\
\llama{} 2 7B & 41.8 & 12.0 & 39.3 & 51.2 & 12.8$^\vardiamond$ & 0.0 & 26.2 \\
\llama{} 2 13B & 52.0 & 25.0 & 48.9 & 56.5 & 18.3$^\vardiamond$ & 0.0 & 33.5 \\
\llama{} 2 70B & 64.5 & 55.5 & 66.0 & \textbf{62.6} & 29.9$^\vardiamond$ & 0.0 & 46.4 \\
\llama{} 2 Chat 13B & 53.2 & 9.0 & 40.3 & 32.1 & 19.6$^{\dagger}$ & 91.4 & 40.9 \\
\llama{} 2 Chat 70B & 60.9 & 59.0 & 49.0 & 44.4 & 34.3$^{\dagger}$ & \underline{94.5} & 57.0 \\
\tulu{} 2 7B & 50.4 & 34.0 & 48.5 & 46.4 & 24.5$^{\dagger}$ & 73.9 & 46.3 \\
\tulu{} 2 13B & 55.4 & 46.0 & 49.5 & 53.2 & 31.4 & 78.9 & 52.4 \\
\tulu{} 2 70B & \underline{67.3} & \textbf{73.0} & \underline{68.4} & 53.6 & 41.6 & 86.6 & \underline{65.1} \\
Mistral 7B & 60.1 & 44.5 & 55.6 & 55.8 & 30.5 & 0.0 & 41.1 \\
Mistral 7B Instruct & 53.0 & 36.0 & 38.5 & 27.8 & 34.0 & 75.3 & 44.1 \\
Mixtral 8x7B Instruct & \textbf{68.4} & \underline{65.0} & 55.9 & 24.3 & \textbf{53.5} & \textbf{94.8} & 60.3 \\
\midrule
\multicolumn{8}{c}{\textbf{\model{}}} \\
\midrule
Emb.-only 7B & 23.5 & 1.0 & 0.0 & 21.0 & 0.0 & 0.0 & 7.6 \\
Gen.-only 7B & 57.5 & 52.0 & 55.4 & 56.6 & 34.5 & 75.4 & 55.2 \\
\model{} \textsc{7B} & 57.6 & 57.5 & 54.8 & 55.4 & 32.8 & 74.8 & 55.5 \\
\model{} \textsc{8x7B} & 66.7 & 61.5 & \textbf{70.2} & \underline{58.2} & \underline{53.4} & 84.0 & \textbf{65.7}\\
\bottomrule
\end{tabular}
}
\label{tab:gen}
\end{table}

\paragraph{\method{} leads to a state-of-the-art embedding and generative model} We benchmark \modelbase{}, \modelbig{} and generative- and embedding-only variants with other models in \autoref{tab:emb} and \autoref{tab:gen}. We find that \modelbase{} outperforms all prior open models on the Massive Text Embedding Benchmark~\citep{muennighoff2023mteb} while still outperforming all generative models up to its size of 7 billion parameters. \method{} models are the only ones that can handle both embedding and generation at best-in-class performance (\autoref{fig:performance}). For example, using \llama{} 70B~\cite{touvron2023llama} for embedding leads to a score of only 35.6 on MTEB as depicted in \autoref{tab:emb}. \model{} almost doubles that performance on MTEB leading to state-of-the-art performance, while still outperforming \llama{} 70B on generative tasks by more than 20\% (\autoref{tab:gen}). Scaling even further, \modelbig{} outperforms all openly available models on our generative average. Its embedding performance slightly decreases from \modelbase{}. This is likely because we had to decrease its embedding batch size from 2048 for \modelbase{} to only 256 for \modelbig{} due to compute limitations (\autoref{sec:setup}). We also train embedding-only and generative-only variants of \model{} that only use representational or generative instruction tuning but are otherwise equivalent. Benchmarking the embedding-only variant or SGPT BE 5.8B~\cite{muennighoff2022sgpt} on generative tasks in \autoref{tab:gen} by simply re-adding the language modeling head that was dropped during embedding finetuning leads to around random performance (25.0 is the random baseline on MMLU). Similarly, benchmarking the embedding performance of the generative-only model only leads to a score of 41.2 in \autoref{tab:emb}. Thus, joint optimization via the \method{} approach is critical to achieve strong performance for both embedding and generation. We note, however, that with 7 billion parameters \modelbase{} is significantly more costly to run than many other embedding models in \autoref{tab:emb}, such as BGE Large with only 335 million parameters~\cite{xiao2023cpack}. In addition, \modelbase{} produces representations of 4096 dimensions, which require 4$\times$ more storage than the 1024-dimensional embeddings of BGE Large.

\paragraph{\model{} matches embedding-only and generative-only variants} We find that unifying the two objectives via \model{} matches both the generative-only and the embedding-only variants. This is similar to observations made for visual models~\cite{yu2022coca}. However, while \model{} is trained for the same number of steps as the embedding-only and generative-only variants, it requires more compute per training step as it does a forward and backward pass on both embedding and generative data.

\begin{wraptable}{l}{7cm}
\caption{\textbf{Reranking (Rerank) using \model{} as both Bi- and Cross-Encoder.}}
\begin{tabular}{l|cc}
\toprule
MTEB DS ($\downarrow$) & No Rerank & Rerank top 10 \\
\midrule
ArguAna & 63.24 & \textbf{64.39} \\
ClimateFEVER & 30.91 & \textbf{31.85} \\
CQADupstack & 49.42 & \textbf{50.05} \\
DBPedia & 46.60 & \textbf{47.82} \\
FiQA2018 & 59.95 & \textbf{60.39} \\
FEVER & 82.74 & \textbf{82.85} \\
HotpotQA & 79.40 & \textbf{80.46} \\
NFCorpus & 40.89 & \textbf{41.23} \\
NQ & 70.30 & \textbf{71.49} \\
MSMARCO & 41.96 & \textbf{42.47} \\
QuoraRetrieval & \textbf{89.47} & 88.67 \\
SCIDOCS & 24.41 & \textbf{24.54} \\
SciFact & 79.17 & \textbf{79.28} \\
TRECCOVID & 74.80 & \textbf{75.24} \\
Touche2020 & 27.93 & \textbf{28.41} \\
\midrule
Average & 57.4 & \textbf{57.9} \\
\bottomrule
\end{tabular}
\label{tab:rr}
\end{wraptable}

\paragraph{Reranking with \model{}} For retrieval tasks, it is common to follow the embedding-based retrieval stage by a reranking stage~\cite{nogueira2020passage}. In the reranking stage, for each query, the top-$k$ chosen documents are reranked based on a usually more expensive but more performant method. For LLMs, prior work has shown that this can be done by passing each of the $k$ documents together with the query to the model and scoring the pair with log probabilities~\cite{muennighoff2022sgpt}. Note that this scales quadratically with the number of documents and queries and is thus usually too expensive for the first stage (``Cross-Encoder''). Meanwhile, using embeddings for the first stage is much cheaper as it only requires passing each query and each document once and thus scales linearly (``Bi-Encoder''). More recent work relies on instructions to use LLMs for reranking~\cite{sun2023chatgpt,ma2023zeroshot,pradeep2023rankvicuna,pradeep2023rankzephyr}. While prior work uses separate models for the embedding and reranking stages, \model{} can be used for both stages due to its unified capabilities. In \autoref{tab:rr}, we display the retrieval performance of \modelbase{} when additionally allowing it to rerank its top 10 documents for each query. For reranking, we use the model's generative capabilities following the permutation generation approach from \citet{sun2023chatgpt} and reusing their prompt. We find that reranking via the generative capabilities of \modelbase{} allows it to improve on its own embedding performance on almost every retrieval dataset. Increasing the top-$k$ documents beyond ten is likely to further improve results, however, at the cost of more compute~\cite{muennighoff2022sgpt}.

\begin{wraptable}{r}{7cm}
\vspace{-1.5em}
\caption{\textbf{Few-shot embedding.} The 12 MTEB datasets (``DS'') are grouped by the 7 main MTEB tasks in the same order as in \autoref{tab:emb}.}
\begin{tabular}{l|cc|cc}
\toprule
Train DS ($\rightarrow$) & \multicolumn{2}{c}{E5S} & \multicolumn{2}{c}{MEDI2} \\
MTEB DS ($\downarrow$) & 0 FS & 1 FS & 0 FS & 1 FS \\
\midrule
Banking77 & \textbf{88.5} & 88.3 & \textbf{88.1} & 87.9 \\
Emotion & \textbf{52.8} & 51.0 & \textbf{52.5} & 51.9 \\
IMDB & \textbf{95.0} & 93.9 & \textbf{94.3} & 92.2 \\
\midrule
BiorxivS2S & \textbf{39.8} & 39.4 & \textbf{37.6} & 37.4 \\
\midrule
SprintDup. & 93.0 & \textbf{94.9} & 95.2 & \textbf{95.7} \\
TwitterSem & \textbf{81.1} & 77.9 & \textbf{76.8} & 73.9 \\
TwitterURL & \textbf{87.4} & 87.1 & 85.9 & \textbf{86.1} \\
\midrule
ArguAna & \textbf{63.2} & 51.7 & \textbf{53.5} & 53.2 \\
SCIDOCS & \textbf{24.4} & 19.7 & \textbf{25.5} & 25.5 \\
\midrule
AskUbuntu & \textbf{67.3} & 64.7 & \textbf{66.6} & 66.0 \\
\midrule
STS12 & 77.3 & \textbf{78.0} & \textbf{76.6} & 73.5 \\
\midrule
SummEval & \textbf{30.4} & 29.5 & 29.1 & \textbf{31.5} \\
\bottomrule
\end{tabular}
\label{tab:fs}
\vspace{-2em}
\end{wraptable}

\paragraph{Few-shot embedding does not work} For generative models it has been well-established that providing in-context examples (``few-shots'', FS) improves performance \cite{brown2020language}. However, to the best of our knowledge, there has been no work on in-context learning with embedding models. In \autoref{tab:fs}, we benchmark the default 0-shot format versus providing a single few-shot example following the task instruction. We take the few-shot example from the respective evaluation dataset (see \autoref{sec:promptsembfewshot} for the prompts). We find that providing few-shot examples overall worsens performance. While there are small gains among PairClassification tasks (SprintDup. and TwitterURL), these are marginal and inconsistent. For the model trained on MEDI2, we even include few-shot embedding samples in the training data for around 5\% of training samples. However, the model seems not to have learned to make good use of the few-shot examples.\\

\paragraph{Aligning \model{}} It is common to follow the instruction finetuning stage of generative language models by an alignment tuning stage using methods like PPO~\cite{schulman2017proximal}, DPO~\cite{rafailov2023direct}, or KTO~\cite{ethayarajh2024kto} (``HALOs''~\cite{ethayarajh2024kto}). We experiment with further finetuning \model{} using KTO and benchmark the resulting models in \autoref{tab:kto}. We train using the binarized UltraFeedback dataset~\cite{cui2023ultrafeedback}. During this KTO stage, no further embedding training is performed, thus it leads to a slight performance drop on the MTEB average (66.8 to 66.7 and 65.7 to 65.2). However, the average generative performance of the KTO-tuned models is stronger. Notably, AlpacaEval jumps by >10 points for both models. On the more recent Alpaca 2.0~\cite{dubois2024length}, GritLM-8x7B-KTO has a length-controlled win rate of 18.5 with an average length of 1662 (not depicted). Thus, the KTO-finetuned models may be more useful for use cases where the generative performance is more important. Future work may consider continuing the embedding training during the alignment tuning stage. It may also be possible to develop an alignment tuning method specifically for embedding performance and combine it with generative alignment via KTO.

\begin{table}[htbp]
\centering
\caption{\textbf{Aligning \model{} with KTO after GRIT.} The upper table depicts embedding performance while the lower depicts generative performance.}
\resizebox{\textwidth}{!}{
\begin{tabular}{l|ccccccc|c}
\toprule
Task ($\rightarrow$) & CLF & Clust. & PairCLF & Rerank & Retrieval & STS & Summ. & Avg. \\
Metric ($\rightarrow$) & Acc. & V-Meas. & AP & MAP & nDCG & Spear. & Spear. & \\
Dataset \# ($\rightarrow$) & 12 & 11 & 3 & 4 & 15 & 10 & 1 & 56 \\
\midrule
\modelbase{} & 79.5 & 50.6 & 87.2 & 60.5 & 57.4 & 83.4 & 30.4 & 66.8 \\ 
\modelbase{} KTO & 79.6 & 50.1 & 87.1 & 60.5 & 57.1 & 83.5 & 30.5 & 66.7 \\ 
\modelbig{} & 78.5 & 50.1 & 85.0 & 59.8 & 55.1 & 83.3 & 29.8 & 65.7 \\ 
\modelbig{} KTO & 78.7 & 50.0 & 84.4 & 59.4 & 54.1 & 82.5 & 30.8 & 65.2 \\ 
\bottomrule
\end{tabular}
}
\resizebox{\textwidth}{!}{
\begin{tabular}{l|llllll|l}
\toprule
Dataset ($\rightarrow$) & MMLU & GSM8K & BBH & TyDi QA & HumanEval & Alpaca & Avg. \\
Setup ($\rightarrow$) & 0 FS & 8 FS, CoT & 3 FS, CoT & 1 FS, GP & 0 FS & 0 FS, 1.0 & \\
Metric ($\rightarrow$) & EM & EM & EM & F1 & pass@1 & \% Win & \\
\midrule
\modelbase{} & 57.6 & 57.5 & 54.8 & 55.4 & 32.8 & 74.8 & 55.5 \\
\modelbase{} KTO & 57.6 & 57.5 & 55.4 & 55.8 & 31.5 & 86.7 & 57.4 \\
\modelbig{} & 66.7 & 61.5 & 70.2 & 58.2 & 53.4 & 84.0 & 65.7 \\
\modelbig{} KTO & 66.8 & 79.5 & 67.1 & 31.4 & 56.8 & 95.3 & 66.2 \\
\bottomrule
\end{tabular}
}
\label{tab:kto}
\end{table}

\FloatBarrier

\subsection{Ablations}
\label{sec:ablations}

\begin{table*}[htbp]
\centering
\subfloat[
Attention and pooling ablations. Wmean is position-weighted mean pooling~\cite{muennighoff2022sgpt}.
\label{tab:attn}
]{
\centering
\begin{minipage}{\linewidth}{\begin{center}
\begin{tabular}{ccccc|cc}
\toprule 
\multicolumn{2}{c}{Attention Emb} & \multicolumn{2}{c}{Attention Gen} & \multirow{2}{*}{Pooling} & \multirow{2}{*}{Emb} & \multirow{2}{*}{Gen} \\
Instruction & Sample & Instruction & Sample & & & \\
\midrule
\multicolumn{7}{c}{\textit{Embedding Only}} \\
\midrule
\multicolumn{2}{c}{Causal} & & & Wmean & 60.0 & - \\
Causal & Bidirectional & & & Mean & 61.0 & - \\
\multicolumn{2}{c}{Bidirectional} & & & Mean & 61.8 & - \\
\midrule
\multicolumn{7}{c}{\textit{Generative Only}} \\
\midrule
& & \multicolumn{2}{c}{Causal} & & - & \textbf{55.2} \\
& & Bidirectional & Causal & & - & 50.7 \\
\midrule
\multicolumn{7}{c}{\textit{Unified}} \\
\midrule
\multicolumn{2}{c}{Causal} & \multicolumn{2}{c}{Causal} & Last token & 61.2 & 53.0 \\    
\multicolumn{2}{c}{Causal} & \multicolumn{2}{c}{Causal} & Wmean & 62.8 & 52.8 \\
\multicolumn{2}{c}{\textbf{Bidirectional}} & 
\multicolumn{2}{c}{\textbf{Causal}} & \textbf{Mean} & \textbf{64.0} & 52.9 \\
\bottomrule
\end{tabular}
\end{center}}\end{minipage}
}
\\
\centering
\subfloat[
Base model
\label{tab:basemodel}
]{
\begin{minipage}{0.301\linewidth}{\begin{center}
\begin{tabular}{c|cc}
\toprule 
Variant & Emb & Gen \\
\midrule
\textbf{Mistral 7B} & \textbf{54.6} & \textbf{22.4} \\
\llama{} 2 7B & 48.2 & 20.8 \\
GPT-J 6B & 51.9 & 14.0 \\
\bottomrule
\end{tabular}
\end{center}}\end{minipage}
}
\subfloat[
Embedding dataset
\label{tab:embds}
]{
\begin{minipage}{0.301\linewidth}{\begin{center}
\begin{tabular}{c|cc}
\toprule 
Dataset & Emb \\
\midrule
MEDI & 64.0 \\
MEDI2 & 64.7 \\
\textbf{E5} & \textbf{66.0} \\    
\bottomrule
\end{tabular}
\end{center}}\end{minipage}
}
\subfloat[
Generative dataset
\label{tab:gends}
]{
\begin{minipage}{0.301\linewidth}{\begin{center}
\begin{tabular}{c|c}
\toprule 
Dataset & Gen \\
\midrule
\textbf{\tulu{} 2} & \textbf{55.2} \\
OASST & 37.7 \\
UltraChat & 47.4 \\
\bottomrule
\end{tabular}
\end{center}}\end{minipage}
}
\\
\centering
\subfloat[
Embedding head
\label{tab:embhead}
]{
\begin{minipage}{0.301\linewidth}{\begin{center}
\begin{tabular}{c|cc}
\toprule 
Variant & Emb & Gen \\
\midrule
\textbf{No head} & \textbf{62.7} & \textbf{49.2} \\
-> 1024 & 62.1 & 48.0 \\
\bottomrule
\end{tabular}
\end{center}}\end{minipage}
}
\subfloat[
Batch size (BS)
\label{tab:bs}
]{
\centering
\begin{minipage}{0.301\linewidth}{\begin{center}
\begin{tabular}{c|cc}
\toprule 
BS Emb:Gen & Emb & Gen \\
\midrule
256:256 & 63.2 & \textbf{53.4} \\
4096:256 & \textbf{64.2} & 53.3 \\
\bottomrule
\end{tabular}
\end{center}}\end{minipage}
}
\subfloat[
Precision
\label{tab:precision}
]{
\begin{minipage}{0.301\linewidth}{\begin{center}
\begin{tabular}{c|cc}
\toprule 
Precision & Emb & Gen \\
\midrule
FP32 & 66.3 & 52.4 \\
\textbf{BF16$^*$} & \textbf{66.5} & \textbf{55.0}  \\
\bottomrule
\end{tabular}
\end{center}}\end{minipage}
}
\\
\centering
\subfloat[
In-batch negatives (IBN)
\label{tab:inbatch}
]{
\begin{minipage}{0.301\linewidth}{\begin{center}
\begin{tabular}{c|cc}
\toprule 
IBN origin & Emb & Gen \\
\midrule
Any dataset & \textbf{66.0} & 50.9 \\
\textbf{Same dataset} & \textbf{66.0}& \textbf{51.1} \\
\bottomrule
\end{tabular}
\end{center}}\end{minipage}
}
\subfloat[
Format
\label{tab:fmt}
]{
\begin{minipage}{0.301\linewidth}{\begin{center}
\begin{tabular}{c|cc}
\toprule 
Format & Gen \\
\midrule
\textbf{\tulu{} 2} & \textbf{55.2} \\
Zephyr $\beta$ & 49.0 \\
\bottomrule
\end{tabular}
\end{center}}\end{minipage}
}
\subfloat[
Emb training max tokens
\label{tab:sq}
]{
\centering
\begin{minipage}{0.301\linewidth}{\begin{center}
\begin{tabular}{c|cc}
\toprule 
Tokens & Emb & Gen \\
\midrule
512 & 64.1 & 52.2 \\
\textbf{2048} & \textbf{64.7} & \textbf{53.8} \\
\bottomrule
\end{tabular}
\end{center}}\end{minipage}
}
\\
\centering
\subfloat[
Loss ablations. $\mathcal{L}_{\text{Rep}}/\mathcal{L}_{\text{Gen}}$ is the loss ratio of the 1st step adjusted via $\lambda_{\text{Rep}}$ and $\lambda_{\text{Gen}}$. Mix refers to mixing sample and token level loss, e.g. (32->8) is token level loss across 32 samples and then sample level loss across 8 sub-batches for a total batch size of 256.
\label{tab:lossweights}
]{

\begin{minipage}{\linewidth}{\begin{center}
\begin{tabular}{cc|cc}
\toprule 
Gen loss type & $\mathcal{L}_{\text{Rep}}/\mathcal{L}_{\text{Gen}}$ & Emb & Gen \\
\midrule
Token & 2.4 & 66.1 & 54.4 \\
Token & 6.0 & 66.5 & 55.0 \\
\textbf{Mix (32~->~8)} & 4.1 & \textbf{66.7} & \textbf{55.4} \\ 
\bottomrule
\end{tabular}
\hspace{1em}
\begin{tabular}{c|c}
\toprule 
Gen loss type & AlpacaEval \\
\midrule
Mix (4~->~64) & 67.6 \\
\textbf{Mix (32~->~8)} & \textbf{74.7} \\
\bottomrule
\end{tabular}
\end{center}}\end{minipage}
}
\caption{\textbf{\method{} ablations.} Emb corresponds to the MTEB average, while Gen corresponds to the average across generative tasks (\autoref{sec:eval}). The embedding head variant ``->~1024'' corresponds to down-projecting the final hidden state with a linear layer from 4096 to 1024 dimensions, only for embedding tasks. BF16$^*$ means that some computations are still in FP32 as explained in \autoref{sec:ablations}. The setting chosen for \model{} is \textbf{bold}. Once an ablation was successful, we adopted its setting, thus the bold performance slightly varies from one table to the next. For example, the base model ablation (b) is done for just 100 hundred steps with sub-optimal formatting. Full results are in \autoref{sec:fullresults}.}
\label{tab:ablations} 
\end{table*}

\paragraph{Attention and pooling} We train \model{} starting from a pretrained decoder language model which has been trained with causal attention. Prior work has shown that while embeddings of causal LLMs are competitive, they are outperformed by BERT-like encoders with bidirectional attention at the same number of parameters~\cite{muennighoff2022sgpt,devlin2019bert}. This lines up with intuition, as bidirectional attention allows the model to adjust the representation of the first tokens based on information obtained from future tokens. Meanwhile, causal attention only allows information to propagate one way. Thus, for causal attention early tokens may yield poor representations due to a lack of understanding of the entire sample. To counter this issue, we experiment with adapting the model during finetuning to learn to use bidirectional attention. In \autoref{tab:ablations} we find that \textbf{adapting the causally pretrained LLM with bidirectional attention provides the best embedding performance}. For fully causal embeddings, we confirm findings from~\citet{muennighoff2022sgpt} that position-weighted mean pooling (``Wmean'') leads to better embedding performance than taking the embedding of the last token despite recent work finding the opposite~\cite{zhang2023language,ma2023finetuning}. For last token pooling, we follow \citet{zhang2023language} and use a special token. We find that adapting the model to be a PrefixLM~\cite{raffel2023exploring}, whereby the attention over the generative instruction is bidirectional but still causal for the response (``Sample'') worsens performance in contrast to prior work~\cite{wang2022language}. Thus, we stick with fully causal generation. The unified variant significantly outperforms the embedding-only variants, while underperforming the best generative-only variant. However, once we switched from MEDI to the E5 dataset in later ablations the embedding-only variant matched the unified variant. Meanwhile, the worse generative performance of the unified model was due to a suboptimal loss setting that we fixed in the loss ablations.

\paragraph{Base model} The \model{} approach generalizes to any generative language model, thus we ablate initializing from GPT-J 6B~\cite{wang2021gpt}, \llama{} 2 7B or Mistral 7B~\cite{jiang2023mistral}. Using Mistral 7B leads to the best performance for both embedding and generative tasks. For generative tasks, this is expected as the pretrained Mistral 7B performs the best among the three (\autoref{tab:gen}). However, for embedding tasks, GPT-J outperforms Mistral 7B (\autoref{tab:emb}). Thus, \textbf{the embedding performance of a pretrained model is not predictive of its embedding performance after finetuning}. Rather, its generative performance appears to be a more reliable indicator of its embedding performance after finetuning.

\paragraph{Generative dataset} We benchmark our filtered \tulu{} 2 introduced in \autoref{sec:setup}~\cite{ivison2023camels} with UltraChat~\cite{ding2023enhancing,tunstall2023zephyr} and the OpenAssistant version from OctoPack~\cite{muennighoff2023octopack,köpf2023openassistant,longpre2023data}. Using \tulu{} 2 leads to better performance on every generative task considered (see \autoref{sec:fullresults} for per-task results). This is likely due to \tulu{} 2 containing a larger diversity of tasks~\cite{ivison2023camels}. Another possible reason is that \tulu{} 2 may have been carefully tuned on the generative evaluation datasets, as we use largely the same evaluation setup as the creators of \tulu{} 2~\cite{ivison2023camels}.

\paragraph{Embedding dataset} We benchmark MEDI~\cite{su2023embedder}, a new version of MEDI with better negatives which we build and call MEDI2, and the E5 dataset~\cite{wang2024improving}. While MEDI and MEDI2 always preface instructions with ``Represent'' (see e.g. \autoref{fig:medi2}), the E5 dataset places no constraint on the instruction prefix (see e.g. \autoref{fig:e5}). Thus, when using the E5 dataset the ``<|embed|>'' formatting is critical to tell the model that it will be subject to the representation loss, not the generative loss~(\autoref{fig:format}). Further, MEDI and MEDI2 always contain instructions for both queries and documents, which we refer to as \textbf{two-sided instructions}. Meanwhile, the E5 dataset uses \textbf{one-sided instructions} for asymmetric datasets~\cite{muennighoff2022sgpt}, whereby the documents receive no instructions, only the queries. The advantage of not using document instructions is that the document corpus can be encoded once and then cached and reused across a variety of tasks. During training on E5, symmetric tasks are also in a one-sided setting, but we still evaluate them in the two-sided format. This should not be a problem as the cosine similarity function we use during training is transitive: if sentence A with instruction is similar to sentence B without instruction, and sentence B without instruction is similar to sentence C with instruction, then we can confidently say that sentence A with instruction is also similar to sentence C with instruction. As depicted in \autoref{tab:ablations}, using the E5 dataset performs best by a wide margin. An inspection of samples, suggests that this is likely due to its superior hard negatives and diversity of tasks generated by GPT-4 (\autoref{sec:datasetsamples}). For our final runs with the E5 dataset, we additionally add scientific data (\autoref{sec:setup}).

\paragraph{Embedding head} The cost of caching the embeddings of a large document corpus is directly proportional to the embedding dimensionality. To minimize such costs, we experiment with adding an embedding head consisting of a linear layer with activation that down-projects the embedding~\cite{ni2021large,muennighoff2022sgpt}. This layer is only used for embedding tasks. Down-projecting the embeddings four-fold (from 4096 to 1024) leads to an embedding performance decrease of around 1\%. This may be acceptable for certain use cases where the saved storage is more important. However, for our final model, we do not use such a head to keep it simple and achieve maximum performance. Search techniques~\cite{arya1998optimal,johnson2017billionscale,douze2024faiss} or dimensionality reduction techniques such as Principal Component Analysis still allow for reducing the embedding dimension of our final model post-training while maintaining most of the performance. Similar to the storage cost-performance trade-off we explore here, we hypothesize that there is a speed/cost-performance trade-off with taking the embedding from different layers of our model. For example, we could train using the embedding after half the layers of the model, thus speeding up the embedding model by 50\% while likely only incurring a small drop in embedding performance.

\paragraph{Batch size} Due to the utilization of in-batch negatives for contrastive training (\autoref{sec:method}), a larger batch size provides a more accurate gradient. Thus, scaling up the batch size is a key ingredient in most well-performing embedding models~\cite{xiao2023cpack,wang2022text}. We experiment with scaling up the embedding batch size to 4096 while keeping it at 256 for generative data. This leads to a 1.0 gain on the embedding average while generative performance remains stable. Especially the 15 retrieval datasets that are part of the embedding average benefit from the increase in batch size (see \autoref{tab:appbs}). For our final model, we use a batch size of 2048 for embedding and 256 for generative data.

\paragraph{Precision} The parameters of the Mistral 7B model are in bfloat16 (BF16) precision as it was pretrained in this format. We experiment with finetuning it with float32 (FP32) precision versus keeping the BF16 format and training with mixed precision. FP32 training is more costly, however, the additional precision may result in a better model. Our intuition is that more precision is important for embedding but not as much for generation. This is because while for generative tasks evaluated greedily, the model output is a discretionary argmax over the predictions of the language modeling head, for embedding tasks it is a continuous representation. Thus, small differences due to a lack of precision may not change the model's generation but will affect its representation. Hence, for embedding tasks, we always cast the hidden states to FP32 during the pooling operation and keep them this way for the similarity computation. Not keeping them in FP32 after pooling worsens performance slightly, but may be necessary for cheap storage (see \autoref{sec:precision}). In addition, some operations such as layer normalization~\cite{ba2016layer} are also performed in FP32 even for BF16 training due to PyTorch autocast~\cite{zhao2023pytorch}. In \autoref{tab:ablations}, we find that there is no benefit from doing even more computations in FP32 besides the ones listed above. Thus, we train and evaluate all our other models in BF16 mixed precision to speed up training and inference.

\paragraph{In-batch negatives} We always use in-batch negatives for embedding training (\autoref{sec:method}), however, we ablate whether or not they come from the same dataset. We hypothesize that making them all come from the same dataset leads to better negatives as the model needs to distinguish them based on more nuanced differences. In practice, we find that the average embedding performance remains around the same. However, we notice a 1.3 jump on the 15-dataset Retrieval average (\autoref{tab:appibn}). Thus, we stick with the variant where in-batch negatives stem from the same dataset.

\paragraph{Format} Our chosen format is depicted in \autoref{fig:format}, which is equivalent to \tulu{} 2~\cite{ivison2023camels} for generative tasks. We also benchmark the Zephyr $\beta$ format \cite{tunstall2023zephyr}, which has an additional end-of-sequence token (``\verb|</s>|'') after each user utterance. We find that it performs worse on generative tasks. The additional end-of-sequence after the user utterance increases the likelihood of the model generating another end-of-sequence token earlier than necessary. This significantly harms HumanEvalSynthesize performance and slightly reduces AlpacaEval, where long generations can be critical (see \autoref{sec:fullresults} for task-specific performance).


\paragraph{Max tokens} Our base model, Mistral 7B, can handle sequences of arbitrary length due to its sliding window attention~\cite{jiang2023mistral}. As finetuning with longer sequences is more expensive we ablate its benefits. We compare training with a maximum token limit of 512 versus 2048 for embedding documents. For embedding queries, we always use 256, and for generative data, we always use 2048. We find that increasing the embedding document sequence length during training slightly boosts performance on both embedding and generation even though we still evaluate embedding tasks with 512. This boost likely comes from our training data containing many documents beyond 512 tokens, which need to be truncated if the maximum sequence length is 512. Such truncation may remove the critical parts that make two texts a positive or a negative contrastive pair and thus hinder learning. As our embedding evaluation (MTEB) contains few documents longer than 512 tokens there is little truncation happening at evaluation~\cite{muennighoff2023mteb,günther2024jina,günther2023jina}. Note that just like their base models, our final models \modelbase{} and \modelbig{} can produce embeddings for sequences of arbitrary length. However, due to a lack of benchmarks, we do not know how well the embeddings of our models perform for input sequences longer than 512 tokens.

\paragraph{Loss ablations} As detailed in \autoref{sec:method}, we experiment with both token and sample level generative loss. Further, we ablate the representation and generative loss weights, $\lambda_{\text{Rep}}$ and $\lambda_{\text{Gen}}$. For the unified visual model CoCa, the authors find that giving a weight of 2 to generation and 1 to embedding boosts performance on both streams~\cite{yu2022coca}. However, rather than the weights, we argue that the loss ratio, $\mathcal{L}_{\text{Rep}}/\mathcal{L}_{\text{Gen}}$, is of more interest as it reveals which objective has a larger impact on the optimization of the model. We maintain a ratio of $\mathcal{L}_{\text{Rep}}/\mathcal{L}_{\text{Gen}}~>~1$ i.e. giving more weight to the representation loss. This is because the model has already been pretrained with the generative loss, thus we expect less additional generative training to be necessary. Meanwhile, the contrastive loss for embedding data is new to the model, thus we expect more learning to be needed on the embedding side. Further, the embedding loss drops off extremely quickly as can be seen in the loss graphs in \autoref{sec:loss}. Thus, even though the representation loss has a higher weight at the start, throughout training they have very similar weights with both hovering around a loss of 1.0. We find that mixing sample and token level generative loss leads to the best performance by a small margin. As expected in \autoref{sec:method}, token level loss to some degree is critical for good performance on AlpacaEval. For ``Mix (4~->~64)''  token level loss is applied across only 4 samples and then sample level loss across 64 sub-batches, which leads to a 7-point drop in AlpacaEval performance. This drop is accompanied by a decrease in median AlpacaEval generation length from 941 to 865. Thus, token level loss across many samples is critical to maintaining long generations, which directly impacts the AlpacaEval score.

\FloatBarrier

\section{RAG with \method{}}
\label{sec:rag}

\begin{figure*}[tbp]
\centering
\begin{center}
\includegraphics[width=\textwidth]{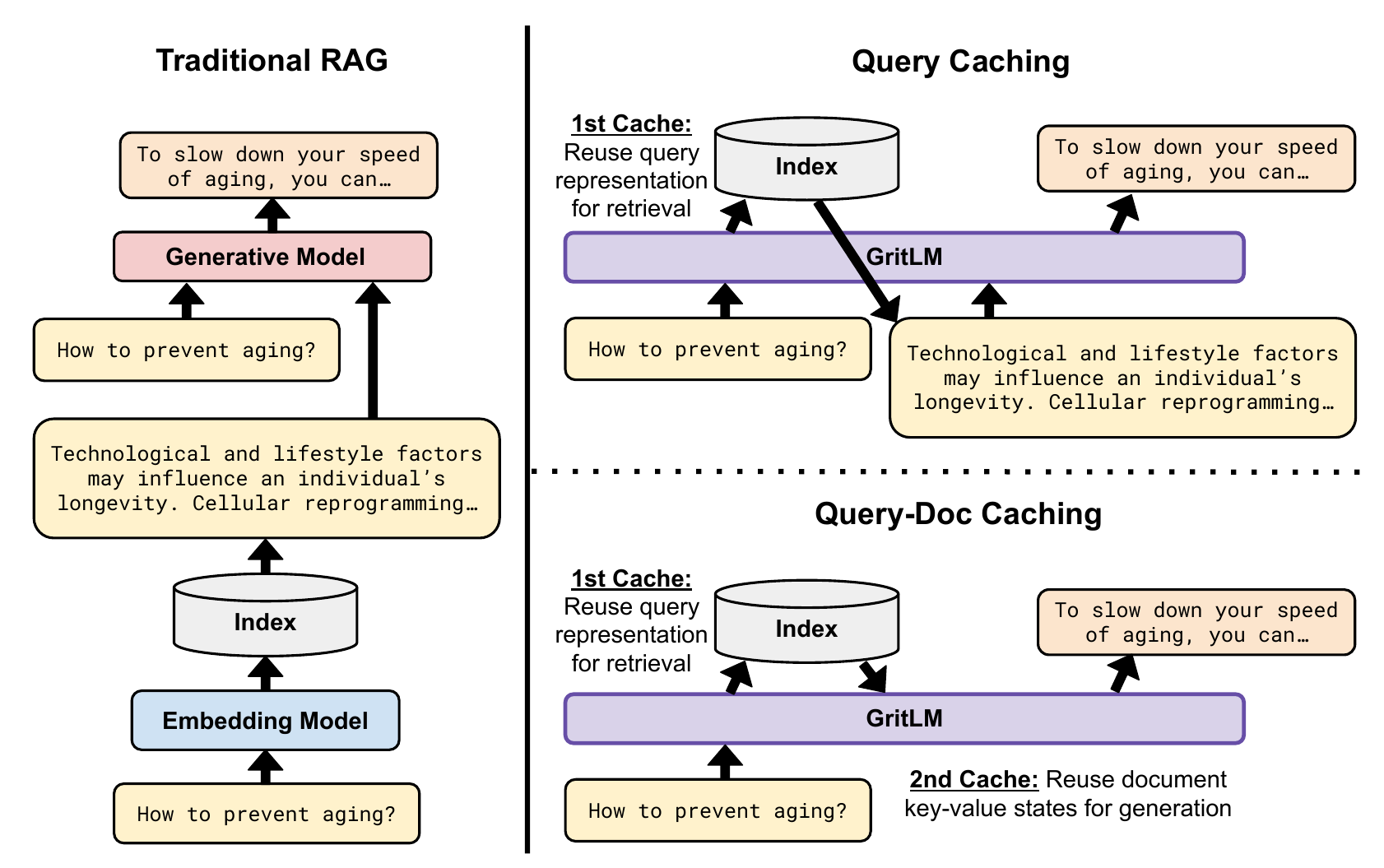}
\caption{\textbf{RAG with \method{}.} \emph{Left:} Traditional Retrieval-Augmented Generation (RAG) relies on a separate embedding model and generative model. \emph{Right:} \model{} simplifies RAG as it handles both embedding and generation. Query Caching removes the duplicate forward pass of the query by reusing its representation. Query-Doc Caching also removes the forward pass on the document during inference, as the cached index also stores the document key-value states.}
\label{fig:rag}
\end{center}
\end{figure*}

\paragraph{Method} By unifying embedding and generation, \model{} simplifies Retrieval-Augmented Generation (RAG). \autoref{fig:rag} displays how forward passes can be reduced by caching. Specifically, we break down the caching alternatives into:\\
\textbf{(a) Query Caching:} In traditional RAG, the query needs to be passed both through the embedding model and later through the generative model. In Query Caching, we cache the key-value states from the embedding forward pass and reuse them for the generative pass, exploiting the property that both are the same model: \model{}. Thus, we save compute equivalent to one forward pass of the query. Equivalently, we can also perform the generative forward pass over the query first and use its representation to retrieve the document on the fly (depicted in \autoref{fig:rag}). Note that Query Caching can be completely equivalent to RAG if the query is placed at the beginning of the prompt such that it only attends to itself through causal attention.\\
\textbf{(b) Doc Caching:} Here we cache the documents, D. When the index is created, we also save the key-value states of every document and add them to the index. Thus, the index consists of the document embeddings and key-value states. Note that the computational cost of creating the index remains the same as the key-value states have to be computed even if only embeddings are desired. At inference, we still retrieve based on embedding similarity but the index returns the key-value states instead of the text passage. These key-value states are then provided to the model to avoid having to recompute them. This effectively saves a forward pass for every in-context document at inference. However, this method increases the necessary storage. While the text passages no longer need to be stored, the key-value states now need to be stored and they usually require more storage depending on the model. We note that Document Caching also works for models other than \model{}. However, for such models, one needs to pass all documents through the generation model ahead of time, thus increasing the cost of creating the index. To maintain equivalence with RAG, the document should be at the beginning of the prompt for Document Caching (opposite of Query Caching).\\
\textbf{(b) Query-Doc Caching / Doc-Query Caching:} We can also combine Query Caching and Doc Caching to save even more inference costs. However, combining them inevitably leads to discrepancies compared to RAG, as in traditional RAG either the query or the document is conditioned on the other one. Meanwhile, if both are cached then they are not conditioned on one another via the self-attention mechanism. We refer to Query-Doc Caching if the query is followed by the document in the prompt and to Doc-Query Caching if the document comes first.

\FloatBarrier

\begin{table}[htbp]
\centering
\caption{\textbf{RAG benchmarking on Natural Questions with \modelbase{}.} For RAG, the retrieved context is simply placed in the context of the language model in contrast to our caching alternatives (\autoref{fig:rag}). We measure CPU and GPU latencies on an ``Intel(R) Xeon(R) Platinum 8481C CPU @ 2.70GHz'' and one ``NVIDIA H100 80GB HBM3'', respectively. Sample A has a query of 1 token and a document of 4000 tokens, and sample B is the inverse. For each approach, we generate 16 tokens. Storage consists of the index and passages, except for Doc Caching variants where it is the index and key-value states. The index is stored in float32, while key-value states are stored in bfloat16.}
\resizebox{\textwidth}{!}{
\begin{tabular}{l|c|cc|cc|c}
\toprule
& Match & \multicolumn{2}{c|}{CPU Latency (s, $\downarrow$)} & \multicolumn{2}{c|}{GPU Latency (s, $\downarrow$)} & Storage ($\downarrow$) \\
& (0-shot, $\uparrow$) & Sample A & Sample B & Sample A & Sample B & \\
\midrule
No RAG & 21.00 & 4.3 ± 0.36 & 13.69 ± 1.0 & 0.24 ± 0.04 & 0.38 ± 0.04 & \textbf{0GB} \\
\midrule
\multicolumn{7}{c}{\textit{Query then document prompt}} \\
\midrule
RAG & \underline{30.50} & 11.64 ± 0.74 & 14.88 ± 0.87 & 0.39 ± 0.02 & 0.40 ± 0.02 & \underline{43GB} \\
Query Caching & 25.46 & 18.30 ± 0.76 & 6.87 ± 0.89 & 0.44 ± 0.03
& \underline{\textbf{0.27 ± 0.02}} & \underline{43GB} \\
Query-Doc Caching & 21.63 & \textbf{5.12 ± 0.23} & \underline{6.62 ± 0.97} & \underline{0.27 ± 0.03} & 0.29 ± 0.01 & 30TB \\
\midrule
\multicolumn{7}{c}{\textit{Document then query prompt}} \\
\midrule
RAG & 30.47 & 14.18 ± 1.01 & 15.33 ± 0.87 & 0.39 ± 0.01 & 0.4 ± 0.01 & \underline{43GB} \\
Doc Caching & \textbf{33.38} & 5.25 ± 0.34 & 23.23 ± 1.05 & \underline{0.27 ± 0.03} & 0.45 ± 0.02 & 30TB \\
Doc-Query Caching & 18.39 & \underline{5.23 ± 0.37} & \textbf{6.41 ± 0.96} & \textbf{0.26 ± 0.03} & \underline{\textbf{0.27 ± 0.02}} & 30TB \\
\bottomrule
\end{tabular}
}
\label{tab:rag}
\end{table}

\begin{figure*}[htbp]
\centering
\begin{center}
\includegraphics[width=\textwidth]{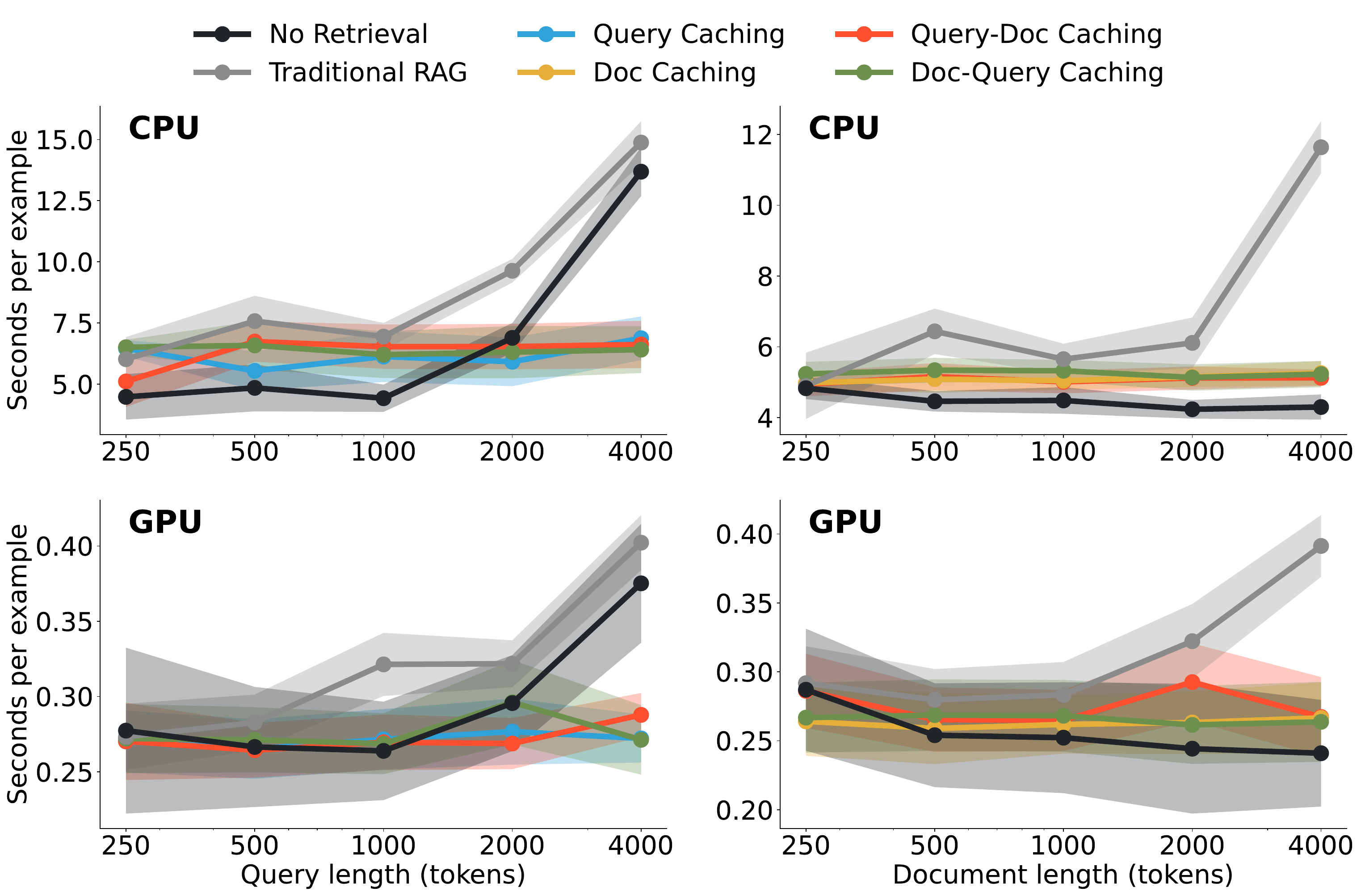}
\caption{\textbf{Inference latency of RAG with \modelbase{}.} When benchmarking scaling query length (left), document length is fixed at 1, whereas query length is fixed at 1 when scaling document length (right). In addition to the query/doc lengths, the formatting and prompt take up around 40 tokens. We visualize the standard deviation across 100 runs as the shaded area. For each approach, we generate 16 tokens. We measure CPU and GPU latencies on an ``Intel(R) Xeon(R) Platinum 8481C CPU @ 2.70GHz'' and one ``NVIDIA H100 80GB HBM3'', respectively.}
\label{fig:latency}
\end{center}
\end{figure*}


\paragraph{Setup} We benchmark the different caching variants using data from Natural Questions~\cite{kwiatkowski2019natural}. Our implementation is adapted from \citet{izacard2022atlas}, however, we use a significantly smaller index of only 2,681,468 documents stemming from the BEIR NQ corpus~\cite{thakur2021beir}. We score models using the match score, whereby we check if any of the correct answers are anywhere in the generation. Prior work usually uses exact match, whereby they check if the generation exactly matches the answer. However, as our model is more chatty, it tends to answer in a few sentences and thus exact match often fails to give credit to correct answers. Inspecting the first 20 samples of the ``No RAG'' baseline, we find that exact match leads to 4 false negatives that are correctly credited by the match metric. We did not find any false positives from choosing the match metric in those samples. We do not use any instructions for embedding, solely the format as presented in \autoref{fig:format}.

\paragraph{Performance} As depicted in \autoref{tab:rag}, RAG performs better than the ``No RAG'' baseline where the model is not provided any context. This validates that despite its small size compared to prior work~\cite{lin2023radit}, our index is still valuable. While Query and Doc Caching can theoretically lead to the exact same performance as RAG, we experience differences stemming from two reasons: \textbf{1) Attention:} Our model is trained to embed with bidirectional attention (\autoref{sec:method}) and thus we use bidirectional attention when embedding query or document. Meanwhile, the generative model expects causal key-value states. In the Query-Doc/Doc-Query setup, there is an additional mismatch in either the documents or the queries not having attended to the other one, as both need to be embedded and cached separately. \textbf{2) Formatting:} The query is formatted in the embedding format as depicted in \autoref{fig:format}, which the model has not seen in conjunction with a generative task during training. This could further lead to a performance drop. Due to 1) and 2), Query Caching leads to a performance drop compared to traditional RAG. However, the Query Caching performance of 25.46 is still better than not using RAG, thus it comes down to a speed-performance trade-off. Formatting the RAG baseline using the embedding format (\autoref{fig:format}) reduces its score from 30.50 to 29.36 (not depicted), thus the additional four-point discrepancy of Query Caching and the majority of the damage is because of the attention issue. Meanwhile, Doc Caching slightly improves performance resulting in the best match score among all methods considered. This is possibly because, unlike the query, the document does not need to be as thoroughly understood, and skimming it may suffice. Thus, the slightly corrupted key-value states do not result in a performance drop. Query-Doc and Doc-Query Caching only perform near the ``No RAG'' baseline in our experiments, which may limit their usefulness in practice. This is likely caused by the additional attention mismatch that they introduce. This issue as well as the formatting issue could likely be solved by an additional RAG finetuning stage on top of \model{}, which we leave to future work.

\paragraph{Latency} In \autoref{fig:latency}, we show how caching leads to significant speed-ups over RAG on both CPUs and GPUs for long sequences. If only 250 tokens are cached, however, we find the speed-up to be negligible. In \autoref{tab:rag}, we display that for 4000 tokens, Query Caching is 54\% and 33\% faster on CPUs and GPUs, respectively (Sample B). For Doc Caching it is 63\% and 31\% (Sample A). If going beyond 4000 tokens the speed-ups will be even larger. However, for the opposite samples in \autoref{tab:rag} speed remains around the same. This is because while for Sample A, Doc Caching caches 4000 tokens, for Sample B it caches only 1 token, which does not provide any speed-up. Thus, Doc Caching should be used when documents are expected to be very long, while Query Caching should be used when queries are expected to be very long. In a production setting, a simple input length check could switch from one caching mode to the other. As is the case in  \autoref{tab:rag}, caching can match or even be faster than not using retrieval at all (``No RAG''). This could be due to the embedding forward pass not using the language modeling head. For Query Caching, the language modeling head is only used for the tokens that are generated, while for ``RAG'' and ``No RAG'' it is used for the entire input. The matrix multiplication with the language modeling head is computationally expensive due to its high dimensionality, which could cause the slower speed of the no retrieval baseline. Query-Doc Caching and Doc-Query Caching cache both documents and queries and thus lead to major speed-ups for both Sample A and Sample B in \autoref{tab:rag}. Overall, speed-ups are larger on CPUs, as GPUs can process the entire sequence in parallel, thus the advantage of caching parts of it is smaller. We also note that our RAG baseline uses our 7B parameter model for both the embedding and generative model but without caching. In practice, it is often common to have an embedding model that is much smaller and cheaper than the generative model. Nonetheless, as caching with \model{}-7B approaches the No RAG latency in \autoref{tab:rag}, we still expect it to be faster than setups with smaller embedding models for long sequences. In addition, it would lead to significantly better performance in that case due to the state-of-the-art retrieval performance of \model{}.

\paragraph{Storage} In most RAG setups the embeddings of all documents are computed ahead of time and stored to be later used at inference. This is referred to as the index. In traditional RAG, the documents themselves still need to be stored, as the index is only used for finding the document ID, which is then used to fetch the document text and pass it to the generative model. For Doc Caching variants documents no longer need to be stored, however, the key-value states need to be stored together with the index. The key-value states take up a lot of storage, as for each batch they consist of two tensors of shape (batch size, number of heads, sequence length, dimension per head). For our 2,681,468 documents and the 7-billion parameter \model{} model, this leads to around 30TB of key-value states. However, unlike the index, the key-value states can be fully offloaded to disk and do not need to be kept in memory. Once the document ID has been determined via the index, the corresponding key-value state can be simply loaded from disk. For a single sample, this corresponds to loading around 12.5MB of key-value states into memory.

\section{Discussion}

\paragraph{Further unification} To the best of our knowledge, \model{} is the first model to unify text embedding and generation, and thus all text-based language problems, into a single model at strong performance. However, many adjacent directions remain to be improved or unified. \textbf{(a) Multilinguality:} Our model is also capable of embedding and generation in non-English languages as seen in its TyDi QA performance (\autoref{tab:gen}). However, major performance gains on non-English tasks are likely possible through both data~\cite{muennighoff2023crosslingual,yong2023bloom1} and architecture changes~\cite{chen2024bge,feng2022languageagnostic,duquenne2023sonar} targeting multilinguality.  \textbf{(b) Multimodality:} Many embedding and generative problems are not purely text-based, such as joint embedding of images and text~\cite{radford2021learning}, generative image captioning~\cite{hossain2018comprehensive}, image-text pair classification~\cite{muennighoff2020vilio,kiela2021hateful} or speech versions of every text problem~\cite{kamath2019deep}. It remains to be explored whether they can be as easily unified as text embedding and generation in this work.

\paragraph{Why does \method{} work?} \method{} unifies embedding and generative tasks into a single model at no performance loss on either one, which may seem surprising. When the embedding dataset is MEDI2, we show that embedding performance even improves once the generative objective is added compared to an otherwise equivalent embedding-only model (\autoref{sec:ablations}). We think that our results confirm that generative language modeling and text embeddings are two sides of the same coin. Both tasks require a model to have a deep understanding of natural language and only differ in the way that understanding is expressed. Possibly, our unified model contains a small number of parameters that act as a switch to make the final representations either useful for mean pooling and subsequent embedding tasks or primed for the language modeling head and subsequent generative tasks. We are excited about future work exploring what is happening inside of \model{}. To support such research, we release all our work freely.

\paragraph{Optimizing RAG with \model{}} RAG and the caching variants we have presented in this work operate on a frozen language model. Meanwhile, there has been extensive work on optimizing a generative model specifically for interaction with a retrieval system~\cite{gao2024retrievalaugmented,zhu2024large,asai2023retrieval}. These works commonly optimize only the retriever~\cite{shi2023replug} or only the reader~\cite{borgeaud2022improving,yasunaga2023retrievalaugmented,asai2023selfrag,luo2023sail}. However, recent work has shown that jointly optimizing both models leads to the best performance~\cite{lin2023radit}. With its state-of-the-art retrieval and generative performance, \model{} can act as both the retriever and reader in a single model. Thus, optimizing either one also changes the parameters of the other. This has the potential to significantly simplify the joint optimization of the retriever and reader. For example, it may suffice to only use the next-token objective (\autoref{eq:gen}) to penalize the retriever for providing irrelevant context and at the same time the reader for poor use of the given context. This is in contrast to separate models and objective functions used in~\citet{lin2023radit}.

\section{Related Work}

The story of text embedding and text generation has been a story of unification.

\textbf{Embedding Models} used to focus on word representations~\cite{pennington2014glove,mikolov2013distributed} that struggled generalizing to entire sentences or passages~\cite{conneau2018senteval}. InferSent~\cite{conneau2018supervised}, SBERT~\cite{reimers2019sentencebert} and similar models~\cite{ni2021sentencet5,ni2021large} emerged that handle both the embedding of words and sentences at good quality by considering context when present. However, for strong performance, they require separate models for symmetric and asymmetric tasks~\cite{muennighoff2023mteb,muennighoff2022sgpt}. Symmetric embedding tasks are ones where the query and document are expected to come from the same distribution, such as STS. Meanwhile, for asymmetric tasks, they come from different distributions and as such could have very different sequence lengths like in retrieval. For example, the MTEB benchmark~\cite{muennighoff2023mteb} revealed that SentT5~\cite{ni2021sentencet5} only performs well at symmetric tasks, while GTR~\cite{ni2021large} only at asymmetric tasks despite both using T5~\cite{raffel2023exploring} as their base model. Recent embedding models have been able to unify symmetric and asymmetric tasks into a single model by differentiating them in the prompt~\cite{xiao2023cpack,wang2022text}. Further, including detailed instructions in the prompt has allowed unifying practically any embedding task into a single model~\cite{su2023embedder}.

\textbf{Generative Models} used to be tailored to a single task, such as translation~\cite{sutskever2014sequence} or question answering~\cite{yin2016neural}. \citet{mccann2018natural} cast multiple generative tasks as question answering to unify them within a single model, however, performance was still limited and it did not generalize to arbitrary tasks. Large-scale self-supervised pretraining has enabled the use of a single large language model (LLM) for practically any generative task~\cite{brown2020language,chowdhery2022palm,rae2022scaling,workshop2023bloom,scao2022language,groeneveld2024olmo,soldaini2024dolma,allal2023santacoder,li2023starcoder}. However, using an LLM without careful prompting often leads to poor performance~\cite{rubin2022learning,min2022rethinking}. Finetuning LLMs on instructions has emerged as a method to significantly ease the usage of the models to apply them to any generative task with strong results~\cite{wei2022finetuned,sanh2022multitask,min2022metaicl,wang2022supernaturalinstructions,mishra2022crosstask,muennighoff2023crosslingual,iyer2023optiml,üstün2024aya,singh2024aya,zhou2023lima}.

The two streams of embedding and generative models have respectively been unified into a single model that handles any task within its stream. Unifying the two streams into a single model that handles any task both for embedding and generation is the natural next step toward a general multi-task model. Besides generation, LLMs have also shown promise for text embeddings~\cite{muennighoff2022sgpt,neelakantan2022text,jiang2023scaling,li2023deelm,li2023angleoptimized}. SGPT~\cite{muennighoff2022sgpt} was an early work in that direction. SGPT only changes 0.01\% of the parameters of a large language model via BitFit~\cite{zaken2022bitfit} to adapt it to produce well-performing embeddings. Thus, one only needs to change this small amount of parameters to switch from one stream to the other. However, SGPT still required separate asymmetric and symmetric models and did not consider the full breadth of embedding tasks. \model{} addresses these deficiencies. \model{} does not require switching out biases, leverages instructions to handle asymmetric or symmetric use cases, and considers the full breadth of embedding and generative tasks.

\section{Conclusion}

We present \method{} to unify text embedding and generation, and thus all text-based language problems, into a single model: \model{}. \modelbase{} achieves state-of-the-art performance on the Massive Text Embedding Benchmark among open models, while at the same time beating all generative models up to its size. Notably, it matches the performance of otherwise equivalent embedding-only and generative-only variants allowing us to unify the two streams at no performance loss. By adding only 5B parameters at inference, \modelbig{} is the best open generative language model among the many we have tried including much larger models based on \llama{} 2 with 70B parameters. Unlike the other generative models, \modelbig{} also boasts very strong embedding performance thanks to the \method{} approach. Due to its unified capabilities, \model{} can be used as both the Bi-Encoder and Cross-Encoder in a reranking pipeline leading to performance improvements on 15 out of 16 retrieval datasets. Further, we conduct extensive ablations uncovering key insights for researchers of both embedding and generative models: causal language models for embedding should be finetuned with bidirectional attention and mean pooling, embedding performance of language models before finetuning is not predictive of embedding performance after finetuning, finetuning embedding models in BF16 mixed precision can match FP32 finetuning, pooling and similarity computations should be performed in FP32 precision during BF16 finetuning, generative models should be instruction-tuned with some form of token level loss, etc. Finally, we show that \method{} simplifies the field using the examples of reranking and RAG. For reranking, we are able to improve retrieval performance by around 10\% by reusing \model{} as reranker instead of having to rely on a separate model. For RAG, by unifying the retriever and reader into a single model, \model{} allows caching operations leading to inference speed-ups of >~60\% for long sequences at no performance loss with \method{} Doc Caching. We believe \method{} paves the way for a paradigm shift in language modeling, where embedding and generation seamlessly coexist in a single model. As such, we highlight the various limitations of this work and point the community to potential future research in \autoref{sec:limits}.

\begin{ack}

We thank everyone at Contextual AI, the University of Hong Kong, and Microsoft for their support. We are grateful to Hamish Ivison and Yizhong Wang for help with using the \tulu{} 2 repository. We thank Akari Asai, Alexander M. Rush, Brian Lester, Colin Raffel, Danqi Chen, Derek Tam, John X. Morris, Haokun Liu, Hong Liu, Mengzhou Xia, Michael Matena, Muqeeth Mohammed, Omar Khattab, Shayne Longpre, Tengyu Ma, Teven Le Scao, and Tianyu Gao for discussions and feedback.

\end{ack}

\newpage

\bibliography{custom}

\begin{thebibliography}{189}
\expandafter\ifx\csname natexlab\endcsname\relax\def\natexlab#1{#1}\fi

\bibitem[{Allal et~al.(2023)Allal, Li, Kocetkov, Mou, Akiki, Ferrandis, Muennighoff, Mishra, Gu, Dey et~al.}]{allal2023santacoder}
Loubna~Ben Allal, Raymond Li, Denis Kocetkov, Chenghao Mou, Christopher Akiki, Carlos~Munoz Ferrandis, Niklas Muennighoff, Mayank Mishra, Alex Gu, Manan Dey, et~al. 2023.
\newblock \href {http://arxiv.org/abs/2301.03988} {SantaCoder: don't reach for the stars!}

\bibitem[{Anil et~al.(2023)Anil, Dai, Firat, Johnson, Lepikhin, Passos, Shakeri, Taropa, Bailey, Chen et~al.}]{anil2023palm}
Rohan Anil, Andrew~M. Dai, Orhan Firat, Melvin Johnson, Dmitry Lepikhin, Alexandre Passos, Siamak Shakeri, Emanuel Taropa, Paige Bailey, Zhifeng Chen, et~al. 2023.
\newblock \href {http://arxiv.org/abs/2305.10403} {PaLM 2 Technical Report}.

\bibitem[{Arya et~al.(1998)Arya, Mount, Netanyahu, Silverman, and Wu}]{arya1998optimal}
Sunil Arya, David~M Mount, Nathan~S Netanyahu, Ruth Silverman, and Angela~Y Wu. 1998.
\newblock \href {https://graphics.stanford.edu/courses/cs468-06-fall/Papers/03%20AMNSW%20-%20JACM.pdf} {An optimal algorithm for approximate nearest neighbor searching fixed dimensions}.

\bibitem[{Asai et~al.(2023{\natexlab{a}})Asai, Min, Zhong, and Chen}]{asai2023retrieval}
Akari Asai, Sewon Min, Zexuan Zhong, and Danqi Chen. 2023{\natexlab{a}}.
\newblock \href {https://aclanthology.org/2023.acl-tutorials.6/} {Retrieval-based Language Models and Applications}.

\bibitem[{Asai et~al.(2022)Asai, Schick, Lewis, Chen, Izacard, Riedel, Hajishirzi, and tau Yih}]{asai2022taskaware}
Akari Asai, Timo Schick, Patrick Lewis, Xilun Chen, Gautier Izacard, Sebastian Riedel, Hannaneh Hajishirzi, and Wen tau Yih. 2022.
\newblock \href {http://arxiv.org/abs/2211.09260} {Task-aware Retrieval with Instructions}.

\bibitem[{Asai et~al.(2023{\natexlab{b}})Asai, Wu, Wang, Sil, and Hajishirzi}]{asai2023selfrag}
Akari Asai, Zeqiu Wu, Yizhong Wang, Avirup Sil, and Hannaneh Hajishirzi. 2023{\natexlab{b}}.
\newblock \href {http://arxiv.org/abs/2310.11511} {Self-RAG: Learning to Retrieve, Generate, and Critique through Self-Reflection}.

\bibitem[{Ba et~al.(2016)Ba, Kiros, and Hinton}]{ba2016layer}
Jimmy~Lei Ba, Jamie~Ryan Kiros, and Geoffrey~E. Hinton. 2016.
\newblock \href {http://arxiv.org/abs/1607.06450} {Layer Normalization}.

\bibitem[{Bajaj et~al.(2018)Bajaj, Campos, Craswell, Deng, Gao, Liu, Majumder, McNamara, Mitra, Nguyen, Rosenberg, Song, Stoica, Tiwary, and Wang}]{bajaj2018ms}
Payal Bajaj, Daniel Campos, Nick Craswell, Li~Deng, Jianfeng Gao, Xiaodong Liu, Rangan Majumder, Andrew McNamara, Bhaskar Mitra, Tri Nguyen, Mir Rosenberg, Xia Song, Alina Stoica, Saurabh Tiwary, and Tong Wang. 2018.
\newblock \href {http://arxiv.org/abs/1611.09268} {MS MARCO: A Human Generated MAchine Reading COmprehension Dataset}.

\bibitem[{Beltagy et~al.(2020)Beltagy, Peters, and Cohan}]{beltagy2020longformer}
Iz~Beltagy, Matthew~E. Peters, and Arman Cohan. 2020.
\newblock \href {http://arxiv.org/abs/2004.05150} {Longformer: The Long-Document Transformer}.

\bibitem[{Ben~Allal et~al.(2022)Ben~Allal, Muennighoff, Kumar~Umapathi, Lipkin, and von Werra}]{bigcode-evaluation-harness}
Loubna Ben~Allal, Niklas Muennighoff, Logesh Kumar~Umapathi, Ben Lipkin, and Leandro von Werra. 2022.
\newblock \href {https://github.com/bigcode-project/bigcode-evaluation-harness} {A framework for the evaluation of code generation models}.

\bibitem[{BigScience~Workshop et~al.(2023)BigScience~Workshop, Fan, Akiki, Pavlick, Ilić, Hesslow, Castagné, Luccioni, Yvon, Gallé, Tow, Rush, Biderman, Webson, Ammanamanchi, Wang, Sagot, Muennighoff et~al.}]{workshop2023bloom}
Teven Le~Scao BigScience~Workshop, Angela Fan, Christopher Akiki, Ellie Pavlick, Suzana Ilić, Daniel Hesslow, Roman Castagné, Alexandra~Sasha Luccioni, François Yvon, Matthias Gallé, Jonathan Tow, Alexander~M. Rush, Stella Biderman, Albert Webson, Pawan~Sasanka Ammanamanchi, Thomas Wang, Benoît Sagot, Niklas Muennighoff, et~al. 2023.
\newblock \href {http://arxiv.org/abs/2211.05100} {BLOOM: A 176B-Parameter Open-Access Multilingual Language Model}.

\bibitem[{Borgeaud et~al.(2022)Borgeaud, Mensch, Hoffmann, Cai, Rutherford, Millican, van~den Driessche, Lespiau, Damoc, Clark et~al.}]{borgeaud2022improving}
Sebastian Borgeaud, Arthur Mensch, Jordan Hoffmann, Trevor Cai, Eliza Rutherford, Katie Millican, George van~den Driessche, Jean-Baptiste Lespiau, Bogdan Damoc, Aidan Clark, et~al. 2022.
\newblock \href {http://arxiv.org/abs/2112.04426} {Improving language models by retrieving from trillions of tokens}.

\bibitem[{Brown et~al.(2020)Brown, Mann, Ryder, Subbiah, Kaplan, Dhariwal, Neelakantan, Shyam, Sastry, Askell et~al.}]{brown2020language}
Tom~B. Brown, Benjamin Mann, Nick Ryder, Melanie Subbiah, Jared Kaplan, Prafulla Dhariwal, Arvind Neelakantan, Pranav Shyam, Girish Sastry, Amanda Askell, et~al. 2020.
\newblock \href {http://arxiv.org/abs/2005.14165} {Language Models are Few-Shot Learners}.

\bibitem[{Cachola et~al.(2020)Cachola, Lo, Cohan, and Weld}]{cachola2020tldr}
Isabel Cachola, Kyle Lo, Arman Cohan, and Daniel~S. Weld. 2020.
\newblock \href {http://arxiv.org/abs/2004.15011} {TLDR: Extreme Summarization of Scientific Documents}.

\bibitem[{Chen et~al.(2024)Chen, Xiao, Zhang, Luo, Lian, and Liu}]{chen2024bge}
Jianlv Chen, Shitao Xiao, Peitian Zhang, Kun Luo, Defu Lian, and Zheng Liu. 2024.
\newblock \href {http://arxiv.org/abs/2402.03216} {BGE M3-Embedding: Multi-Lingual, Multi-Functionality, Multi-Granularity Text Embeddings Through Self-Knowledge Distillation}.

\bibitem[{Chen et~al.(2023)Chen, Zaharia, and Zou}]{chen2023chatgpts}
Lingjiao Chen, Matei Zaharia, and James Zou. 2023.
\newblock \href {http://arxiv.org/abs/2307.09009} {How is ChatGPT's behavior changing over time?}

\bibitem[{Chen et~al.(2021)Chen, Tworek, Jun, Yuan, de~Oliveira~Pinto, Kaplan, Edwards, Burda et~al.}]{chen2021evaluating}
Mark Chen, Jerry Tworek, Heewoo Jun, Qiming Yuan, Henrique~Ponde de~Oliveira~Pinto, Jared Kaplan, Harri Edwards, Yuri Burda, et~al. 2021.
\newblock \href {http://arxiv.org/abs/2107.03374} {Evaluating Large Language Models Trained on Code}.

\bibitem[{Chen et~al.(2020)Chen, Kornblith, Norouzi, and Hinton}]{chen2020simple}
Ting Chen, Simon Kornblith, Mohammad Norouzi, and Geoffrey Hinton. 2020.
\newblock \href {http://arxiv.org/abs/2002.05709} {A Simple Framework for Contrastive Learning of Visual Representations}.

\bibitem[{Chen et~al.(2015)Chen, Fang, Lin, Vedantam, Gupta, Dollar, and Zitnick}]{chen2015microsoft}
Xinlei Chen, Hao Fang, Tsung-Yi Lin, Ramakrishna Vedantam, Saurabh Gupta, Piotr Dollar, and C.~Lawrence Zitnick. 2015.
\newblock \href {http://arxiv.org/abs/1504.00325} {Microsoft COCO Captions: Data Collection and Evaluation Server}.

\bibitem[{Child et~al.(2019)Child, Gray, Radford, and Sutskever}]{child2019generating}
Rewon Child, Scott Gray, Alec Radford, and Ilya Sutskever. 2019.
\newblock \href {http://arxiv.org/abs/1904.10509} {Generating Long Sequences with Sparse Transformers}.

\bibitem[{Cho et~al.(2021)Cho, Lei, Tan, and Bansal}]{cho2021unifying}
Jaemin Cho, Jie Lei, Hao Tan, and Mohit Bansal. 2021.
\newblock \href {http://arxiv.org/abs/2102.02779} {Unifying Vision-and-Language Tasks via Text Generation}.

\bibitem[{Chowdhery et~al.(2022)Chowdhery, Narang, Devlin, Bosma, Mishra, Roberts, Barham, Chung, Sutton, Gehrmann et~al.}]{chowdhery2022palm}
Aakanksha Chowdhery, Sharan Narang, Jacob Devlin, Maarten Bosma, Gaurav Mishra, Adam Roberts, Paul Barham, Hyung~Won Chung, Charles Sutton, Sebastian Gehrmann, et~al. 2022.
\newblock \href {http://arxiv.org/abs/2204.02311} {PaLM: Scaling Language Modeling with Pathways}.

\bibitem[{Chung et~al.(2022)Chung, Hou, Longpre, Zoph, Tay, Fedus, Li, Wang et~al.}]{chung2022scaling}
Hyung~Won Chung, Le~Hou, Shayne Longpre, Barret Zoph, Yi~Tay, William Fedus, Yunxuan Li, Xuezhi Wang, et~al. 2022.
\newblock \href {http://arxiv.org/abs/2210.11416} {Scaling Instruction-Finetuned Language Models}.

\bibitem[{Clark et~al.(2020)Clark, Choi, Collins, Garrette, Kwiatkowski, Nikolaev, and Palomaki}]{clark2020tydi}
Jonathan~H. Clark, Eunsol Choi, Michael Collins, Dan Garrette, Tom Kwiatkowski, Vitaly Nikolaev, and Jennimaria Palomaki. 2020.
\newblock \href {http://arxiv.org/abs/2003.05002} {TyDi QA: A Benchmark for Information-Seeking Question Answering in Typologically Diverse Languages}.

\bibitem[{Cobbe et~al.(2021)Cobbe, Kosaraju, Bavarian, Chen, Jun, Kaiser, Plappert, Tworek, Hilton, Nakano, Hesse, and Schulman}]{cobbe2021training}
Karl Cobbe, Vineet Kosaraju, Mohammad Bavarian, Mark Chen, Heewoo Jun, Lukasz Kaiser, Matthias Plappert, Jerry Tworek, Jacob Hilton, Reiichiro Nakano, Christopher Hesse, and John Schulman. 2021.
\newblock \href {http://arxiv.org/abs/2110.14168} {Training Verifiers to Solve Math Word Problems}.

\bibitem[{Cohan et~al.(2020)Cohan, Feldman, Beltagy, Downey, and Weld}]{cohan2020specter}
Arman Cohan, Sergey Feldman, Iz~Beltagy, Doug Downey, and Daniel~S. Weld. 2020.
\newblock \href {http://arxiv.org/abs/2004.07180} {SPECTER: Document-level Representation Learning using Citation-informed Transformers}.

\bibitem[{Conneau and Kiela(2018)}]{conneau2018senteval}
Alexis Conneau and Douwe Kiela. 2018.
\newblock \href {http://arxiv.org/abs/1803.05449} {SentEval: An Evaluation Toolkit for Universal Sentence Representations}.

\bibitem[{Conneau et~al.(2018)Conneau, Kiela, Schwenk, Barrault, and Bordes}]{conneau2018supervised}
Alexis Conneau, Douwe Kiela, Holger Schwenk, Loic Barrault, and Antoine Bordes. 2018.
\newblock \href {http://arxiv.org/abs/1705.02364} {Supervised Learning of Universal Sentence Representations from Natural Language Inference Data}.

\bibitem[{Coster and Kauchak(2011)}]{Coster2011SimpleEW}
William Coster and David Kauchak. 2011.
\newblock \href {https://api.semanticscholar.org/CorpusID:9128245} {Simple English Wikipedia: A New Text Simplification Task}.

\bibitem[{Cui et~al.(2023)Cui, Yuan, Ding, Yao, Zhu, Ni, Xie, Liu, and Sun}]{cui2023ultrafeedback}
Ganqu Cui, Lifan Yuan, Ning Ding, Guanming Yao, Wei Zhu, Yuan Ni, Guotong Xie, Zhiyuan Liu, and Maosong Sun. 2023.
\newblock \href {http://arxiv.org/abs/2310.01377} {UltraFeedback: Boosting Language Models with High-quality Feedback}.

\bibitem[{Dao(2023)}]{dao2023flashattention}
Tri Dao. 2023.
\newblock \href {http://arxiv.org/abs/2307.08691} {FlashAttention-2: Faster Attention with Better Parallelism and Work Partitioning}.

\bibitem[{Dao et~al.(2022)Dao, Fu, Ermon, Rudra, and Ré}]{dao2022flashattention}
Tri Dao, Daniel~Y. Fu, Stefano Ermon, Atri Rudra, and Christopher Ré. 2022.
\newblock \href {http://arxiv.org/abs/2205.14135} {FlashAttention: Fast and Memory-Efficient Exact Attention with IO-Awareness}.

\bibitem[{DataCanary et~al.(2017)DataCanary, hilfialkaff, Lili~Jiang, Dandekar, and tomtung}]{quora-question-pairs}
DataCanary, hilfialkaff, Meg~Risdal Lili~Jiang, Nikhil Dandekar, and tomtung. 2017.
\newblock \href {https://kaggle.com/competitions/quora-question-pairs} {Quora Question Pairs}.

\bibitem[{Devlin et~al.(2019)Devlin, Chang, Lee, and Toutanova}]{devlin2019bert}
Jacob Devlin, Ming-Wei Chang, Kenton Lee, and Kristina Toutanova. 2019.
\newblock \href {http://arxiv.org/abs/1810.04805} {BERT: Pre-training of Deep Bidirectional Transformers for Language Understanding}.

\bibitem[{Dhole et~al.(2022)Dhole, Gangal, Gehrmann, Gupta, Li, Mahamood, Mahendiran, Mille, Shrivastava, Tan et~al.}]{dhole2022nlaugmenter}
Kaustubh~D. Dhole, Varun Gangal, Sebastian Gehrmann, Aadesh Gupta, Zhenhao Li, Saad Mahamood, Abinaya Mahendiran, Simon Mille, Ashish Shrivastava, Samson Tan, et~al. 2022.
\newblock \href {http://arxiv.org/abs/2112.02721} {NL-Augmenter: A Framework for Task-Sensitive Natural Language Augmentation}.

\bibitem[{Ding et~al.(2023)Ding, Chen, Xu, Qin, Zheng, Hu, Liu, Sun, and Zhou}]{ding2023enhancing}
Ning Ding, Yulin Chen, Bokai Xu, Yujia Qin, Zhi Zheng, Shengding Hu, Zhiyuan Liu, Maosong Sun, and Bowen Zhou. 2023.
\newblock \href {http://arxiv.org/abs/2305.14233} {Enhancing Chat Language Models by Scaling High-quality Instructional Conversations}.

\bibitem[{Douze et~al.(2024)Douze, Guzhva, Deng, Johnson, Szilvasy, Mazaré, Lomeli, Hosseini, and Jégou}]{douze2024faiss}
Matthijs Douze, Alexandr Guzhva, Chengqi Deng, Jeff Johnson, Gergely Szilvasy, Pierre-Emmanuel Mazaré, Maria Lomeli, Lucas Hosseini, and Hervé Jégou. 2024.
\newblock \href {http://arxiv.org/abs/2401.08281} {The Faiss library}.

\bibitem[{Du et~al.(2021)Du, Qian, Liu, Ding, Qiu, Yang, and Tang}]{du2021all}
Zhengxiao Du, Yujie Qian, Xiao Liu, Ming Ding, Jiezhong Qiu, Zhilin Yang, and Jie Tang. 2021.
\newblock \href {https://arxiv.org/abs/2103.10360v1} {All nlp tasks are generation tasks: A general pretraining framework}.

\bibitem[{Dubois et~al.(2024)Dubois, Galambosi, Liang, and Hashimoto}]{dubois2024length}
Yann Dubois, Bal{\'a}zs Galambosi, Percy Liang, and Tatsunori~B Hashimoto. 2024.
\newblock Length-Controlled AlpacaEval: A Simple Way to Debias Automatic Evaluators.
\newblock \emph{arXiv preprint arXiv:2404.04475}.

\bibitem[{Dubois et~al.(2023)Dubois, Li, Taori, Zhang, Gulrajani, Ba, Guestrin, Liang, and Hashimoto}]{dubois2023alpacafarm}
Yann Dubois, Xuechen Li, Rohan Taori, Tianyi Zhang, Ishaan Gulrajani, Jimmy Ba, Carlos Guestrin, Percy Liang, and Tatsunori~B. Hashimoto. 2023.
\newblock \href {http://arxiv.org/abs/2305.14387} {AlpacaFarm: A Simulation Framework for Methods that Learn from Human Feedback}.

\bibitem[{Dunn et~al.(2017)Dunn, Sagun, Higgins, G{"{u}}ney, Cirik, and Cho}]{DBLP:journals/corr/DunnSHGCC17}
Matthew Dunn, Levent Sagun, Mike Higgins, V.~Ugur G{"{u}}ney, Volkan Cirik, and Kyunghyun Cho. 2017.
\newblock \href {http://arxiv.org/abs/1704.05179} {SearchQA: {A} New Q{\&}A Dataset Augmented with Context from a Search Engine}.

\bibitem[{Duquenne et~al.(2023)Duquenne, Schwenk, and Sagot}]{duquenne2023sonar}
Paul-Ambroise Duquenne, Holger Schwenk, and Benoît Sagot. 2023.
\newblock \href {http://arxiv.org/abs/2308.11466} {SONAR: Sentence-Level Multimodal and Language-Agnostic Representations}.

\bibitem[{ElSahar et~al.(2018)ElSahar, Vougiouklis, Remaci, Gravier, Hare, Laforest, and Simperl}]{ElSahar2018TRExAL}
Hady ElSahar, Pavlos Vougiouklis, Arslen Remaci, Christophe Gravier, Jonathon~S. Hare, Fr{\'e}d{\'e}rique Laforest, and Elena Paslaru~Bontas Simperl. 2018.
\newblock \href {https://api.semanticscholar.org/CorpusID:4612975} {T-REx: A Large Scale Alignment of Natural Language with Knowledge Base Triples}.

\bibitem[{Ethayarajh et~al.(2024)Ethayarajh, Xu, Muennighoff, Jurafsky, and Kiela}]{ethayarajh2024kto}
Kawin Ethayarajh, Winnie Xu, Niklas Muennighoff, Dan Jurafsky, and Douwe Kiela. 2024.
\newblock \href {http://arxiv.org/abs/2402.01306} {KTO: Model Alignment as Prospect Theoretic Optimization}.

\bibitem[{Fabbri et~al.(2021)Fabbri, Kryściński, McCann, Xiong, Socher, and Radev}]{fabbri2021summeval}
Alexander~R. Fabbri, Wojciech Kryściński, Bryan McCann, Caiming Xiong, Richard Socher, and Dragomir Radev. 2021.
\newblock \href {http://arxiv.org/abs/2007.12626} {SummEval: Re-evaluating Summarization Evaluation}.

\bibitem[{Fader et~al.(2014)Fader, Zettlemoyer, and Etzioni}]{Fader2014OpenQA}
Anthony Fader, Luke Zettlemoyer, and Oren Etzioni. 2014.
\newblock \href {https://api.semanticscholar.org/CorpusID:207214527} {Open question answering over curated and extracted knowledge bases}.

\bibitem[{Fan et~al.(2019)Fan, Jernite, Perez, Grangier, Weston, and Auli}]{fan2019eli5}
Angela Fan, Yacine Jernite, Ethan Perez, David Grangier, Jason Weston, and Michael Auli. 2019.
\newblock \href {http://arxiv.org/abs/1907.09190} {ELI5: Long Form Question Answering}.

\bibitem[{Feng et~al.(2022)Feng, Yang, Cer, Arivazhagan, and Wang}]{feng2022languageagnostic}
Fangxiaoyu Feng, Yinfei Yang, Daniel Cer, Naveen Arivazhagan, and Wei Wang. 2022.
\newblock \href {http://arxiv.org/abs/2007.01852} {Language-agnostic BERT Sentence Embedding}.

\bibitem[{Filippova and Altun(2013)}]{Filippova2013OvercomingTL}
Katja Filippova and Yasemin Altun. 2013.
\newblock \href {https://api.semanticscholar.org/CorpusID:9751546} {Overcoming the Lack of Parallel Data in Sentence Compression}.

\bibitem[{Gao et~al.(2021{\natexlab{a}})Gao, Tow, Biderman, Black, DiPofi, Foster, Golding, Hsu, McDonell, Muennighoff, Phang, Reynolds, Tang, Thite, Wang, Wang, and Zou}]{eval-harness}
Leo Gao, Jonathan Tow, Stella Biderman, Sid Black, Anthony DiPofi, Charles Foster, Laurence Golding, Jeffrey Hsu, Kyle McDonell, Niklas Muennighoff, Jason Phang, Laria Reynolds, Eric Tang, Anish Thite, Ben Wang, Kevin Wang, and Andy Zou. 2021{\natexlab{a}}.
\newblock \href {https://doi.org/10.5281/zenodo.5371628} {A framework for few-shot language model evaluation}.

\bibitem[{Gao et~al.(2021{\natexlab{b}})Gao, Zhang, Han, and Callan}]{gao2021scaling}
Luyu Gao, Yunyi Zhang, Jiawei Han, and Jamie Callan. 2021{\natexlab{b}}.
\newblock \href {http://arxiv.org/abs/2101.06983} {Scaling Deep Contrastive Learning Batch Size under Memory Limited Setup}.

\bibitem[{Gao et~al.(2022)Gao, Yao, and Chen}]{gao2022simcse}
Tianyu Gao, Xingcheng Yao, and Danqi Chen. 2022.
\newblock \href {http://arxiv.org/abs/2104.08821} {SimCSE: Simple Contrastive Learning of Sentence Embeddings}.

\bibitem[{Gao et~al.(2024)Gao, Xiong, Gao, Jia, Pan, Bi, Dai, Sun, Guo, Wang, and Wang}]{gao2024retrievalaugmented}
Yunfan Gao, Yun Xiong, Xinyu Gao, Kangxiang Jia, Jinliu Pan, Yuxi Bi, Yi~Dai, Jiawei Sun, Qianyu Guo, Meng Wang, and Haofen Wang. 2024.
\newblock \href {http://arxiv.org/abs/2312.10997} {Retrieval-Augmented Generation for Large Language Models: A Survey}.

\bibitem[{Graff et~al.(2003)Graff, Kong, Chen, and Maeda}]{graff2003english}
David Graff, Junbo Kong, Ke~Chen, and Kazuaki Maeda. 2003.
\newblock \href {https://catalog.ldc.upenn.edu/LDC2011T07} {English gigaword}.

\bibitem[{Groeneveld et~al.(2024)Groeneveld, Beltagy, Walsh, Bhagia, Kinney, Tafjord, Jha, Ivison, Magnusson, Wang, Arora, Atkinson, Authur, Chandu, Cohan, Dumas, Elazar, Gu, Hessel, Khot, Merrill, Morrison, Muennighoff, Naik, Nam, Peters, Pyatkin, Ravichander, Schwenk, Shah, Smith, Strubell, Subramani, Wortsman, Dasigi, Lambert, Richardson, Zettlemoyer, Dodge, Lo, Soldaini, Smith, and Hajishirzi}]{groeneveld2024olmo}
Dirk Groeneveld, Iz~Beltagy, Pete Walsh, Akshita Bhagia, Rodney Kinney, Oyvind Tafjord, Ananya~Harsh Jha, Hamish Ivison, Ian Magnusson, Yizhong Wang, Shane Arora, David Atkinson, Russell Authur, Khyathi~Raghavi Chandu, Arman Cohan, Jennifer Dumas, Yanai Elazar, Yuling Gu, Jack Hessel, Tushar Khot, William Merrill, Jacob Morrison, Niklas Muennighoff, Aakanksha Naik, Crystal Nam, Matthew~E. Peters, Valentina Pyatkin, Abhilasha Ravichander, Dustin Schwenk, Saurabh Shah, Will Smith, Emma Strubell, Nishant Subramani, Mitchell Wortsman, Pradeep Dasigi, Nathan Lambert, Kyle Richardson, Luke Zettlemoyer, Jesse Dodge, Kyle Lo, Luca Soldaini, Noah~A. Smith, and Hannaneh Hajishirzi. 2024.
\newblock \href {http://arxiv.org/abs/2402.00838} {OLMo: Accelerating the Science of Language Models}.

\bibitem[{Gupta et~al.(2019)Gupta, Kulkarni, Chanda, Rayasam, and Lipton}]{gupta2019amazonqa}
Mansi Gupta, Nitish Kulkarni, Raghuveer Chanda, Anirudha Rayasam, and Zachary~C Lipton. 2019.
\newblock \href {http://arxiv.org/abs/1908.04364} {AmazonQA: A Review-Based Question Answering Task}.

\bibitem[{Guu et~al.(2020)Guu, Lee, Tung, Pasupat, and Chang}]{guu2020realm}
Kelvin Guu, Kenton Lee, Zora Tung, Panupong Pasupat, and Ming-Wei Chang. 2020.
\newblock \href {http://arxiv.org/abs/2002.08909} {REALM: Retrieval-Augmented Language Model Pre-Training}.

\bibitem[{Günther et~al.(2023)Günther, Milliken, Geuter, Mastrapas, Wang, and Xiao}]{günther2023jina}
Michael Günther, Louis Milliken, Jonathan Geuter, Georgios Mastrapas, Bo~Wang, and Han Xiao. 2023.
\newblock \href {http://arxiv.org/abs/2307.11224} {Jina Embeddings: A Novel Set of High-Performance Sentence Embedding Models}.

\bibitem[{Günther et~al.(2024)Günther, Ong, Mohr, Abdessalem, Abel, Akram, Guzman, Mastrapas, Sturua, Wang, Werk, Wang, and Xiao}]{günther2024jina}
Michael Günther, Jackmin Ong, Isabelle Mohr, Alaeddine Abdessalem, Tanguy Abel, Mohammad~Kalim Akram, Susana Guzman, Georgios Mastrapas, Saba Sturua, Bo~Wang, Maximilian Werk, Nan Wang, and Han Xiao. 2024.
\newblock \href {http://arxiv.org/abs/2310.19923} {Jina Embeddings 2: 8192-Token General-Purpose Text Embeddings for Long Documents}.

\bibitem[{Hamborg et~al.(2017)Hamborg, Meuschke, Breitinger, and Gipp}]{Hamborg2017newspleaseA}
Felix Hamborg, Norman Meuschke, Corinna Breitinger, and Bela Gipp. 2017.
\newblock \href {https://api.semanticscholar.org/CorpusID:5830937} {news-please - A Generic News Crawler and Extractor}.

\bibitem[{Hendrycks et~al.(2022)Hendrycks, Carlini, Schulman, and Steinhardt}]{hendrycks2022unsolved}
Dan Hendrycks, Nicholas Carlini, John Schulman, and Jacob Steinhardt. 2022.
\newblock \href {http://arxiv.org/abs/2109.13916} {Unsolved Problems in ML Safety}.

\bibitem[{Hidey and McKeown(2016)}]{hidey-mckeown-2016-identifying}
Christopher Hidey and Kathy McKeown. 2016.
\newblock \href {https://doi.org/10.18653/v1/P16-1135} {Identifying Causal Relations Using Parallel {W}ikipedia Articles}.

\bibitem[{Hossain et~al.(2018)Hossain, Sohel, Shiratuddin, and Laga}]{hossain2018comprehensive}
Md.~Zakir Hossain, Ferdous Sohel, Mohd~Fairuz Shiratuddin, and Hamid Laga. 2018.
\newblock \href {http://arxiv.org/abs/1810.04020} {A Comprehensive Survey of Deep Learning for Image Captioning}.

\bibitem[{Huang et~al.(2020)Huang, Sharma, Sun, Xia, Zhang, Pronin, Padmanabhan, Ottaviano, and Yang}]{huang2020embedding}
Jui-Ting Huang, Ashish Sharma, Shuying Sun, Li~Xia, David Zhang, Philip Pronin, Janani Padmanabhan, Giuseppe Ottaviano, and Linjun Yang. 2020.
\newblock \href {https://arxiv.org/abs/2006.11632} {Embedding-based retrieval in facebook search}.

\bibitem[{Ivison et~al.(2023)Ivison, Wang, Pyatkin, Lambert, Peters, Dasigi, Jang, Wadden, Smith, Beltagy, and Hajishirzi}]{ivison2023camels}
Hamish Ivison, Yizhong Wang, Valentina Pyatkin, Nathan Lambert, Matthew Peters, Pradeep Dasigi, Joel Jang, David Wadden, Noah~A. Smith, Iz~Beltagy, and Hannaneh Hajishirzi. 2023.
\newblock \href {http://arxiv.org/abs/2311.10702} {Camels in a Changing Climate: Enhancing LM Adaptation with Tulu 2}.

\bibitem[{Iyer et~al.(2023)Iyer, Lin, Pasunuru, Mihaylov, Simig, Yu, Shuster, Wang, Liu, Koura, Li, O'Horo, Pereyra, Wang, Dewan, Celikyilmaz, Zettlemoyer, and Stoyanov}]{iyer2023optiml}
Srinivasan Iyer, Xi~Victoria Lin, Ramakanth Pasunuru, Todor Mihaylov, Daniel Simig, Ping Yu, Kurt Shuster, Tianlu Wang, Qing Liu, Punit~Singh Koura, Xian Li, Brian O'Horo, Gabriel Pereyra, Jeff Wang, Christopher Dewan, Asli Celikyilmaz, Luke Zettlemoyer, and Ves Stoyanov. 2023.
\newblock \href {http://arxiv.org/abs/2212.12017} {OPT-IML: Scaling Language Model Instruction Meta Learning through the Lens of Generalization}.

\bibitem[{Izacard et~al.(2022)Izacard, Lewis, Lomeli, Hosseini, Petroni, Schick, Dwivedi-Yu, Joulin, Riedel, and Grave}]{izacard2022atlas}
Gautier Izacard, Patrick Lewis, Maria Lomeli, Lucas Hosseini, Fabio Petroni, Timo Schick, Jane Dwivedi-Yu, Armand Joulin, Sebastian Riedel, and Edouard Grave. 2022.
\newblock \href {http://arxiv.org/abs/2208.03299} {Atlas: Few-shot Learning with Retrieval Augmented Language Models}.

\bibitem[{Jaegle et~al.(2021)Jaegle, Gimeno, Brock, Zisserman, Vinyals, and Carreira}]{jaegle2021perceiver}
Andrew Jaegle, Felix Gimeno, Andrew Brock, Andrew Zisserman, Oriol Vinyals, and Joao Carreira. 2021.
\newblock \href {http://arxiv.org/abs/2103.03206} {Perceiver: General Perception with Iterative Attention}.

\bibitem[{Jiang et~al.(2023{\natexlab{a}})Jiang, Sablayrolles, Mensch, Bamford, Chaplot, de~las Casas, Bressand, Lengyel, Lample, Saulnier, Lavaud, Lachaux, Stock, Scao, Lavril, Wang, Lacroix, and Sayed}]{jiang2023mistral}
Albert~Q. Jiang, Alexandre Sablayrolles, Arthur Mensch, Chris Bamford, Devendra~Singh Chaplot, Diego de~las Casas, Florian Bressand, Gianna Lengyel, Guillaume Lample, Lucile Saulnier, Lélio~Renard Lavaud, Marie-Anne Lachaux, Pierre Stock, Teven~Le Scao, Thibaut Lavril, Thomas Wang, Timothée Lacroix, and William~El Sayed. 2023{\natexlab{a}}.
\newblock \href {http://arxiv.org/abs/2310.06825} {Mistral 7B}.

\bibitem[{Jiang et~al.(2024)Jiang, Sablayrolles, Roux, Mensch, Savary, Bamford, Chaplot, de~las Casas, Hanna, Bressand, Lengyel, Bour, Lample, Lavaud, Saulnier, Lachaux, Stock, Subramanian, Yang, Antoniak, Scao, Gervet, Lavril, Wang, Lacroix, and Sayed}]{jiang2024mixtral}
Albert~Q. Jiang, Alexandre Sablayrolles, Antoine Roux, Arthur Mensch, Blanche Savary, Chris Bamford, Devendra~Singh Chaplot, Diego de~las Casas, Emma~Bou Hanna, Florian Bressand, Gianna Lengyel, Guillaume Bour, Guillaume Lample, Lélio~Renard Lavaud, Lucile Saulnier, Marie-Anne Lachaux, Pierre Stock, Sandeep Subramanian, Sophia Yang, Szymon Antoniak, Teven~Le Scao, Théophile Gervet, Thibaut Lavril, Thomas Wang, Timothée Lacroix, and William~El Sayed. 2024.
\newblock \href {http://arxiv.org/abs/2401.04088} {Mixtral of Experts}.

\bibitem[{Jiang et~al.(2023{\natexlab{b}})Jiang, Huang, Luan, Wang, and Zhuang}]{jiang2023scaling}
Ting Jiang, Shaohan Huang, Zhongzhi Luan, Deqing Wang, and Fuzhen Zhuang. 2023{\natexlab{b}}.
\newblock \href {http://arxiv.org/abs/2307.16645} {Scaling Sentence Embeddings with Large Language Models}.

\bibitem[{Jin et~al.(2019)Jin, Dhingra, Liu, Cohen, and Lu}]{jin2019pubmedqa}
Qiao Jin, Bhuwan Dhingra, Zhengping Liu, William~W. Cohen, and Xinghua Lu. 2019.
\newblock \href {http://arxiv.org/abs/1909.06146} {PubMedQA: A Dataset for Biomedical Research Question Answering}.

\bibitem[{Johnson et~al.(2017)Johnson, Douze, and Jégou}]{johnson2017billionscale}
Jeff Johnson, Matthijs Douze, and Hervé Jégou. 2017.
\newblock \href {http://arxiv.org/abs/1702.08734} {Billion-scale similarity search with GPUs}.

\bibitem[{Joshi et~al.(2017)Joshi, Choi, Weld, and Zettlemoyer}]{joshi2017triviaqalargescaledistantly}
Mandar Joshi, Eunsol Choi, Daniel~S. Weld, and Luke Zettlemoyer. 2017.
\newblock \href {http://arxiv.org/abs/1705.03551} {TriviaQA: A Large Scale Distantly Supervised Challenge Dataset for Reading Comprehension}.

\bibitem[{Kaiser et~al.(2017)Kaiser, Gomez, Shazeer, Vaswani, Parmar, Jones, and Uszkoreit}]{kaiser2017model}
Lukasz Kaiser, Aidan~N. Gomez, Noam Shazeer, Ashish Vaswani, Niki Parmar, Llion Jones, and Jakob Uszkoreit. 2017.
\newblock \href {http://arxiv.org/abs/1706.05137} {One Model To Learn Them All}.

\bibitem[{Kamath et~al.(2019)Kamath, Liu, and Whitaker}]{kamath2019deep}
Uday Kamath, John Liu, and James Whitaker. 2019.
\newblock \href {https://link.springer.com/book/10.1007/978-3-030-14596-5} {Deep learning for NLP and speech recognition}.

\bibitem[{Karpukhin et~al.(2020)Karpukhin, Oğuz, Min, Lewis, Wu, Edunov, Chen, and tau Yih}]{karpukhin2020dense}
Vladimir Karpukhin, Barlas Oğuz, Sewon Min, Patrick Lewis, Ledell Wu, Sergey Edunov, Danqi Chen, and Wen tau Yih. 2020.
\newblock \href {http://arxiv.org/abs/2004.04906} {Dense Passage Retrieval for Open-Domain Question Answering}.

\bibitem[{Keung et~al.(2020)Keung, Lu, Szarvas, and Smith}]{keung2020multilingual}
Phillip Keung, Yichao Lu, György Szarvas, and Noah~A. Smith. 2020.
\newblock \href {http://arxiv.org/abs/2010.02573} {The Multilingual Amazon Reviews Corpus}.

\bibitem[{Khashabi et~al.(2021)Khashabi, Ng, Khot, Sabharwal, Hajishirzi, and Callison-Burch}]{khashabi2021gooaq}
Daniel Khashabi, Amos Ng, Tushar Khot, Ashish Sabharwal, Hannaneh Hajishirzi, and Chris Callison-Burch. 2021.
\newblock \href {http://arxiv.org/abs/2104.08727} {GooAQ: Open Question Answering with Diverse Answer Types}.

\bibitem[{Kiela et~al.(2021)Kiela, Firooz, Mohan, Goswami, Singh, Fitzpatrick, Bull, Lipstein, Nelli, Zhu et~al.}]{kiela2021hateful}
Douwe Kiela, Hamed Firooz, Aravind Mohan, Vedanuj Goswami, Amanpreet Singh, Casey~A Fitzpatrick, Peter Bull, Greg Lipstein, Tony Nelli, Ron Zhu, et~al. 2021.
\newblock \href {https://proceedings.mlr.press/v133/kiela21a.html} {The hateful memes challenge: Competition report}.

\bibitem[{Kingma and Ba(2017)}]{kingma2017adam}
Diederik~P. Kingma and Jimmy Ba. 2017.
\newblock \href {http://arxiv.org/abs/1412.6980} {Adam: A Method for Stochastic Optimization}.

\bibitem[{Koupaee and Wang(2018)}]{koupaee2018wikihow}
Mahnaz Koupaee and William~Yang Wang. 2018.
\newblock \href {http://arxiv.org/abs/1810.09305} {WikiHow: A Large Scale Text Summarization Dataset}.

\bibitem[{Kwiatkowski et~al.(2019)Kwiatkowski, Palomaki, Redfield, Collins, Parikh, Alberti, Epstein, Polosukhin, Devlin, Lee et~al.}]{kwiatkowski2019natural}
Tom Kwiatkowski, Jennimaria Palomaki, Olivia Redfield, Michael Collins, Ankur Parikh, Chris Alberti, Danielle Epstein, Illia Polosukhin, Jacob Devlin, Kenton Lee, et~al. 2019.
\newblock \href {https://aclanthology.org/Q19-1026/} {Natural questions: a benchmark for question answering research}.

\bibitem[{Köpf et~al.(2023)Köpf, Kilcher, von Rütte, Anagnostidis, Tam, Stevens, Barhoum, Duc, Stanley, Nagyfi, ES, Suri, Glushkov, Dantuluri, Maguire, Schuhmann, Nguyen, and Mattick}]{köpf2023openassistant}
Andreas Köpf, Yannic Kilcher, Dimitri von Rütte, Sotiris Anagnostidis, Zhi-Rui Tam, Keith Stevens, Abdullah Barhoum, Nguyen~Minh Duc, Oliver Stanley, Richárd Nagyfi, Shahul ES, Sameer Suri, David Glushkov, Arnav Dantuluri, Andrew Maguire, Christoph Schuhmann, Huu Nguyen, and Alexander Mattick. 2023.
\newblock \href {http://arxiv.org/abs/2304.07327} {OpenAssistant Conversations -- Democratizing Large Language Model Alignment}.

\bibitem[{Levy et~al.(2017)Levy, Seo, Choi, and Zettlemoyer}]{levy2017zeroshot}
Omer Levy, Minjoon Seo, Eunsol Choi, and Luke Zettlemoyer. 2017.
\newblock \href {http://arxiv.org/abs/1706.04115} {Zero-Shot Relation Extraction via Reading Comprehension}.

\bibitem[{Lewis et~al.(2021{\natexlab{a}})Lewis, Perez, Piktus, Petroni, Karpukhin, Goyal, Küttler, Lewis, tau Yih, Rocktäschel, Riedel, and Kiela}]{lewis2021retrievalaugmented}
Patrick Lewis, Ethan Perez, Aleksandra Piktus, Fabio Petroni, Vladimir Karpukhin, Naman Goyal, Heinrich Küttler, Mike Lewis, Wen tau Yih, Tim Rocktäschel, Sebastian Riedel, and Douwe Kiela. 2021{\natexlab{a}}.
\newblock \href {http://arxiv.org/abs/2005.11401} {Retrieval-Augmented Generation for Knowledge-Intensive NLP Tasks}.

\bibitem[{Lewis et~al.(2021{\natexlab{b}})Lewis, Wu, Liu, Minervini, Küttler, Piktus, Stenetorp, and Riedel}]{lewis2021paq}
Patrick Lewis, Yuxiang Wu, Linqing Liu, Pasquale Minervini, Heinrich Küttler, Aleksandra Piktus, Pontus Stenetorp, and Sebastian Riedel. 2021{\natexlab{b}}.
\newblock \href {http://arxiv.org/abs/2102.07033} {PAQ: 65 Million Probably-Asked Questions and What You Can Do With Them}.

\bibitem[{Li et~al.(2023{\natexlab{a}})Li, Allal, Zi, Muennighoff, Kocetkov, Mou, Marone, Akiki, Li, Chim et~al.}]{li2023starcoder}
Raymond Li, Loubna~Ben Allal, Yangtian Zi, Niklas Muennighoff, Denis Kocetkov, Chenghao Mou, Marc Marone, Christopher Akiki, Jia Li, Jenny Chim, et~al. 2023{\natexlab{a}}.
\newblock \href {http://arxiv.org/abs/2305.06161} {StarCoder: may the source be with you!}

\bibitem[{Li and Li(2023{\natexlab{a}})}]{li2023angleoptimized}
Xianming Li and Jing Li. 2023{\natexlab{a}}.
\newblock \href {http://arxiv.org/abs/2309.12871} {AnglE-optimized Text Embeddings}.

\bibitem[{Li and Li(2023{\natexlab{b}})}]{li2023deelm}
Xianming Li and Jing Li. 2023{\natexlab{b}}.
\newblock \href {http://arxiv.org/abs/2311.05296} {DeeLM: Dependency-enhanced Large Language Model for Sentence Embeddings}.

\bibitem[{Li et~al.(2023{\natexlab{b}})Li, Zhang, Dubois, Taori, Gulrajani, Guestrin, Liang, and Hashimoto}]{alpaca_eval}
Xuechen Li, Tianyi Zhang, Yann Dubois, Rohan Taori, Ishaan Gulrajani, Carlos Guestrin, Percy Liang, and Tatsunori~B. Hashimoto. 2023{\natexlab{b}}.
\newblock \href {https://github.com/tatsu-lab/alpaca_eval} {AlpacaEval: An Automatic Evaluator of Instruction-following Models}.

\bibitem[{Lin et~al.(2023)Lin, Chen, Chen, Shi, Lomeli, James, Rodriguez, Kahn, Szilvasy, Lewis, Zettlemoyer, and Yih}]{lin2023radit}
Xi~Victoria Lin, Xilun Chen, Mingda Chen, Weijia Shi, Maria Lomeli, Rich James, Pedro Rodriguez, Jacob Kahn, Gergely Szilvasy, Mike Lewis, Luke Zettlemoyer, and Scott Yih. 2023.
\newblock \href {http://arxiv.org/abs/2310.01352} {RA-DIT: Retrieval-Augmented Dual Instruction Tuning}.

\bibitem[{Lo et~al.(2020)Lo, Wang, Neumann, Kinney, and Weld}]{lo2020s2orc}
Kyle Lo, Lucy~Lu Wang, Mark Neumann, Rodney Kinney, and Dan~S. Weld. 2020.
\newblock \href {http://arxiv.org/abs/1911.02782} {S2ORC: The Semantic Scholar Open Research Corpus}.

\bibitem[{Longpre et~al.(2023)Longpre, Mahari, Chen, Obeng-Marnu, Sileo, Brannon, Muennighoff, Khazam, Kabbara, Perisetla, Wu, Shippole, Bollacker, Wu, Villa, Pentland, Roy, and Hooker}]{longpre2023data}
Shayne Longpre, Robert Mahari, Anthony Chen, Naana Obeng-Marnu, Damien Sileo, William Brannon, Niklas Muennighoff, Nathan Khazam, Jad Kabbara, Kartik Perisetla, Xinyi Wu, Enrico Shippole, Kurt Bollacker, Tongshuang Wu, Luis Villa, Sandy Pentland, Deb Roy, and Sara Hooker. 2023.
\newblock \href {http://arxiv.org/abs/2310.16787} {The Data Provenance Initiative: A Large Scale Audit of Dataset Licensing \& Attribution in AI}.

\bibitem[{Luo et~al.(2023)Luo, Chuang, Gong, Zhang, Kim, Wu, Fox, Meng, and Glass}]{luo2023sail}
Hongyin Luo, Yung-Sung Chuang, Yuan Gong, Tianhua Zhang, Yoon Kim, Xixin Wu, Danny Fox, Helen Meng, and James Glass. 2023.
\newblock \href {http://arxiv.org/abs/2305.15225} {SAIL: Search-Augmented Instruction Learning}.

\bibitem[{Luukkonen et~al.(2023)Luukkonen, Komulainen, Luoma, Eskelinen, Kanerva, Kupari, Ginter, Laippala, Muennighoff, Piktus, Wang, Tazi, Scao, Wolf, Suominen, Sairanen, Merioksa, Heinonen, Vahtola, Antao, and Pyysalo}]{luukkonen2023fingpt}
Risto Luukkonen, Ville Komulainen, Jouni Luoma, Anni Eskelinen, Jenna Kanerva, Hanna-Mari Kupari, Filip Ginter, Veronika Laippala, Niklas Muennighoff, Aleksandra Piktus, Thomas Wang, Nouamane Tazi, Teven~Le Scao, Thomas Wolf, Osma Suominen, Samuli Sairanen, Mikko Merioksa, Jyrki Heinonen, Aija Vahtola, Samuel Antao, and Sampo Pyysalo. 2023.
\newblock \href {http://arxiv.org/abs/2311.05640} {FinGPT: Large Generative Models for a Small Language}.

\bibitem[{Ma et~al.(2023{\natexlab{a}})Ma, Wang, Yang, Wei, and Lin}]{ma2023finetuning}
Xueguang Ma, Liang Wang, Nan Yang, Furu Wei, and Jimmy Lin. 2023{\natexlab{a}}.
\newblock \href {http://arxiv.org/abs/2310.08319} {Fine-Tuning LLaMA for Multi-Stage Text Retrieval}.

\bibitem[{Ma et~al.(2023{\natexlab{b}})Ma, Zhang, Pradeep, and Lin}]{ma2023zeroshot}
Xueguang Ma, Xinyu Zhang, Ronak Pradeep, and Jimmy Lin. 2023{\natexlab{b}}.
\newblock \href {http://arxiv.org/abs/2305.02156} {Zero-Shot Listwise Document Reranking with a Large Language Model}.

\bibitem[{McCann et~al.(2018)McCann, Keskar, Xiong, and Socher}]{mccann2018natural}
Bryan McCann, Nitish~Shirish Keskar, Caiming Xiong, and Richard Socher. 2018.
\newblock \href {http://arxiv.org/abs/1806.08730} {The Natural Language Decathlon: Multitask Learning as Question Answering}.

\bibitem[{Mikolov et~al.(2013)Mikolov, Sutskever, Chen, Corrado, and Dean}]{mikolov2013distributed}
Tomas Mikolov, Ilya Sutskever, Kai Chen, Greg Corrado, and Jeffrey Dean. 2013.
\newblock \href {http://arxiv.org/abs/1310.4546} {Distributed Representations of Words and Phrases and their Compositionality}.

\bibitem[{Min et~al.(2022{\natexlab{a}})Min, Lewis, Zettlemoyer, and Hajishirzi}]{min2022metaicl}
Sewon Min, Mike Lewis, Luke Zettlemoyer, and Hannaneh Hajishirzi. 2022{\natexlab{a}}.
\newblock \href {http://arxiv.org/abs/2110.15943} {MetaICL: Learning to Learn In Context}.

\bibitem[{Min et~al.(2022{\natexlab{b}})Min, Lyu, Holtzman, Artetxe, Lewis, Hajishirzi, and Zettlemoyer}]{min2022rethinking}
Sewon Min, Xinxi Lyu, Ari Holtzman, Mikel Artetxe, Mike Lewis, Hannaneh Hajishirzi, and Luke Zettlemoyer. 2022{\natexlab{b}}.
\newblock \href {http://arxiv.org/abs/2202.12837} {Rethinking the Role of Demonstrations: What Makes In-Context Learning Work?}

\bibitem[{Mishra et~al.(2022)Mishra, Khashabi, Baral, and Hajishirzi}]{mishra2022crosstask}
Swaroop Mishra, Daniel Khashabi, Chitta Baral, and Hannaneh Hajishirzi. 2022.
\newblock \href {http://arxiv.org/abs/2104.08773} {Cross-Task Generalization via Natural Language Crowdsourcing Instructions}.

\bibitem[{Morris et~al.(2023)Morris, Kuleshov, Shmatikov, and Rush}]{morris2023text}
John~X. Morris, Volodymyr Kuleshov, Vitaly Shmatikov, and Alexander~M. Rush. 2023.
\newblock \href {http://arxiv.org/abs/2310.06816} {Text Embeddings Reveal (Almost) As Much As Text}.

\bibitem[{Muennighoff(2020)}]{muennighoff2020vilio}
Niklas Muennighoff. 2020.
\newblock \href {http://arxiv.org/abs/2012.07788} {Vilio: State-of-the-art Visio-Linguistic Models applied to Hateful Memes}.

\bibitem[{Muennighoff(2022)}]{muennighoff2022sgpt}
Niklas Muennighoff. 2022.
\newblock \href {http://arxiv.org/abs/2202.08904} {SGPT: GPT Sentence Embeddings for Semantic Search}.

\bibitem[{Muennighoff et~al.(2023{\natexlab{a}})Muennighoff, Liu, Zebaze, Zheng, Hui, Zhuo, Singh, Tang, von Werra, and Longpre}]{muennighoff2023octopack}
Niklas Muennighoff, Qian Liu, Armel Zebaze, Qinkai Zheng, Binyuan Hui, Terry~Yue Zhuo, Swayam Singh, Xiangru Tang, Leandro von Werra, and Shayne Longpre. 2023{\natexlab{a}}.
\newblock \href {http://arxiv.org/abs/2308.07124} {OctoPack: Instruction Tuning Code Large Language Models}.

\bibitem[{Muennighoff et~al.(2023{\natexlab{b}})Muennighoff, Rush, Barak, Scao, Piktus, Tazi, Pyysalo, Wolf, and Raffel}]{muennighoff2023scaling}
Niklas Muennighoff, Alexander~M. Rush, Boaz Barak, Teven~Le Scao, Aleksandra Piktus, Nouamane Tazi, Sampo Pyysalo, Thomas Wolf, and Colin Raffel. 2023{\natexlab{b}}.
\newblock \href {http://arxiv.org/abs/2305.16264} {Scaling Data-Constrained Language Models}.

\bibitem[{Muennighoff et~al.(2023{\natexlab{c}})Muennighoff, Tazi, Magne, and Reimers}]{muennighoff2023mteb}
Niklas Muennighoff, Nouamane Tazi, Loïc Magne, and Nils Reimers. 2023{\natexlab{c}}.
\newblock \href {http://arxiv.org/abs/2210.07316} {MTEB: Massive Text Embedding Benchmark}.

\bibitem[{Muennighoff et~al.(2023{\natexlab{d}})Muennighoff, Wang, Sutawika, Roberts, Biderman, Scao, Bari, Shen, Yong, Schoelkopf, Tang, Radev, Aji, Almubarak, Albanie, Alyafeai, Webson, Raff, and Raffel}]{muennighoff2023crosslingual}
Niklas Muennighoff, Thomas Wang, Lintang Sutawika, Adam Roberts, Stella Biderman, Teven~Le Scao, M~Saiful Bari, Sheng Shen, Zheng-Xin Yong, Hailey Schoelkopf, Xiangru Tang, Dragomir Radev, Alham~Fikri Aji, Khalid Almubarak, Samuel Albanie, Zaid Alyafeai, Albert Webson, Edward Raff, and Colin Raffel. 2023{\natexlab{d}}.
\newblock \href {http://arxiv.org/abs/2211.01786} {Crosslingual Generalization through Multitask Finetuning}.

\bibitem[{Narayan et~al.(2018)Narayan, Cohen, and Lapata}]{narayan2018dont}
Shashi Narayan, Shay~B. Cohen, and Mirella Lapata. 2018.
\newblock \href {http://arxiv.org/abs/1808.08745} {Don't Give Me the Details, Just the Summary! Topic-Aware Convolutional Neural Networks for Extreme Summarization}.

\bibitem[{Neelakantan et~al.(2022)Neelakantan, Xu, Puri, Radford, Han, Tworek, Yuan, Tezak, Kim, Hallacy, Heidecke, Shyam, Power, Nekoul, Sastry, Krueger, Schnurr, Such, Hsu, Thompson, Khan, Sherbakov, Jang, Welinder, and Weng}]{neelakantan2022text}
Arvind Neelakantan, Tao Xu, Raul Puri, Alec Radford, Jesse~Michael Han, Jerry Tworek, Qiming Yuan, Nikolas Tezak, Jong~Wook Kim, Chris Hallacy, Johannes Heidecke, Pranav Shyam, Boris Power, Tyna~Eloundou Nekoul, Girish Sastry, Gretchen Krueger, David Schnurr, Felipe~Petroski Such, Kenny Hsu, Madeleine Thompson, Tabarak Khan, Toki Sherbakov, Joanne Jang, Peter Welinder, and Lilian Weng. 2022.
\newblock \href {http://arxiv.org/abs/2201.10005} {Text and Code Embeddings by Contrastive Pre-Training}.

\bibitem[{Ni et~al.(2021{\natexlab{a}})Ni, Qu, Lu, Dai, Ábrego, Ma, Zhao, Luan, Hall, Chang, and Yang}]{ni2021large}
Jianmo Ni, Chen Qu, Jing Lu, Zhuyun Dai, Gustavo~Hernández Ábrego, Ji~Ma, Vincent~Y. Zhao, Yi~Luan, Keith~B. Hall, Ming-Wei Chang, and Yinfei Yang. 2021{\natexlab{a}}.
\newblock \href {http://arxiv.org/abs/2112.07899} {Large Dual Encoders Are Generalizable Retrievers}.

\bibitem[{Ni et~al.(2021{\natexlab{b}})Ni, Ábrego, Constant, Ma, Hall, Cer, and Yang}]{ni2021sentencet5}
Jianmo Ni, Gustavo~Hernández Ábrego, Noah Constant, Ji~Ma, Keith~B. Hall, Daniel Cer, and Yinfei Yang. 2021{\natexlab{b}}.
\newblock \href {http://arxiv.org/abs/2108.08877} {Sentence-T5: Scalable Sentence Encoders from Pre-trained Text-to-Text Models}.

\bibitem[{Nogueira and Cho(2020)}]{nogueira2020passage}
Rodrigo Nogueira and Kyunghyun Cho. 2020.
\newblock \href {http://arxiv.org/abs/1901.04085} {Passage Re-ranking with BERT}.

\bibitem[{OpenAI et~al.(2023)OpenAI, Achiam, Adler, Agarwal, Ahmad, Akkaya, Aleman, Almeida, Altenschmidt, Altman et~al.}]{openai2023gpt4}
OpenAI, Josh Achiam, Steven Adler, Sandhini Agarwal, Lama Ahmad, Ilge Akkaya, Florencia~Leoni Aleman, Diogo Almeida, Janko Altenschmidt, Sam Altman, et~al. 2023.
\newblock \href {http://arxiv.org/abs/2303.08774} {GPT-4 Technical Report}.

\bibitem[{Pal et~al.(2022)Pal, Umapathi, and Sankarasubbu}]{pal2022medmcqa}
Ankit Pal, Logesh~Kumar Umapathi, and Malaikannan Sankarasubbu. 2022.
\newblock \href {http://arxiv.org/abs/2203.14371} {MedMCQA : A Large-scale Multi-Subject Multi-Choice Dataset for Medical domain Question Answering}.

\bibitem[{Paszke et~al.(2019)Paszke, Gross, Massa, Lerer, Bradbury, Chanan, Killeen, Lin, Gimelshein, Antiga, Desmaison, Köpf, Yang, DeVito, Raison, Tejani, Chilamkurthy, Steiner, Fang, Bai, and Chintala}]{paszke2019pytorch}
Adam Paszke, Sam Gross, Francisco Massa, Adam Lerer, James Bradbury, Gregory Chanan, Trevor Killeen, Zeming Lin, Natalia Gimelshein, Luca Antiga, Alban Desmaison, Andreas Köpf, Edward Yang, Zach DeVito, Martin Raison, Alykhan Tejani, Sasank Chilamkurthy, Benoit Steiner, Lu~Fang, Junjie Bai, and Soumith Chintala. 2019.
\newblock \href {http://arxiv.org/abs/1912.01703} {PyTorch: An Imperative Style, High-Performance Deep Learning Library}.

\bibitem[{Pennington et~al.(2014)Pennington, Socher, and Manning}]{pennington2014glove}
Jeffrey Pennington, Richard Socher, and Christopher~D Manning. 2014.
\newblock \href {https://aclanthology.org/D14-1162/} {Glove: Global vectors for word representation}.

\bibitem[{Petroni et~al.(2021)Petroni, Piktus, Fan, Lewis, Yazdani, Cao, Thorne, Jernite, Karpukhin, Maillard, Plachouras, Rocktäschel, and Riedel}]{petroni2021kilt}
Fabio Petroni, Aleksandra Piktus, Angela Fan, Patrick Lewis, Majid Yazdani, Nicola~De Cao, James Thorne, Yacine Jernite, Vladimir Karpukhin, Jean Maillard, Vassilis Plachouras, Tim Rocktäschel, and Sebastian Riedel. 2021.
\newblock \href {http://arxiv.org/abs/2009.02252} {KILT: a Benchmark for Knowledge Intensive Language Tasks}.

\bibitem[{Pradeep et~al.(2023{\natexlab{a}})Pradeep, Sharifymoghaddam, and Lin}]{pradeep2023rankvicuna}
Ronak Pradeep, Sahel Sharifymoghaddam, and Jimmy Lin. 2023{\natexlab{a}}.
\newblock \href {http://arxiv.org/abs/2309.15088} {RankVicuna: Zero-Shot Listwise Document Reranking with Open-Source Large Language Models}.

\bibitem[{Pradeep et~al.(2023{\natexlab{b}})Pradeep, Sharifymoghaddam, and Lin}]{pradeep2023rankzephyr}
Ronak Pradeep, Sahel Sharifymoghaddam, and Jimmy Lin. 2023{\natexlab{b}}.
\newblock \href {http://arxiv.org/abs/2312.02724} {RankZephyr: Effective and Robust Zero-Shot Listwise Reranking is a Breeze!}

\bibitem[{Qiu et~al.(2022)Qiu, Li, Qu, Chen, She, Liu, Wu, and Wang}]{qiu2022dureaderretrieval}
Yifu Qiu, Hongyu Li, Yingqi Qu, Ying Chen, Qiaoqiao She, Jing Liu, Hua Wu, and Haifeng Wang. 2022.
\newblock \href {http://arxiv.org/abs/2203.10232} {DuReader\_retrieval: A Large-scale Chinese Benchmark for Passage Retrieval from Web Search Engine}.

\bibitem[{Radford et~al.(2021)Radford, Kim, Hallacy, Ramesh, Goh, Agarwal, Sastry, Askell, Mishkin, Clark, Krueger, and Sutskever}]{radford2021learning}
Alec Radford, Jong~Wook Kim, Chris Hallacy, Aditya Ramesh, Gabriel Goh, Sandhini Agarwal, Girish Sastry, Amanda Askell, Pamela Mishkin, Jack Clark, Gretchen Krueger, and Ilya Sutskever. 2021.
\newblock \href {http://arxiv.org/abs/2103.00020} {Learning Transferable Visual Models From Natural Language Supervision}.

\bibitem[{Radford et~al.(2018)Radford, Narasimhan, Salimans, Sutskever et~al.}]{radford2018improving}
Alec Radford, Karthik Narasimhan, Tim Salimans, Ilya Sutskever, et~al. 2018.
\newblock \href {https://cdn.openai.com/research-covers/language-unsupervised/language_understanding_paper.pdf} {Improving language understanding by generative pre-training}.

\bibitem[{Radford et~al.(2019)Radford, Wu, Child, Luan, Amodei, Sutskever et~al.}]{radford2019language}
Alec Radford, Jeffrey Wu, Rewon Child, David Luan, Dario Amodei, Ilya Sutskever, et~al. 2019.
\newblock \href {https://d4mucfpksywv.cloudfront.net/better-language-models/language_models_are_unsupervised_multitask_learners.pdf} {Language models are unsupervised multitask learners}.

\bibitem[{Rae et~al.(2022)Rae, Borgeaud, Cai, Millican, Hoffmann, Song, Aslanides, Henderson, Ring, Young et~al.}]{rae2022scaling}
Jack~W. Rae, Sebastian Borgeaud, Trevor Cai, Katie Millican, Jordan Hoffmann, Francis Song, John Aslanides, Sarah Henderson, Roman Ring, Susannah Young, et~al. 2022.
\newblock \href {http://arxiv.org/abs/2112.11446} {Scaling Language Models: Methods, Analysis \& Insights from Training Gopher}.

\bibitem[{Rafailov et~al.(2023)Rafailov, Sharma, Mitchell, Ermon, Manning, and Finn}]{rafailov2023direct}
Rafael Rafailov, Archit Sharma, Eric Mitchell, Stefano Ermon, Christopher~D. Manning, and Chelsea Finn. 2023.
\newblock \href {http://arxiv.org/abs/2305.18290} {Direct Preference Optimization: Your Language Model is Secretly a Reward Model}.

\bibitem[{Raffel et~al.(2023)Raffel, Shazeer, Roberts, Lee, Narang, Matena, Zhou, Li, and Liu}]{raffel2023exploring}
Colin Raffel, Noam Shazeer, Adam Roberts, Katherine Lee, Sharan Narang, Michael Matena, Yanqi Zhou, Wei Li, and Peter~J. Liu. 2023.
\newblock \href {http://arxiv.org/abs/1910.10683} {Exploring the Limits of Transfer Learning with a Unified Text-to-Text Transformer}.

\bibitem[{Rajpurkar et~al.(2016)Rajpurkar, Zhang, Lopyrev, and Liang}]{rajpurkar2016squad}
Pranav Rajpurkar, Jian Zhang, Konstantin Lopyrev, and Percy Liang. 2016.
\newblock \href {http://arxiv.org/abs/1606.05250} {SQuAD: 100,000+ Questions for Machine Comprehension of Text}.

\bibitem[{Reed et~al.(2022)Reed, Zolna, Parisotto, Colmenarejo, Novikov, Barth-Maron, Gimenez, Sulsky, Kay, Springenberg, Eccles, Bruce, Razavi, Edwards, Heess, Chen, Hadsell, Vinyals, Bordbar, and de~Freitas}]{reed2022generalist}
Scott Reed, Konrad Zolna, Emilio Parisotto, Sergio~Gomez Colmenarejo, Alexander Novikov, Gabriel Barth-Maron, Mai Gimenez, Yury Sulsky, Jackie Kay, Jost~Tobias Springenberg, Tom Eccles, Jake Bruce, Ali Razavi, Ashley Edwards, Nicolas Heess, Yutian Chen, Raia Hadsell, Oriol Vinyals, Mahyar Bordbar, and Nando de~Freitas. 2022.
\newblock \href {http://arxiv.org/abs/2205.06175} {A Generalist Agent}.

\bibitem[{Reimers and Gurevych(2019)}]{reimers2019sentencebert}
Nils Reimers and Iryna Gurevych. 2019.
\newblock \href {http://arxiv.org/abs/1908.10084} {Sentence-BERT: Sentence Embeddings using Siamese BERT-Networks}.

\bibitem[{Rubin et~al.(2022)Rubin, Herzig, and Berant}]{rubin2022learning}
Ohad Rubin, Jonathan Herzig, and Jonathan Berant. 2022.
\newblock \href {http://arxiv.org/abs/2112.08633} {Learning To Retrieve Prompts for In-Context Learning}.

\bibitem[{Rush et~al.(2015)Rush, Chopra, and Weston}]{rush2015neural}
Alexander~M. Rush, Sumit Chopra, and Jason Weston. 2015.
\newblock \href {http://arxiv.org/abs/1509.00685} {A Neural Attention Model for Abstractive Sentence Summarization}.

\bibitem[{Sanh et~al.(2022)Sanh, Webson, Raffel, Bach, Sutawika, Alyafeai, Chaffin, Stiegler, Scao, Raja et~al.}]{sanh2022multitask}
Victor Sanh, Albert Webson, Colin Raffel, Stephen~H. Bach, Lintang Sutawika, Zaid Alyafeai, Antoine Chaffin, Arnaud Stiegler, Teven~Le Scao, Arun Raja, et~al. 2022.
\newblock \href {http://arxiv.org/abs/2110.08207} {Multitask Prompted Training Enables Zero-Shot Task Generalization}.

\bibitem[{Scao et~al.(2022)Scao, Wang, Hesslow, Saulnier, Bekman, Bari, Biderman, Elsahar, Muennighoff, Phang, Press, Raffel, Sanh, Shen, Sutawika, Tae, Yong, Launay, and Beltagy}]{scao2022language}
Teven~Le Scao, Thomas Wang, Daniel Hesslow, Lucile Saulnier, Stas Bekman, M~Saiful Bari, Stella Biderman, Hady Elsahar, Niklas Muennighoff, Jason Phang, Ofir Press, Colin Raffel, Victor Sanh, Sheng Shen, Lintang Sutawika, Jaesung Tae, Zheng~Xin Yong, Julien Launay, and Iz~Beltagy. 2022.
\newblock \href {http://arxiv.org/abs/2210.15424} {What Language Model to Train if You Have One Million GPU Hours?}

\bibitem[{Schick et~al.(2023)Schick, Dwivedi-Yu, Dessì, Raileanu, Lomeli, Zettlemoyer, Cancedda, and Scialom}]{schick2023toolformer}
Timo Schick, Jane Dwivedi-Yu, Roberto Dessì, Roberta Raileanu, Maria Lomeli, Luke Zettlemoyer, Nicola Cancedda, and Thomas Scialom. 2023.
\newblock \href {http://arxiv.org/abs/2302.04761} {Toolformer: Language Models Can Teach Themselves to Use Tools}.

\bibitem[{Schulman et~al.(2017)Schulman, Wolski, Dhariwal, Radford, and Klimov}]{schulman2017proximal}
John Schulman, Filip Wolski, Prafulla Dhariwal, Alec Radford, and Oleg Klimov. 2017.
\newblock Proximal policy optimization algorithms.
\newblock \emph{arXiv preprint arXiv:1707.06347}.

\bibitem[{Shen et~al.(2022)Shen, Lo, Yu, Dahlberg, Schlanger, and Downey}]{shen2022multilexsum}
Zejiang Shen, Kyle Lo, Lauren Yu, Nathan Dahlberg, Margo Schlanger, and Doug Downey. 2022.
\newblock \href {http://arxiv.org/abs/2206.10883} {Multi-LexSum: Real-World Summaries of Civil Rights Lawsuits at Multiple Granularities}.

\bibitem[{Shi et~al.(2023)Shi, Min, Yasunaga, Seo, James, Lewis, Zettlemoyer, and tau Yih}]{shi2023replug}
Weijia Shi, Sewon Min, Michihiro Yasunaga, Minjoon Seo, Rich James, Mike Lewis, Luke Zettlemoyer, and Wen tau Yih. 2023.
\newblock \href {http://arxiv.org/abs/2301.12652} {REPLUG: Retrieval-Augmented Black-Box Language Models}.

\bibitem[{Singh et~al.(2022)Singh, Hu, Goswami, Couairon, Galuba, Rohrbach, and Kiela}]{singh2022flava}
Amanpreet Singh, Ronghang Hu, Vedanuj Goswami, Guillaume Couairon, Wojciech Galuba, Marcus Rohrbach, and Douwe Kiela. 2022.
\newblock \href {http://arxiv.org/abs/2112.04482} {FLAVA: A Foundational Language And Vision Alignment Model}.

\bibitem[{Singh et~al.(2024)Singh, Vargus, Dsouza, Karlsson, Mahendiran, Ko, Shandilya, Patel, Mataciunas, OMahony, Zhang, Hettiarachchi, Wilson, Machado, Moura, Krzemiński, Fadaei, Ergün, Okoh, Alaagib, Mudannayake, Alyafeai, Chien, Ruder, Guthikonda, Alghamdi, Gehrmann, Muennighoff, Bartolo, Kreutzer, Üstün, Fadaee, and Hooker}]{singh2024aya}
Shivalika Singh, Freddie Vargus, Daniel Dsouza, Börje~F. Karlsson, Abinaya Mahendiran, Wei-Yin Ko, Herumb Shandilya, Jay Patel, Deividas Mataciunas, Laura OMahony, Mike Zhang, Ramith Hettiarachchi, Joseph Wilson, Marina Machado, Luisa~Souza Moura, Dominik Krzemiński, Hakimeh Fadaei, Irem Ergün, Ifeoma Okoh, Aisha Alaagib, Oshan Mudannayake, Zaid Alyafeai, Vu~Minh Chien, Sebastian Ruder, Surya Guthikonda, Emad~A. Alghamdi, Sebastian Gehrmann, Niklas Muennighoff, Max Bartolo, Julia Kreutzer, Ahmet Üstün, Marzieh Fadaee, and Sara Hooker. 2024.
\newblock \href {http://arxiv.org/abs/2402.06619} {Aya Dataset: An Open-Access Collection for Multilingual Instruction Tuning}.

\bibitem[{Soldaini et~al.(2024)Soldaini, Kinney, Bhagia, Schwenk, Atkinson, Authur, Bogin, Chandu, Dumas, Elazar, Hofmann, Jha, Kumar, Lucy, Lyu, Lambert, Magnusson, Morrison, Muennighoff, Naik, Nam, Peters, Ravichander, Richardson, Shen, Strubell, Subramani, Tafjord, Walsh, Zettlemoyer, Smith, Hajishirzi, Beltagy, Groeneveld, Dodge, and Lo}]{soldaini2024dolma}
Luca Soldaini, Rodney Kinney, Akshita Bhagia, Dustin Schwenk, David Atkinson, Russell Authur, Ben Bogin, Khyathi Chandu, Jennifer Dumas, Yanai Elazar, Valentin Hofmann, Ananya~Harsh Jha, Sachin Kumar, Li~Lucy, Xinxi Lyu, Nathan Lambert, Ian Magnusson, Jacob Morrison, Niklas Muennighoff, Aakanksha Naik, Crystal Nam, Matthew~E. Peters, Abhilasha Ravichander, Kyle Richardson, Zejiang Shen, Emma Strubell, Nishant Subramani, Oyvind Tafjord, Pete Walsh, Luke Zettlemoyer, Noah~A. Smith, Hannaneh Hajishirzi, Iz~Beltagy, Dirk Groeneveld, Jesse Dodge, and Kyle Lo. 2024.
\newblock \href {http://arxiv.org/abs/2402.00159} {Dolma: an Open Corpus of Three Trillion Tokens for Language Model Pretraining Research}.

\bibitem[{Srivastava et~al.(2023)Srivastava, Rastogi, Rao, Shoeb, Abid, Fisch, Brown, Santoro, Gupta, Garriga-Alonso et~al.}]{srivastava2023imitation}
Aarohi Srivastava, Abhinav Rastogi, Abhishek Rao, Abu Awal~Md Shoeb, Abubakar Abid, Adam Fisch, Adam~R. Brown, Adam Santoro, Aditya Gupta, Adrià Garriga-Alonso, et~al. 2023.
\newblock \href {http://arxiv.org/abs/2206.04615} {Beyond the Imitation Game: Quantifying and extrapolating the capabilities of language models}.

\bibitem[{Su et~al.(2023)Su, Shi, Kasai, Wang, Hu, Ostendorf, tau Yih, Smith, Zettlemoyer, and Yu}]{su2023embedder}
Hongjin Su, Weijia Shi, Jungo Kasai, Yizhong Wang, Yushi Hu, Mari Ostendorf, Wen tau Yih, Noah~A. Smith, Luke Zettlemoyer, and Tao Yu. 2023.
\newblock \href {http://arxiv.org/abs/2212.09741} {One Embedder, Any Task: Instruction-Finetuned Text Embeddings}.

\bibitem[{Su et~al.(2017)Su, Wu, Huang, Hong, and Wang}]{su2017chatbot}
Ming-Hsiang Su, Chung-Hsien Wu, Kun-Yi Huang, Qian-Bei Hong, and Hsin-Min Wang. 2017.
\newblock \href {https://ieeexplore.ieee.org/document/8336091} {A chatbot using LSTM-based multi-layer embedding for elderly care}.

\bibitem[{Sun et~al.(2023)Sun, Yan, Ma, Wang, Ren, Chen, Yin, and Ren}]{sun2023chatgpt}
Weiwei Sun, Lingyong Yan, Xinyu Ma, Shuaiqiang Wang, Pengjie Ren, Zhumin Chen, Dawei Yin, and Zhaochun Ren. 2023.
\newblock \href {http://arxiv.org/abs/2304.09542} {Is ChatGPT Good at Search? Investigating Large Language Models as Re-Ranking Agents}.

\bibitem[{Sutskever et~al.(2014)Sutskever, Vinyals, and Le}]{sutskever2014sequence}
Ilya Sutskever, Oriol Vinyals, and Quoc~V. Le. 2014.
\newblock \href {http://arxiv.org/abs/1409.3215} {Sequence to Sequence Learning with Neural Networks}.

\bibitem[{Suzgun et~al.(2022)Suzgun, Scales, Schärli, Gehrmann, Tay, Chung, Chowdhery, Le, Chi, Zhou, and Wei}]{suzgun2022challenging}
Mirac Suzgun, Nathan Scales, Nathanael Schärli, Sebastian Gehrmann, Yi~Tay, Hyung~Won Chung, Aakanksha Chowdhery, Quoc~V. Le, Ed~H. Chi, Denny Zhou, and Jason Wei. 2022.
\newblock \href {http://arxiv.org/abs/2210.09261} {Challenging BIG-Bench Tasks and Whether Chain-of-Thought Can Solve Them}.

\bibitem[{Team(2021{\natexlab{a}})}]{StackExchangeDataset}
Flax Sentence~Embeddings Team. 2021{\natexlab{a}}.
\newblock \href {https://hf.co/datasets/flax-sentence-embeddings/} {Stack Exchange question pairs}.

\bibitem[{Team et~al.(2023)Team, Anil, Borgeaud, Wu, Alayrac, Yu, Soricut, Schalkwyk, Dai, Hauth et~al.}]{geminiteam2023gemini}
Gemini Team, Rohan Anil, Sebastian Borgeaud, Yonghui Wu, Jean-Baptiste Alayrac, Jiahui Yu, Radu Soricut, Johan Schalkwyk, Andrew~M. Dai, Anja Hauth, et~al. 2023.
\newblock \href {http://arxiv.org/abs/2312.11805} {Gemini: A Family of Highly Capable Multimodal Models}.

\bibitem[{Team(2021{\natexlab{b}})}]{Reddit}
Sentence~Transformers Team. 2021{\natexlab{b}}.
\newblock \href {https://hf.co/datasets/sentence-transformers/reddit-title-body} {Reddit Title Body}.

\bibitem[{Team(2021{\natexlab{c}})}]{NPR}
Sentence~Transformers Team. 2021{\natexlab{c}}.
\newblock \href {https://hf.co/datasets/sentence-transformers/embedding-training-data} {(Title, Body) pairs from the npr.org website}.

\bibitem[{Thakur et~al.(2021)Thakur, Reimers, Rücklé, Srivastava, and Gurevych}]{thakur2021beir}
Nandan Thakur, Nils Reimers, Andreas Rücklé, Abhishek Srivastava, and Iryna Gurevych. 2021.
\newblock \href {http://arxiv.org/abs/2104.08663} {BEIR: A Heterogenous Benchmark for Zero-shot Evaluation of Information Retrieval Models}.

\bibitem[{Thorne et~al.(2018)Thorne, Vlachos, Christodoulopoulos, and Mittal}]{thorne2018fever}
James Thorne, Andreas Vlachos, Christos Christodoulopoulos, and Arpit Mittal. 2018.
\newblock \href {http://arxiv.org/abs/1803.05355} {FEVER: a large-scale dataset for Fact Extraction and VERification}.

\bibitem[{Touvron et~al.(2023)Touvron, Lavril, Izacard, Martinet, Lachaux, Lacroix, Rozière, Goyal, Hambro, Azhar, Rodriguez, Joulin, Grave, and Lample}]{touvron2023llama}
Hugo Touvron, Thibaut Lavril, Gautier Izacard, Xavier Martinet, Marie-Anne Lachaux, Timothée Lacroix, Baptiste Rozière, Naman Goyal, Eric Hambro, Faisal Azhar, Aurelien Rodriguez, Armand Joulin, Edouard Grave, and Guillaume Lample. 2023.
\newblock \href {http://arxiv.org/abs/2302.13971} {LLaMA: Open and Efficient Foundation Language Models}.

\bibitem[{Tunstall et~al.(2023)Tunstall, Beeching, Lambert, Rajani, Rasul, Belkada, Huang, von Werra, Fourrier, Habib, Sarrazin, Sanseviero, Rush, and Wolf}]{tunstall2023zephyr}
Lewis Tunstall, Edward Beeching, Nathan Lambert, Nazneen Rajani, Kashif Rasul, Younes Belkada, Shengyi Huang, Leandro von Werra, Clémentine Fourrier, Nathan Habib, Nathan Sarrazin, Omar Sanseviero, Alexander~M. Rush, and Thomas Wolf. 2023.
\newblock \href {http://arxiv.org/abs/2310.16944} {Zephyr: Direct Distillation of LM Alignment}.

\bibitem[{Vaswani et~al.(2023)Vaswani, Shazeer, Parmar, Uszkoreit, Jones, Gomez, Kaiser, and Polosukhin}]{vaswani2023attention}
Ashish Vaswani, Noam Shazeer, Niki Parmar, Jakob Uszkoreit, Llion Jones, Aidan~N. Gomez, Lukasz Kaiser, and Illia Polosukhin. 2023.
\newblock \href {http://arxiv.org/abs/1706.03762} {Attention Is All You Need}.

\bibitem[{Wang and Komatsuzaki(2021)}]{wang2021gpt}
Ben Wang and Aran Komatsuzaki. 2021.
\newblock \href {https://github.com/kingoflolz/mesh-transformer-jax} {GPT-J-6B: A 6 Billion Parameter Autoregressive Language Model}.

\bibitem[{Wang and Kuo(2020)}]{wang2020sbertwk}
Bin Wang and C.~C.~Jay Kuo. 2020.
\newblock \href {http://arxiv.org/abs/2002.06652} {SBERT-WK: A Sentence Embedding Method by Dissecting BERT-based Word Models}.

\bibitem[{Wang et~al.(2022{\natexlab{a}})Wang, Yang, Huang, Jiao, Yang, Jiang, Majumder, and Wei}]{wang2022text}
Liang Wang, Nan Yang, Xiaolong Huang, Binxing Jiao, Linjun Yang, Daxin Jiang, Rangan Majumder, and Furu Wei. 2022{\natexlab{a}}.
\newblock \href {http://arxiv.org/abs/2212.03533} {Text Embeddings by Weakly-Supervised Contrastive Pre-training}.

\bibitem[{Wang et~al.(2024)Wang, Yang, Huang, Yang, Majumder, and Wei}]{wang2024improving}
Liang Wang, Nan Yang, Xiaolong Huang, Linjun Yang, Rangan Majumder, and Furu Wei. 2024.
\newblock \href {http://arxiv.org/abs/2401.00368} {Improving Text Embeddings with Large Language Models}.

\bibitem[{Wang et~al.(2022{\natexlab{b}})Wang, Roberts, Hesslow, Scao, Chung, Beltagy, Launay, and Raffel}]{wang2022language}
Thomas Wang, Adam Roberts, Daniel Hesslow, Teven~Le Scao, Hyung~Won Chung, Iz~Beltagy, Julien Launay, and Colin Raffel. 2022{\natexlab{b}}.
\newblock \href {http://arxiv.org/abs/2204.05832} {What Language Model Architecture and Pretraining Objective Work Best for Zero-Shot Generalization?}

\bibitem[{Wang et~al.(2023)Wang, Ivison, Dasigi, Hessel, Khot, Chandu, Wadden, MacMillan, Smith, Beltagy, and Hajishirzi}]{wang2023far}
Yizhong Wang, Hamish Ivison, Pradeep Dasigi, Jack Hessel, Tushar Khot, Khyathi~Raghavi Chandu, David Wadden, Kelsey MacMillan, Noah~A. Smith, Iz~Beltagy, and Hannaneh Hajishirzi. 2023.
\newblock \href {http://arxiv.org/abs/2306.04751} {How Far Can Camels Go? Exploring the State of Instruction Tuning on Open Resources}.

\bibitem[{Wang et~al.(2022{\natexlab{c}})Wang, Mishra, Alipoormolabashi, Kordi, Mirzaei, Arunkumar, Ashok, Dhanasekaran, Naik, Stap et~al.}]{wang2022supernaturalinstructions}
Yizhong Wang, Swaroop Mishra, Pegah Alipoormolabashi, Yeganeh Kordi, Amirreza Mirzaei, Anjana Arunkumar, Arjun Ashok, Arut~Selvan Dhanasekaran, Atharva Naik, David Stap, et~al. 2022{\natexlab{c}}.
\newblock \href {http://arxiv.org/abs/2204.07705} {Super-NaturalInstructions: Generalization via Declarative Instructions on 1600+ NLP Tasks}.

\bibitem[{Wei et~al.(2022)Wei, Bosma, Zhao, Guu, Yu, Lester, Du, Dai, and Le}]{wei2022finetuned}
Jason Wei, Maarten Bosma, Vincent~Y. Zhao, Kelvin Guu, Adams~Wei Yu, Brian Lester, Nan Du, Andrew~M. Dai, and Quoc~V. Le. 2022.
\newblock \href {http://arxiv.org/abs/2109.01652} {Finetuned Language Models Are Zero-Shot Learners}.

\bibitem[{Wei et~al.(2023)Wei, Wang, Schuurmans, Bosma, Ichter, Xia, Chi, Le, and Zhou}]{wei2023chainofthought}
Jason Wei, Xuezhi Wang, Dale Schuurmans, Maarten Bosma, Brian Ichter, Fei Xia, Ed~Chi, Quoc Le, and Denny Zhou. 2023.
\newblock \href {http://arxiv.org/abs/2201.11903} {Chain-of-Thought Prompting Elicits Reasoning in Large Language Models}.

\bibitem[{Xia et~al.(2023)Xia, Gao, Zeng, and Chen}]{xia2023sheared}
Mengzhou Xia, Tianyu Gao, Zhiyuan Zeng, and Danqi Chen. 2023.
\newblock \href {http://arxiv.org/abs/2310.06694} {Sheared LLaMA: Accelerating Language Model Pre-training via Structured Pruning}.

\bibitem[{Xiao and Liu(2022)}]{xiao2022retromae2}
Shitao Xiao and Zheng Liu. 2022.
\newblock \href {http://arxiv.org/abs/2211.08769} {RetroMAE v2: Duplex Masked Auto-Encoder For Pre-Training Retrieval-Oriented Language Models}.

\bibitem[{Xiao et~al.(2022)Xiao, Liu, Shao, and Cao}]{xiao2022retromae}
Shitao Xiao, Zheng Liu, Yingxia Shao, and Zhao Cao. 2022.
\newblock \href {http://arxiv.org/abs/2205.12035} {RetroMAE: Pre-Training Retrieval-oriented Language Models Via Masked Auto-Encoder}.

\bibitem[{Xiao et~al.(2023)Xiao, Liu, Zhang, and Muennighoff}]{xiao2023cpack}
Shitao Xiao, Zheng Liu, Peitian Zhang, and Niklas Muennighoff. 2023.
\newblock \href {http://arxiv.org/abs/2309.07597} {C-Pack: Packaged Resources To Advance General Chinese Embedding}.

\bibitem[{Xie et~al.(2023)Xie, Dong, Wang, Lv, Yao, Gan, Wu, Li, Li, Liu, and Ma}]{xie2023t2ranking}
Xiaohui Xie, Qian Dong, Bingning Wang, Feiyang Lv, Ting Yao, Weinan Gan, Zhijing Wu, Xiangsheng Li, Haitao Li, Yiqun Liu, and Jin Ma. 2023.
\newblock \href {http://arxiv.org/abs/2304.03679} {T2Ranking: A large-scale Chinese Benchmark for Passage Ranking}.

\bibitem[{Yang et~al.(2018)Yang, Qi, Zhang, Bengio, Cohen, Salakhutdinov, and Manning}]{yang2018hotpotqa}
Zhilin Yang, Peng Qi, Saizheng Zhang, Yoshua Bengio, William~W. Cohen, Ruslan Salakhutdinov, and Christopher~D. Manning. 2018.
\newblock \href {http://arxiv.org/abs/1809.09600} {HotpotQA: A Dataset for Diverse, Explainable Multi-hop Question Answering}.

\bibitem[{Yasunaga et~al.(2023)Yasunaga, Aghajanyan, Shi, James, Leskovec, Liang, Lewis, Zettlemoyer, and tau Yih}]{yasunaga2023retrievalaugmented}
Michihiro Yasunaga, Armen Aghajanyan, Weijia Shi, Rich James, Jure Leskovec, Percy Liang, Mike Lewis, Luke Zettlemoyer, and Wen tau Yih. 2023.
\newblock \href {http://arxiv.org/abs/2211.12561} {Retrieval-Augmented Multimodal Language Modeling}.

\bibitem[{Yin et~al.(2016)Yin, Jiang, Lu, Shang, Li, and Li}]{yin2016neural}
Jun Yin, Xin Jiang, Zhengdong Lu, Lifeng Shang, Hang Li, and Xiaoming Li. 2016.
\newblock \href {http://arxiv.org/abs/1512.01337} {Neural Generative Question Answering}.

\bibitem[{Yong et~al.(2023)Yong, Schoelkopf, Muennighoff, Aji, Adelani, Almubarak, Bari, Sutawika, Kasai, Baruwa, Winata, Biderman, Raff, Radev, and Nikoulina}]{yong2023bloom1}
Zheng-Xin Yong, Hailey Schoelkopf, Niklas Muennighoff, Alham~Fikri Aji, David~Ifeoluwa Adelani, Khalid Almubarak, M~Saiful Bari, Lintang Sutawika, Jungo Kasai, Ahmed Baruwa, Genta~Indra Winata, Stella Biderman, Edward Raff, Dragomir Radev, and Vassilina Nikoulina. 2023.
\newblock \href {http://arxiv.org/abs/2212.09535} {BLOOM+1: Adding Language Support to BLOOM for Zero-Shot Prompting}.

\bibitem[{Young et~al.(2014)Young, Lai, Hodosh, and Hockenmaier}]{young2014image}
Peter Young, Alice Lai, Micah Hodosh, and Julia Hockenmaier. 2014.
\newblock \href {https://aclanthology.org/Q14-1006} {From image descriptions to visual denotations: New similarity metrics for semantic inference over event descriptions}.

\bibitem[{Yu et~al.(2022)Yu, Wang, Vasudevan, Yeung, Seyedhosseini, and Wu}]{yu2022coca}
Jiahui Yu, Zirui Wang, Vijay Vasudevan, Legg Yeung, Mojtaba Seyedhosseini, and Yonghui Wu. 2022.
\newblock \href {http://arxiv.org/abs/2205.01917} {CoCa: Contrastive Captioners are Image-Text Foundation Models}.

\bibitem[{Zaken et~al.(2022)Zaken, Ravfogel, and Goldberg}]{zaken2022bitfit}
Elad~Ben Zaken, Shauli Ravfogel, and Yoav Goldberg. 2022.
\newblock \href {http://arxiv.org/abs/2106.10199} {BitFit: Simple Parameter-efficient Fine-tuning for Transformer-based Masked Language-models}.

\bibitem[{Zhang et~al.(2016)Zhang, Zhao, and LeCun}]{zhang2016characterlevel}
Xiang Zhang, Junbo Zhao, and Yann LeCun. 2016.
\newblock \href {http://arxiv.org/abs/1509.01626} {Character-level Convolutional Networks for Text Classification}.

\bibitem[{Zhang et~al.(2023)Zhang, Li, Zhang, Long, Xie, Zhang, and Zhang}]{zhang2023language}
Xin Zhang, Zehan Li, Yanzhao Zhang, Dingkun Long, Pengjun Xie, Meishan Zhang, and Min Zhang. 2023.
\newblock \href {http://arxiv.org/abs/2310.08232} {Language Models are Universal Embedders}.

\bibitem[{Zhang et~al.(2021)Zhang, Ma, Shi, and Lin}]{zhang2021mr}
Xinyu Zhang, Xueguang Ma, Peng Shi, and Jimmy Lin. 2021.
\newblock \href {http://arxiv.org/abs/2108.08787} {Mr. TyDi: A Multi-lingual Benchmark for Dense Retrieval}.

\bibitem[{Zhang et~al.(2022)Zhang, Thakur, Ogundepo, Kamalloo, Alfonso-Hermelo, Li, Liu, Rezagholizadeh, and Lin}]{zhang2022making}
Xinyu Zhang, Nandan Thakur, Odunayo Ogundepo, Ehsan Kamalloo, David Alfonso-Hermelo, Xiaoguang Li, Qun Liu, Mehdi Rezagholizadeh, and Jimmy Lin. 2022.
\newblock \href {http://arxiv.org/abs/2210.09984} {Making a MIRACL: Multilingual Information Retrieval Across a Continuum of Languages}.

\bibitem[{Zhao et~al.(2023)Zhao, Gu, Varma, Luo, Huang, Xu, Wright, Shojanazeri, Ott, Shleifer, Desmaison, Balioglu, Damania, Nguyen, Chauhan, Hao, Mathews, and Li}]{zhao2023pytorch}
Yanli Zhao, Andrew Gu, Rohan Varma, Liang Luo, Chien-Chin Huang, Min Xu, Less Wright, Hamid Shojanazeri, Myle Ott, Sam Shleifer, Alban Desmaison, Can Balioglu, Pritam Damania, Bernard Nguyen, Geeta Chauhan, Yuchen Hao, Ajit Mathews, and Shen Li. 2023.
\newblock \href {http://arxiv.org/abs/2304.11277} {PyTorch FSDP: Experiences on Scaling Fully Sharded Data Parallel}.

\bibitem[{Zheng et~al.(2023)Zheng, Chiang, Sheng, Zhuang, Wu, Zhuang, Lin, Li, Li, Xing, Zhang, Gonzalez, and Stoica}]{zheng2023judging}
Lianmin Zheng, Wei-Lin Chiang, Ying Sheng, Siyuan Zhuang, Zhanghao Wu, Yonghao Zhuang, Zi~Lin, Zhuohan Li, Dacheng Li, Eric~P. Xing, Hao Zhang, Joseph~E. Gonzalez, and Ion Stoica. 2023.
\newblock \href {http://arxiv.org/abs/2306.05685} {Judging LLM-as-a-Judge with MT-Bench and Chatbot Arena}.

\bibitem[{Zhou et~al.(2023)Zhou, Liu, Xu, Iyer, Sun, Mao, Ma, Efrat, Yu, Yu, Zhang, Ghosh, Lewis, Zettlemoyer, and Levy}]{zhou2023lima}
Chunting Zhou, Pengfei Liu, Puxin Xu, Srini Iyer, Jiao Sun, Yuning Mao, Xuezhe Ma, Avia Efrat, Ping Yu, Lili Yu, Susan Zhang, Gargi Ghosh, Mike Lewis, Luke Zettlemoyer, and Omer Levy. 2023.
\newblock \href {http://arxiv.org/abs/2305.11206} {LIMA: Less Is More for Alignment}.

\bibitem[{Zhu et~al.(2024)Zhu, Yuan, Wang, Liu, Liu, Deng, Chen, Dou, and Wen}]{zhu2024large}
Yutao Zhu, Huaying Yuan, Shuting Wang, Jiongnan Liu, Wenhan Liu, Chenlong Deng, Haonan Chen, Zhicheng Dou, and Ji-Rong Wen. 2024.
\newblock \href {http://arxiv.org/abs/2308.07107} {Large Language Models for Information Retrieval: A Survey}.

\bibitem[{Zhuo et~al.(2024)Zhuo, Zebaze, Suppattarachai, von Werra, de~Vries, Liu, and Muennighoff}]{zhuo2024astraios}
Terry~Yue Zhuo, Armel Zebaze, Nitchakarn Suppattarachai, Leandro von Werra, Harm de~Vries, Qian Liu, and Niklas Muennighoff. 2024.
\newblock \href {http://arxiv.org/abs/2401.00788} {Astraios: Parameter-Efficient Instruction Tuning Code Large Language Models}.

\bibitem[{Üstün et~al.(2024)Üstün, Aryabumi, Yong, Ko, D'souza, Onilude, Bhandari, Singh, Ooi, Kayid, Vargus, Blunsom, Longpre, Muennighoff, Fadaee, Kreutzer, and Hooker}]{üstün2024aya}
Ahmet Üstün, Viraat Aryabumi, Zheng-Xin Yong, Wei-Yin Ko, Daniel D'souza, Gbemileke Onilude, Neel Bhandari, Shivalika Singh, Hui-Lee Ooi, Amr Kayid, Freddie Vargus, Phil Blunsom, Shayne Longpre, Niklas Muennighoff, Marzieh Fadaee, Julia Kreutzer, and Sara Hooker. 2024.
\newblock \href {http://arxiv.org/abs/2402.07827} {Aya Model: An Instruction Finetuned Open-Access Multilingual Language Model}.

\end{thebibliography}
\bibliographystyle{acl_natbib}

\newpage
\appendix

\part{}
\section*{\centering \LARGE{Appendix}}
\mtcsettitle{parttoc}{Contents}
\parttoc

\clearpage

\section{Contributions}
\label{sec:contributions}

Niklas Muennighoff conceived of the project and led its execution. He designed the experiments and implemented, trained, and evaluated all models, and wrote most of the paper. Hongjin Su created MEDI2, implemented and ran reranking, and helped with experiments. Liang Wang ran several dataset ablations and contributed to framing. Nan Yang, Furu Wei, Tao Yu, Amanpreet Singh, and Douwe Kiela advised the project. All authors helped edit the paper.

\section{Artifacts}
\label{sec:artifacts}

\begin{longtable}{p{2.9cm} p{10.2cm}}
\caption{\textbf{Produced artifacts.}} \\
\toprule
Artifact & Public Link\\
\midrule
\multicolumn{2}{c}{\textit{\autoref{tab:kto}}} \\
\midrule
7B KTO & \url{https://hf.co/GritLM/GritLM-7B-KTO} \\
8x7B KTO & \url{https://hf.co/GritLM/GritLM-8x7B-KTO} \\
\midrule
\multicolumn{2}{c}{\textit{\autoref{tab:unifiedattn}}} \\
\midrule
CCCC WM & \url{https://hf.co/GritLM/gritlm_m7_sq2048_medi} \\
CCCC LT & \url{https://hf.co/GritLM/gritlm_m7_sq2048_medi_lasttoken} \\
BBCC M & \url{https://hf.co/GritLM/gritlm_m7_sq2048_medi_bbcc} \\
\midrule
\multicolumn{2}{c}{\textit{\autoref{tab:embattn}}} \\
\midrule
CC WM & \url{https://hf.co/GritLM/emb_m7_sq2048_medi} \\
CB M & \url{https://hf.co/GritLM/emb_m7_sq2048_medi_cb} \\
BB M &  \url{https://hf.co/GritLM/emb_m7_sq2048_medi_bb} \\
\midrule
\multicolumn{2}{c}{\textit{\autoref{tab:genattn}}} \\
\midrule
CC & \url{https://hf.co/GritLM/gen_m7_sq2048_tulu2_ep1} \\
BC & \url{https://hf.co/GritLM/gen_m7_sq2048_tulu2_bc} \\
BC IL & \url{https://hf.co/GritLM/gen_m7_sq2048_tulu2_bcil} \\
\midrule
\multicolumn{2}{c}{\textit{\autoref{tab:appbasemodel}}} \\
\midrule
Mistral 7B & \url{https://hf.co/GritLM/gritlm_m7_sq1024_st100_zephfmt} \\
\llama{} 2 7B & \url{https://hf.co/GritLM/gritlm_l7_sq1024_st100_zephfmt} \\
GPT-J 6B & \url{https://hf.co/GritLM/gritlm_g6_sq1024_st100_zephfmt} \\
\midrule
\multicolumn{2}{c}{\textit{\autoref{tab:appembds}}} \\
\midrule
MEDI & \url{https://hf.co/GritLM/emb_m7_sq2048_medi} \\
MEDI2 NNI & \url{https://hf.co/GritLM/emb_m7_sq2048_medi2nni} \\
MEDI2 & \url{https://hf.co/GritLM/emb_m7_sq2048_medi2} \\
MEDI2 + W & \url{https://hf.co/GritLM/emb_m7_sq2048_medi2weights}  \\
\midrule
\multicolumn{2}{c}{\textit{\autoref{tab:appunifiedembds}}} \\
\midrule
MEDI & \url{https://hf.co/GritLM/gritlm_m7_sq2048_medi} \\
MEDI2 & \url{https://hf.co/GritLM/gritlm_m7_sq2048_medi2} \\
BBCC MEDI & \url{https://hf.co/GritLM/gritlm_m7_sq2048_medi_bbcc} \\
BBCC MEDI2 & \url{https://hf.co/GritLM/gritlm_m7_sq2048_medi2_bbcc} \\
BBCC MEDI2BGE & \url{https://hf.co/GritLM/gritlm_m7_sq2048_medi2bge_bbcc} \\
BBCC E5 & \url{https://hf.co/GritLM/gritlm_m7_sq2048_e5_bbcc} \\
\midrule
\multicolumn{2}{c}{\textit{\autoref{tab:appgends}}} \\
\midrule
\tulu{} 2 1 EP & \url{https://hf.co/GritLM/gen_m7_sq2048_tulu2_ep1} \\
\tulu{} 2 2 EP & \url{https://hf.co/GritLM/gen_m7_sq2048_tulu2_ep2} \\
OASST 1 EP & \url{https://hf.co/GritLM/gen_m7_sq2048_oasst_ep1} \\
OASST 2 EP & \url{https://hf.co/GritLM/gen_m7_sq2048_oasst_ep2} \\
UltraChat & \url{https://hf.co/GritLM/gen_m7_sq2048_ultrachat_ep1} \\
\midrule
\multicolumn{2}{c}{\textit{\autoref{tab:appembhead}}} \\
\midrule
No head & \url{https://hf.co/GritLM/gritlm_m7_sq2048_medi_gendups} \\
->~1024 & \url{https://hf.co/GritLM/gritlm_m7_sq2048_medi_proj1024_gendups} \\
\midrule
\multicolumn{2}{c}{\textit{\autoref{tab:appbs}}} \\
\midrule
256 & \url{https://hf.co/GritLM/gritlm_m7_sq2048_medi2} \\
4096 & \url{https://hf.co/GritLM/gritlm_m7_sq2048_medi2_bs4096} \\
\midrule
\multicolumn{2}{c}{\textit{\autoref{tab:appprecision}}} \\
\midrule
BF16 & \url{https://hf.co/GritLM/gritlm_m7_sq2048_e5s_bbcc_bs2048_token6} \\
FP32 & \url{https://hf.co/GritLM/gritlm_m7_sq2048_e5s_bbcc_bs2048_token6_fp32} \\
\midrule
\multicolumn{2}{c}{\textit{\autoref{tab:appibn}}} \\
\midrule
Any dataset & \url{https://hf.co/GritLM/gritlm_m7_sq2048_e5_bbcc_token_anyneg} \\
Same dataset & \url{https://hf.co/GritLM/gritlm_m7_sq2048_e5_bbcc_token} \\
\midrule
\multicolumn{2}{c}{\textit{\autoref{tab:appformat}}} \\
\midrule
\tulu{} 2 & \url{https://hf.co/GritLM/gen_m7_sq2048_tulu2_ep1} \\
Zephyr $\beta$ & \url{https://hf.co/GritLM/gen_m7_sq2048_tulu2_ep1_zephfmt} \\
\midrule
\multicolumn{2}{c}{\textit{\autoref{tab:appsq}}} \\
\midrule
MEDI 2048 & \url{https://hf.co/GritLM/gritlm_m7_sq2048_medi} \\
MEDI 4096 & \url{https://hf.co/GritLM/gritlm_m7_sq4096_medi} \\
BBCC MEDI2 512 & \url{https://hf.co/GritLM/gritlm_m7_sq512_medi2_bbcc} \\
BBCC MEDI2 2048 & \url{https://hf.co/GritLM/gritlm_m7_sq2048_medi2_bbcc} \\
\midrule
\multicolumn{2}{c}{\textit{\autoref{tab:apploss}}} \\
\midrule
E5 Token 4.2 & \url{https://hf.co/GritLM/gritlm_m7_sq2048_e5s_bbcc_bs2048_token4} \\
E5 Token 6.0 & \url{https://hf.co/GritLM/gritlm_m7_sq2048_e5s_bbcc_bs2048_token6} \\
E5 Mix 32~->~8 & \url{https://hf.co/GritLM/GritLM-7b} \\
MEDI2 Mix 4~->~64 & \url{https://hf.co/GritLM/gritlm_m7_sq2048_medi2_bbcc} \\
MEDI2 Mix 32~->~8 & \url{https://hf.co/GritLM/gritlm_m7_sq2048_medi2_bbcc_bs4096} \\
\midrule
\multicolumn{2}{c}{\textit{Other}} \\
\midrule
Code & \url{https://github.com/ContextualAI/gritlm} \\
Logs & \url{https://wandb.ai/muennighoff/gritlm} \\
\tulu{} 2 & \url{https://hf.co/datasets/GritLM/tulu2} \\
MEDI & \url{https://hf.co/datasets/GritLM/medi} \\
MEDI2 & \url{https://hf.co/datasets/GritLM/medi2} \\
MEDI2BGE & \url{https://hf.co/datasets/GritLM/medi2bge} \\    
\modelbase{} NQ Index (\autoref{sec:rag}) & \url{https://hf.co/datasets/GritLM/index} \\
\midrule
\multicolumn{2}{c}{\textit{Main artifacts}} \\
\midrule
\model{} 7B & \url{https://hf.co/GritLM/GritLM-7B} \\
\model{} 8x7B & \url{https://hf.co/GritLM/GritLM-8x7B} \\    
\bottomrule
\label{tab:artifacts}
\end{longtable}

\begin{longtable}{p{3.9cm} p{9.2cm}}
\caption{\textbf{Used artifacts released by others.}} \\
\toprule
Model / Dataset & Public Link \\
\midrule
GPT-4 \citep{openai2023gpt4} & \scriptsize{\url{https://openai.com/gpt-4}} \\
OpenAI v3 \citep{openai2023gpt4} & \scriptsize{\url{https://openai.com/blog/new-embedding-models-and-api-updates}} \\    
Gemini \cite{geminiteam2023gemini} & \scriptsize{\url{https://deepmind.google/technologies/gemini/}} \\
\midrule
\llama{} 2 \citep{touvron2023llama} & \scriptsize{\url{https://hf.co/meta-llama}} \\
Mistral 7B \cite{jiang2023mistral} & \scriptsize{\url{https://hf.co/mistralai/Mistral-7B-v0.1}} \\
Mistral 7B Instruct \cite{jiang2023mistral} & \scriptsize{\url{https://hf.co/mistralai/Mistral-7B-Instruct-v0.1}} \\
Mixtral 8x7B \cite{jiang2024mixtral} & \scriptsize{\url{https://hf.co/mistralai/Mixtral-8x7B-v0.1}} \\
Mixtral 8x7B Instruct \cite{jiang2024mixtral} & \scriptsize{\url{https://hf.co/mistralai/Mixtral-8x7B-Instruct-v0.1}} \\
\tulu{} 2 \citep{ivison2023camels} & \scriptsize{\url{https://hf.co/collections/allenai/tulu-v2-suite-6551b56e743e6349aab45101}} \\
GPT-J 6B \citep{wang2021gpt} & \scriptsize{\url{https://hf.co/EleutherAI/gpt-j-6b}} \\
SGPT BE 5.8B \cite{muennighoff2022sgpt} & \scriptsize{\url{https://hf.co/Muennighoff/SGPT-5.8B-weightedmean-msmarco-specb-bitfit}} \\
Instructor-XL 1.5B \cite{su2023embedder} & \scriptsize{\url{https://hf.co/hkunlp/instructor-xl}} \\
BGE Large 0.34B \cite{xiao2023cpack} & \scriptsize{\url{https://hf.co/BAAI/bge-large-en-v1.5}} \\
Zephyr 7B $\beta$ \cite{tunstall2023zephyr} & \scriptsize{\url{https://hf.co/HuggingFaceH4/zephyr-7b-beta}} \\
E5 Mistral 7B \cite{wang2024improving} & \scriptsize{\url{https://hf.co/intfloat/e5-mistral-7b-instruct}} \\
UltraChat \cite{ding2023enhancing,tunstall2023zephyr} & \scriptsize{\url{https://hf.co/datasets/HuggingFaceH4/ultrachat_200k}} \\
OASST \cite{köpf2023openassistant,muennighoff2023octopack} & \scriptsize{\url{https://hf.co/datasets/bigcode/oasst-octopack}} \\
\bottomrule
\label{tab:usedartifacts}
\end{longtable}

\FloatBarrier

\section{Loss Curves}
\label{sec:loss}

\begin{figure*}[htbp]
\centering
\begin{center}
\includegraphics[width=\textwidth]{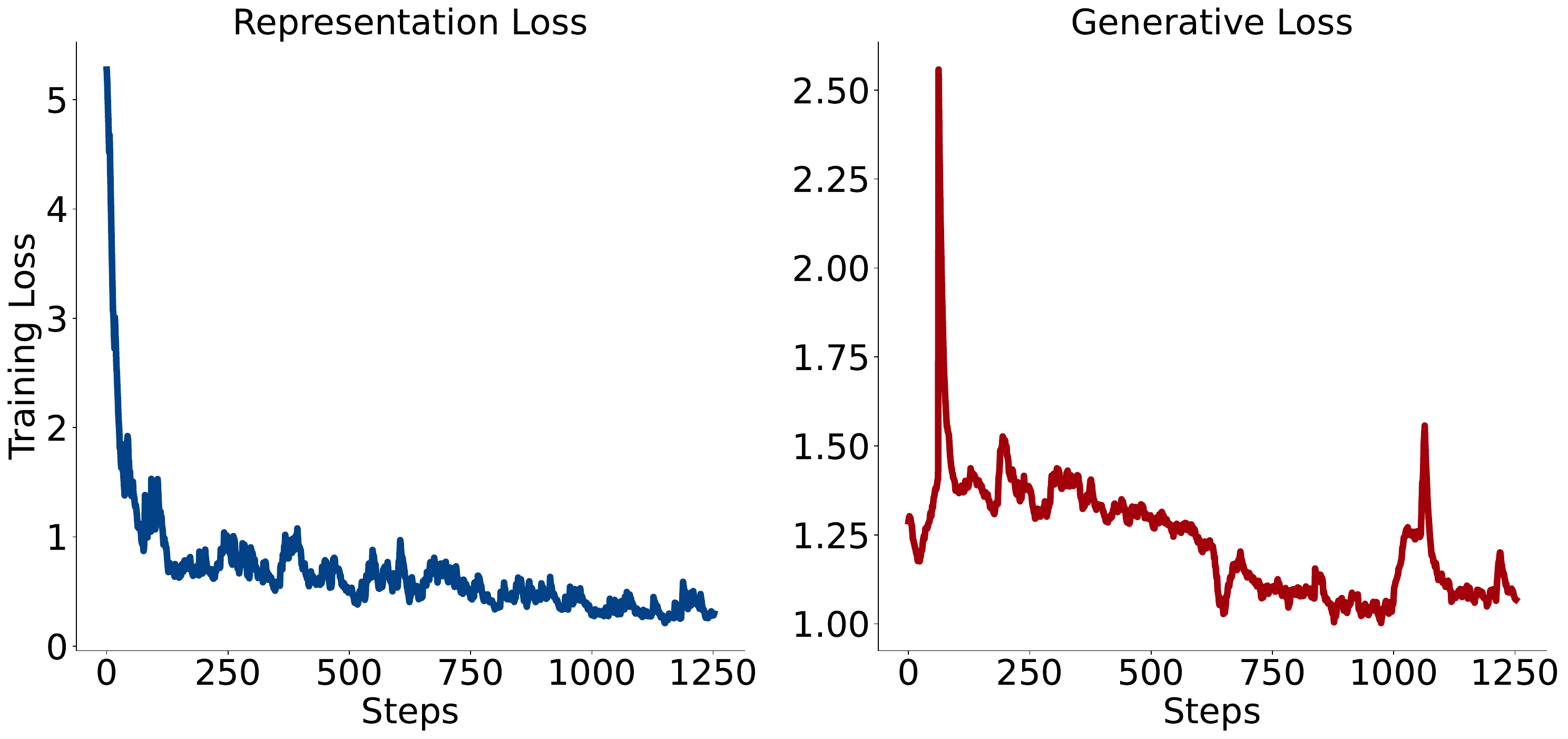}
\caption{\textbf{\modelbase{} training loss smoothed with exponential moving average smoothing and a weight of 0.9.}}
\label{fig:loss7}
\end{center}
\end{figure*}

\begin{figure*}[htbp]
\centering
\begin{center}
\includegraphics[width=\textwidth]{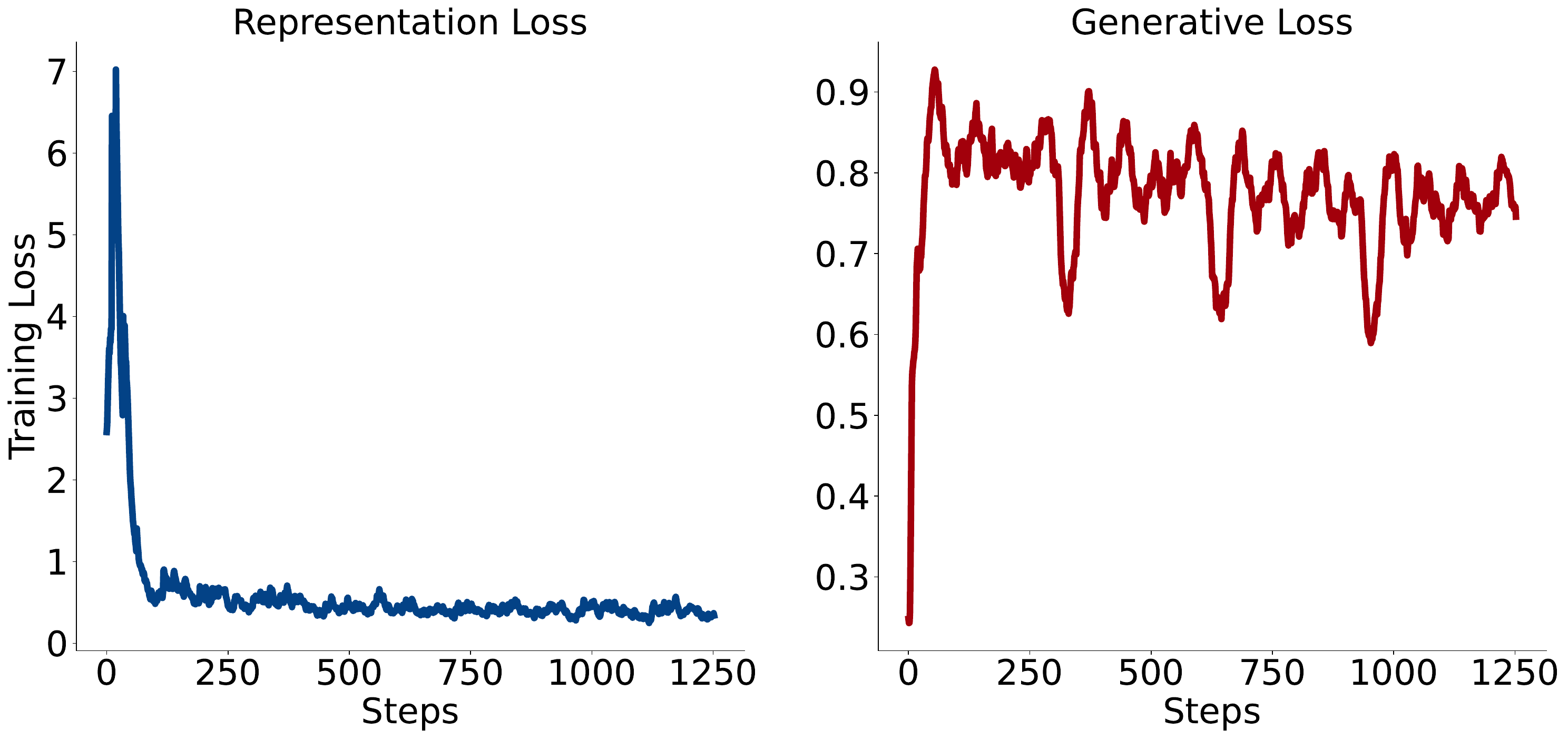}
\caption{\textbf{\modelbig{} training loss smoothed with exponential moving average smoothing and a weight of 0.9.}}
\label{fig:loss8x7}
\end{center}
\end{figure*}

\FloatBarrier

\section{Additional RAG results}
\label{sec:addcaching}

\begin{wraptable}{r}{6cm}
\centering
\caption{\textbf{Additional doc caching results.} We use the same setup as in \autoref{tab:rag} to benchmark doc caching on two additional datasets: TriviaQA~\citep{joshi2017triviaqalargescaledistantly} and MMLU~\citep{hendrycks2022unsolved}.}
\begin{tabular}{l|cc}
\toprule
Dataset ($\rightarrow$) & TriviaQA & MMLU \\
Metric ($\rightarrow$) & \multicolumn{2}{c}{Match (0-shot, $\uparrow$)} \\
\midrule
RAG & 52.12 & 51.10 \\
Doc Caching & \textbf{57.93} & \textbf{53.46} \\
\bottomrule
\end{tabular}
\label{tab:addrag}
\end{wraptable}

In \autoref{sec:rag}, we find that doc caching is the most promising caching variant out of the ones we propose. This is because \textbf{(a)} documents are usually significantly longer than queries, thus caching documents has the highest potential to reduce latency, \textbf{(b)} it maintains performance of regular RAG (\autoref{tab:rag}, and \textbf{(c)} it even works for non-GRIT models though it requires more time to construct the cache for non-GRIT models (\autoref{sec:rag}). Thus we further experiment with doc caching in \autoref{tab:addrag} to verify its performance on other datasets. Similar to Natural Questions in \autoref{tab:rag}, we observe that doc caching maintains performance of regular RAG (even slightly improves) for TriviaQA and MMLU despite the attention mismatch. Note that the attention mismatch problem can always be resolved by simply not using bidirectional attention for the embedding part and thereby guarantee the same performance as not using RAG, however, not using bidirectional attention comes at a slight reduction in embedding performance according to our ablation experiments (\autoref{sec:ablations}).

\begin{wraptable}{r}{6cm}
\centering
\caption{\textbf{Additional RAG results with BGE.} We use the same setup as in \autoref{tab:rag} to benchmark BGE embedding models with the ``Query then document" prompt. The generative model is still \modelbase{}.}
\begin{tabular}{l|c}
\toprule
Dataset ($\rightarrow$) & NQ \\
Metric ($\rightarrow$) & Match (0-shot, $\uparrow$) \\
\midrule
BGE Large 0.34B & 10.39 \\
BGE Base 0.11B & 10.31 \\
BGE Small 0.03B & 10.17 \\
\bottomrule
\end{tabular}
\label{tab:bge}
\end{wraptable}

We also benchmark the BGE series of embedding models~\citep{xiao2023cpack} in \autoref{tab:bge} for RAG. We find performance to be significantly worse than with \model{} in \autoref{tab:rag}. Based on a manual inspection of samples, it appears that the embedding models commonly retrieve irrelevant passages that confuse the generative model. There may be other smaller embedding models or other generative models that may perform better, but overall we expect the RAG performance to be a function of the embedding and generative performance of the individual components (e.g. if an embedding model performs better than \model{}, we would expect it to lead to better RAG performance; BGE generally does not perform better on embedding as shown in \autoref{tab:emb}).

\FloatBarrier

\section{Evaluation}
\label{sec:eval}

For evaluating \model{}, we select the most commonly used embedding and generative benchmarks:

\paragraph{Embedding} To evaluate embedding performance we use the 7 main tasks from MTEB~\cite{muennighoff2023mteb}. \\
\textbf{(1) Classification (CLF):} A logistic regression classifier is trained on embeddings from texts with different labels. The classifier is scored with \textbf{F1}.\\
\textbf{(2)~Clustering (Clust.):} K-means clustering is performed on embeddings from different sources. The agreement of the clusters with respect to the source labels is scored with \textbf{V-measure}.\\
\textbf{(3) Pair Classification (PairCLF):} The cosine similarity of two embeddings with a binary label is computed. The optimal similarity threshold across all samples is found and scored with \textbf{AP} (average precision).\\
\textbf{(4)~Reranking (Rerank)} A query embedding and reference embeddings are compared with cosine similarity. The similarities are scored versus the ground truth ranking of the references via \textbf{MAP} (mean AP).\\
\textbf{(5)~Retrieval:} A query embedding and embeddings of references are compared with cosine similarity. The position of the correct reference(s) in the top ten with the highest cosine similarity is scored with \textbf{nDCG@10} (normalized discounted cumulative gain).\\
\textbf{(6)~STS:} The cosine similarity of two embeddings is compared with a ground truth continuous score of their similarity and scored with \textbf{Spearman} correlation.\\
\textbf{(7)~Summarization (Summ.)} Human-written and machine-written summaries of the same text are embedded. The cosine similarity of the embeddings is compared to human ratings of the machine summaries and scored with \textbf{Spearman} correlation.\\
Among the tasks, Reranking, Retrieval, and Summarization are asymmetric i.e. there are two different kinds of embeddings: queries and documents. Others are symmetric i.e. there is only one kind. We use instructions for every dataset specified in \autoref{sec:promptsemb}. Notably, for some models, we use different instructions for query and document embeddings when dealing with asymmetric tasks. The datasets within each task cover diverse domains ranging from scientific papers to casual conversations.

\paragraph{Generation} For evaluating the generative performance of \model{}, we largely follow the evaluation setup of \tulu{} \cite{wang2023far,ivison2023camels} using open-source frameworks~\cite{eval-harness, bigcode-evaluation-harness}.\\
\textbf{(1)~Multiple-Choice Question Answering via MMLU~\cite{hendrycks2022unsolved}:} Models are tasked to answer knowledge-intensive questions from different fields, such as humanities, social sciences, and hard sciences. No few-shots (FS) are provided and answers are evaluated with \textbf{exact match}.\\
\textbf{(2)~Problem solving via GSM~\cite{cobbe2021training}:} Models are tasked to solve a math problem requiring multi-step reasoning. 8 few-shot (FS) examples with chain-of-thought reasoning (CoT)~\cite{wei2023chainofthought} are provided and \textbf{exact match} is measured.\\
\textbf{(3)~Multilingual Closed-book Question Answering via TyDi QA~\cite{clark2020tydi}:} Models are tasked to answer a question in one of six languages. We evaluate in the Gold Passage and no-context setting following~\citet{anil2023palm}.\\
\textbf{(4) Code Generation via HumanEvalSynthesize~\cite{muennighoff2023octopack,chen2021evaluating}:} We use the HumanEvalSynthesize Python dataset~\cite{muennighoff2023octopack}, which is adapted from HumanEval~\cite{chen2021evaluating} for easy evaluation of instruction-following models. Using the instruction format is different from \citet{ivison2023camels} who use HumanEval without an instruction format which is not how the model is used in practice. Following \citet{muennighoff2023octopack}, we score pass@1 using 20 samples and a temperature of 0.2.\\
\textbf{(5) Boolean Expressions, Causal Judgement, etc. via BBH~\cite{srivastava2023imitation,suzgun2022challenging}} We evaluate a variety of reasoning tasks using BIG-Bench Hard (BBH)~\cite{srivastava2023imitation,suzgun2022challenging}. Similar to GSM8K, 3 FS CoT examples are provided and \textbf{exact match} is measured.\\
\textbf{(6) Open-ended writing, Summarization, Role-playing, etc. via AlpacaEval (Alpaca)~\cite{alpaca_eval,dubois2023alpacafarm}} We evaluate a variety of open-ended generation tasks via the original 1.0 version of AlpacaEval~\cite{alpaca_eval,dubois2023alpacafarm}. GPT-4~\cite{openai2023gpt4} is used to determine the win rate of generations compared to provided GPT-3~\cite{brown2020language} answers. We differ from \citet{ivison2023camels} in that we reduce the maximum token length to 6144 from 8192. We do not use MT-Bench due to its limitations pointed out in \autoref{sec:mtbench}. To ensure reproducibility, we use greedy evaluation throughout.

\FloatBarrier

\section{Ablations Detailed Results}
\label{sec:fullresults}

We display a breakdown of the results from \autoref{tab:ablations} in \autoref{tab:unifiedattn} to \autoref{tab:appformat}. For MTEB per-dataset results, we refer to \autoref{sec:mtebresults}, the MTEB leaderboard (\url{https://hf.co/spaces/mteb/leaderboard}) and our released result files (\url{https://hf.co/datasets/GritLM/results}).

\begin{table}[htbp]
\centering
\caption{\textbf{Unified models attention and pooling ablations.} The sequence of Cs and Bs refers to the attention mechanism for (from left to right): Emb instruction, Emb sample, Gen instruction, Gen sample, where C=Causal, B=Bidirectional, Emb=Embedding and Gen=Generative. WM, LT and M refer to position-weighted mean, last token and mean pooling, respectively.}
\resizebox{\textwidth}{!}{
\begin{tabular}{l|ccccccc|c}
\toprule
Task ($\rightarrow$) & CLF & Clust. & PairCLF & Rerank & Retrieval & STS & Summ. & Avg. \\
Metric ($\rightarrow$) & Acc. & V-Meas. & AP & MAP & nDCG & Spear. & Spear. & \\
Dataset \# ($\rightarrow$) & 12 & 11 & 3 & 4 & 15 & 10 & 1 & 56 \\
\midrule
CCCC WM & 77.9 & 47.9 & 81.5 & 59.0 & 49.4 & 80.3 & 29.4 & 62.8 \\
CCCC LT & 78.8 & 46.9 & 84.5 & 59.6 & 43.9 & 78.7 & 29.3 & 61.2 \\
BBCC M & 79.0 & 48.6 & 86.3 & 59.5 & 49.9 & 81.7 & 30.1 & 63.8 \\
\bottomrule
\end{tabular}
}
\resizebox{\textwidth}{!}{
\begin{tabular}{l|llllll|l}
\toprule
Dataset ($\rightarrow$) & MMLU & GSM8K & BBH & TyDi QA & HumanEval & Alpaca & Avg. \\
Setup ($\rightarrow$) & 0 FS & 8 FS, CoT & 3 FS, CoT & 1 FS, GP & 0 FS & 0 FS, 1.0 & \\
Metric ($\rightarrow$) & EM & EM & EM & F1 & pass@1 & \% Win & \\
\midrule
CCCC WM & 57.5 & 45.0 & 53.1 & 56.0 & 32.3 & 72.9 & 52.8 \\
CCCC LT & 57.2 & 45.5 & 54.7 & 54.0 & 31.1 & 75.7 & 53.0 \\
BBCC M & 57.0 & 46.5 & 54.5 & 55.0 & 30.4 & 73.8 & 52.9 \\
\bottomrule
\end{tabular}
}
\label{tab:unifiedattn}
\end{table}

\begin{table}[htbp]
\centering
\caption{\textbf{Embedding-only models attention and pooling ablations.} The sequence of Cs and Bs refers to the attention mechanism for (from left to right): Emb instruction, Emb sample, where C=Causal, B=Bidirectional and Emb=Embedding. WM and M refer to position-weighted mean and mean pooling, respectively.}
\resizebox{\textwidth}{!}{
\begin{tabular}{l|ccccccc|c}
\toprule
Task ($\rightarrow$) & CLF & Clust. & PairCLF & Rerank & Retrieval & STS & Summ. & Avg. \\
Metric ($\rightarrow$) & Acc. & V-Meas. & AP & MAP & nDCG & Spear. & Spear. & \\
Dataset \# ($\rightarrow$) & 12 & 11 & 3 & 4 & 15 & 10 & 1 & 56 \\
\midrule
CC WM & 77.1 & 44.0 & 83.3 & 57.0 & 43.2 & 79.6 & 29.4 & 60.0 \\
CB M & 76.4 & 45.5 & 83.1 & 56.8 & 45.7 & 80.6 & 30.4 & 61.0 \\
BB M & 77.3 & 46.0 & 83.8 & 58.2 & 46.8 & 81.0 & 32.3 & 61.8 \\
\bottomrule
\end{tabular}
}
\label{tab:embattn}
\end{table}

\begin{table}[htbp]
\centering
\caption{\textbf{Generative-only models attention ablations.} The sequence of Cs and Bs refers to the attention mechanism for (from left to right): Gen instruction, Gen sample, where C=Causal and B=Bidirectional. IL=interleaved, whereby the bidirectional attention is interleaved with causal attention in multi-turn samples (bidirectional for instructions, causal for answers). This allows for faster generation in multi-turn settings as the kv-cache of the answer can be reused.}
\resizebox{\textwidth}{!}{
\begin{tabular}{l|llllll|l}
\toprule
Dataset ($\rightarrow$) & MMLU & GSM8K & BBH & TyDi QA & HumanEval & Alpaca & Avg. \\
Setup ($\rightarrow$) & 0 FS & 8 FS, CoT & 3 FS, CoT & 1 FS, GP & 0 FS & 0 FS, 1.0 & \\
Metric ($\rightarrow$) & EM & EM & EM & F1 & pass@1 & \% Win & \\
\midrule
CC & 57.5 & 52.0 & 55.4 & 56.6 & 34.5 & 75.4 & 55.2 \\
BC & 57.2 & 50.0 & 49.3 & 52.0 & 30.6 & 64.8 & 50.7 \\
BC IL & 52.6 & 41.0 & 46.9 & 45.4 & - & - & - \\
\bottomrule
\end{tabular}
}
\label{tab:genattn}
\end{table}

\begin{table}[htbp]
\centering
\caption{\textbf{Base model ablations.} Models are only trained for 100 steps and with other sub-optimal settings, such as the Zephyr format, that were rectified through later ablations.}
\resizebox{\textwidth}{!}{
\begin{tabular}{l|ccccccc|c}
\toprule
Task ($\rightarrow$) & CLF & Clust. & PairCLF & Rerank & Retrieval & STS & Summ. & Avg. \\
Metric ($\rightarrow$) & Acc. & V-Meas. & AP & MAP & nDCG & Spear. & Spear. & \\
Dataset \# ($\rightarrow$) & 12 & 11 & 3 & 4 & 15 & 10 & 1 & 56 \\
\midrule
Mistral 7B & 70.6 & 43.7 & 74.0 & 54.8 & 35.3 & 72.9 & 31.2 & 54.6 \\
\llama{} 2 7B & 68.1 & 38.0 & 64.1 & 50.2 & 24.2 & 67.7 & 30.5 & 48.2 \\
GPT-J 6B & 70.7 & 41.4 & 69.6 & 53.9 & 29.7 & 70.4 & 29.8 & 51.9 \\
\bottomrule
\end{tabular}
}
\begin{tabular}{l|lllll|l}
\toprule
Dataset ($\rightarrow$) & MMLU & GSM8K & BBH & TyDi QA & HumanEval & Avg. \\
Setup ($\rightarrow$) & 0 FS & 8 FS, CoT & 3 FS, CoT & 1 FS, GP & 0 FS & \\
Metric ($\rightarrow$) & EM & EM & EM & F1 & pass@1 & \\
\midrule
Mistral 7B & 35.0 & 11.0 & 31.6 & 20.5 & 13.8 & 22.4 \\
\llama{} 2 7B & 35.8 & 7.0 & 27.2 & 21.0 & 12.9 & 20.8 \\ 
GPT-J 6B & 27.5 & 3.5 & 22.2 & 8.7 & 8.0 & 14.0 \\
\bottomrule
\end{tabular}
\label{tab:appbasemodel}
\end{table}

\begin{table}[htbp]
\centering
\caption{\textbf{Embedding-only models embedding dataset ablations.} NNI = No Natural Instructions, corresponding to not including natural instructions in the data. II = evaluating with the Instructor-XL instructions~\cite{su2023embedder}. Other models use our new structure with domain, intent, and unit depicted in \autoref{fig:format}. Thus, MEDI2 NNI II and MEDI2 NNI are the same model and only differ in the evaluation instruction set.}
\resizebox{\textwidth}{!}{
\begin{tabular}{l|ccccccc|c}
\toprule
Task ($\rightarrow$) & CLF & Clust. & PairCLF & Rerank & Retrieval & STS & Summ. & Avg. \\
Metric ($\rightarrow$) & Acc. & V-Meas. & AP & MAP & nDCG & Spear. & Spear. & \\
Dataset \# ($\rightarrow$) & 12 & 11 & 3 & 4 & 15 & 10 & 1 & 56 \\
\midrule
MEDI II & 77.1 & 44.0 & 83.3 & 57.0 & 43.2 & 79.6 & 29.4 & 60.0 \\
MEDI2 NNI II & 74.0 & 43.5 & 80.5 & 56.6 & 46.1 & 78.4 & 29.5 & 59.6 \\
MEDI2 NNI & 74.2 & 44.5 & 80.7 & 57.3 & 49.5 & 79.6 & 30.8 & 61.1 \\
MEDI2 & 75.1 & 43.8 & 80.6 & 57.5 & 50.2 & 81.7 & 31.9 & 61.7 \\
MEDI2 + Weights & 74.4 & 42.7 & 78.4 & 57.7 & 50.2 & 81.4 & 30.5 & 61.2 \\
\bottomrule
\end{tabular}
}
\label{tab:appembds}
\end{table}

\begin{table}[htbp]
\centering
\caption{\textbf{Unified models embedding dataset ablations.} The sequence of Cs and Bs refers to the attention mechanism for (from left to right): Emb instruction, Emb sample, where C=Causal, B=Bidirectional, and Emb=Embedding. WM and M refer to position-weighted mean and mean pooling, respectively. MEDI2BGE corresponds to our MEDI2 dataset with negatives coming from the BGE training dataset MTP~\cite{xiao2023cpack}.}
\resizebox{\textwidth}{!}{
\begin{tabular}{l|ccccccc|c}
\toprule
Task ($\rightarrow$) & CLF & Clust. & PairCLF & Rerank & Retrieval & STS & Summ. & Avg. \\
Metric ($\rightarrow$) & Acc. & V-Meas. & AP & MAP & nDCG & Spear. & Spear. & \\
Dataset \# ($\rightarrow$) & 12 & 11 & 3 & 4 & 15 & 10 & 1 & 56 \\
\midrule
CCCC WM MEDI & 77.9 & 47.9 & 81.5 & 59.0 & 49.4 & 80.3 & 29.4 & 62.8 \\
CCCC WM MEDI2 & 76.5 & 47.0 & 82.5 & 59.4 & 51.4 & 81.9 & 30.2 & 63.2 \\
\midrule
BBCC M MEDI & 79.1 & 48.8 & 86.4 & 59.6 & 50.3 & 81.3 & 31.0 & 64.0 \\
BBCC M MEDI2 & 77.0 & 48.7 & 86.0 & 61.0 & 53.6 & 83.0 & 29.1 & 64.7 \\
BBCC M MEDI2BGE & 77.0 & 48.9 & 86.9 & 61.3 & 53.1 & 82.8 & 29.4 & 64.7 \\
BBCC M E5 & 79.7 & 49.5 & 86.2 & 59.6 & 55.3 & 83.6 & 29.9 & 66.0 \\
\bottomrule
\end{tabular}
}
\resizebox{\textwidth}{!}{
\begin{tabular}{l|llllll|l}
\toprule
Dataset ($\rightarrow$) & MMLU & GSM8K & BBH & TyDi QA & HumanEval & Alpaca & Avg. \\
Setup ($\rightarrow$) & 0 FS & 8 FS, CoT & 3 FS, CoT & 1 FS, GP & 0 FS & 0 FS, 1.0 & \\
Metric ($\rightarrow$) & EM & EM & EM & F1 & pass@1 & \% Win & \\
\midrule
CCCC WM MEDI & 57.5 & 45.0 & 53.1 & 56.0 & 32.3 & 72.9 & 52.8 \\
CCCC WM MEDI2 & 57.1 & 49.0 & 53.3 & 55.3 & 32.3 & 73.6 & 53.4 \\
\midrule
BBCC M MEDI & 57.0 & 46.5 & 54.5 & 55.0 & 30.4 & 73.8 & 52.9 \\
BBCC M MEDI2 & 57.0 & 50.5 & 53.8 & 54.7 & 32.3 & 74.7 & 53.8 \\
BBCC M MEDI2BGE & 57.4 & 48.0 & 54.7 & 55.1 & 32.0 & 74.7 & 53.7 \\
BBCC M E5 & 57.3 & 47.5 & 54.2 & 54.6 & 33.6 & 75.4 & 53.8 \\
\bottomrule
\end{tabular}
}
\label{tab:appunifiedembds}
\end{table}

\begin{table}[htbp]
\centering
\caption{\textbf{Generative dataset ablations.} EP = number of epochs.}
\resizebox{\textwidth}{!}{
\begin{tabular}{l|llllll|l}
\toprule
Dataset ($\rightarrow$) & MMLU & GSM8K & BBH & TyDi QA & HumanEval & Alpaca & Avg. \\
Setup ($\rightarrow$) & 0 FS & 8 FS, CoT & 3 FS, CoT & 1 FS, GP & 0 FS & 0 FS, 1.0 & \\
Metric ($\rightarrow$) & EM & EM & EM & F1 & pass@1 & \% Win & \\
\midrule
\tulu{} 2 1 EP & 57.5 & 52.0 & 55.4 & 56.6 & 34.5 & 75.4 & 55.2 \\
\tulu{} 2 2 EP & 58.2 & 53.0 & 51.9 & 54.1 & 37.4 & 80.5 & 55.9 \\
OASST 1 EP & 53.8 & 24.0 & 41.1 & 28.2 & 27.4 & 51.7 & 37.7 \\
OASST 2 EP & 52.4 & 17.5 & 45.7 & 29.2 & 19.8 & 61.3 & 37.7 \\
UltraChat & 56.1 & 43.0 & 53.8 & 35.0 & 25.9 & 70.3 & 47.4 \\
\bottomrule
\end{tabular}
}
\label{tab:appgends}
\end{table}

\begin{table}[htbp]
\centering
\caption{\textbf{Embedding Head.} ``->~1024'' refers to down-projecting the final hidden state with a linear layer from 4096 to 1024 dimensions only for embedding tasks.}
\resizebox{\textwidth}{!}{
\begin{tabular}{l|ccccccc|c}
\toprule
Task ($\rightarrow$) & CLF & Clust. & PairCLF & Rerank & Retrieval & STS & Summ. & Avg. \\
Metric ($\rightarrow$) & Acc. & V-Meas. & AP & MAP & nDCG & Spear. & Spear. & \\
Dataset \# ($\rightarrow$) & 12 & 11 & 3 & 4 & 15 & 10 & 1 & 56 \\
\midrule
No head & 77.7 & 47.9 & 81.3 & 58.6 & 49.2 & 80.4 & 29.5 & 62.7\\
->~1024 & 76.9 & 47.6 & 82.1 & 58.6 & 48.0 & 80.1 & 29.8 & 62.1 \\
\bottomrule
\end{tabular}
}
\resizebox{\textwidth}{!}{
\begin{tabular}{l|llllll|l}
\toprule
Dataset ($\rightarrow$) & MMLU & GSM8K & BBH & TyDi QA & HumanEval & Alpaca & Avg. \\
Setup ($\rightarrow$) & 0 FS & 8 FS, CoT & 3 FS, CoT & 1 FS, GP & 0 FS & 0 FS, 1.0 & \\
Metric ($\rightarrow$) & EM & EM & EM & F1 & pass@1 & \% Win & \\
\midrule
No head & 54.2 & 42.5 & 50.6 & 53.9 & 28.4 & 65.5 & 49.2 \\
->~1024 & 53.6 & 37.0 & 48.8 & 54.4 & 26.6 &  67.3 & 48.0 \\
\bottomrule
\end{tabular}
}
\label{tab:appembhead}
\end{table}

\begin{table}[htbp]
\centering
\caption{\textbf{Embedding batch size ablations.} 256 and 4096 indicate the respective embedding batch size. The generative batch size is always 256.}
\resizebox{\textwidth}{!}{
\begin{tabular}{l|ccccccc|c}
\toprule
Task ($\rightarrow$) & CLF & Clust. & PairCLF & Rerank & Retrieval & STS & Summ. & Avg. \\
Metric ($\rightarrow$) & Acc. & V-Meas. & AP & MAP & nDCG & Spear. & Spear. & \\
Dataset \# ($\rightarrow$) & 12 & 11 & 3 & 4 & 15 & 10 & 1 & 56 \\
\midrule
MEDI2 256 & 76.5 & 47.0 & 82.5 & 59.4 & 51.4 & 81.9 & 30.2 & 63.2 \\
MEDI2 4096 & 77.1 & 48.0 & 84.1 & 60.2 & 52.8 & 82.8 & 30.5 & 64.2 \\
\midrule
\bottomrule
\end{tabular}
}
\resizebox{\textwidth}{!}{
\begin{tabular}{l|llllll|l}
\toprule
Dataset ($\rightarrow$) & MMLU & GSM8K & BBH & TyDi QA & HumanEval & Alpaca & Avg. \\
Setup ($\rightarrow$) & 0 FS & 8 FS, CoT & 3 FS, CoT & 1 FS, GP & 0 FS & 0 FS, 1.0 & \\
Metric ($\rightarrow$) & EM & EM & EM & F1 & pass@1 & \% Win & \\
\midrule
MEDI2 256 & 57.1 & 49.0 & 53.3 & 55.3 & 32.3 & 73.6 & 53.4 \\
MEDI2 4096 & 57.7 & 48.0 & 53.2 & 54.5 & 32.0 & 74.3 & 53.3 \\
\midrule
\bottomrule
\end{tabular}
}
\label{tab:appbs}
\end{table}

\begin{table}[htbp]
\centering
\caption{\textbf{Precision ablations.} BF16 refers to bfloat16 mixed precision and FP32 to float32 precision.}
\resizebox{\textwidth}{!}{
\begin{tabular}{l|ccccccc|c}
\toprule
Task ($\rightarrow$) & CLF & Clust. & PairCLF & Rerank & Retrieval & STS & Summ. & Avg. \\
Metric ($\rightarrow$) & Acc. & V-Meas. & AP & MAP & nDCG & Spear. & Spear. & \\
Dataset \# ($\rightarrow$) & 12 & 11 & 3 & 4 & 15 & 10 & 1 & 56 \\
\midrule
BF16 & 79.7 & 50.2 & 87.6 & 60.2 & 56.5 & 83.4 & 30.8 & 66.5 \\
FP32 & 79.6 & 50.3 & 87.2 & 59.9 & 56.1 & 83.3 & 30.9 & 66.3 \\
\bottomrule
\end{tabular}
}
\resizebox{\textwidth}{!}{
\begin{tabular}{l|llllll|l}
\toprule
Dataset ($\rightarrow$) & MMLU & GSM8K & BBH & TyDi QA & HumanEval & Alpaca & Avg. \\
Setup ($\rightarrow$) & 0 FS & 8 FS, CoT & 3 FS, CoT & 1 FS, GP & 0 FS & 0 FS, 1.0 & \\
Metric ($\rightarrow$) & EM & EM & EM & F1 & pass@1 & \% Win & \\
\midrule
BF16 & 58.2 & 51.5 & 52.8 & 55.9 & 37.3 & 74.4 & 55.0 \\
FP32 & 55.9 & 52.0 & 49.9 & 53.9 & 31.2 & 71.3 & 52.4 \\
\bottomrule
\end{tabular}
}
\label{tab:appprecision}
\end{table}

\begin{table}[htbp]
\centering
\caption{\textbf{In-batch negatives ablations.}}
\resizebox{\textwidth}{!}{
\begin{tabular}{l|ccccccc|c}
\toprule
Task ($\rightarrow$) & CLF & Clust. & PairCLF & Rerank & Retrieval & STS & Summ. & Avg. \\
Metric ($\rightarrow$) & Acc. & V-Meas. & AP & MAP & nDCG & Spear. & Spear. & \\
Dataset \# ($\rightarrow$) & 12 & 11 & 3 & 4 & 15 & 10 & 1 & 56 \\
\midrule
Any dataset & 79.7 & 49.8 & 85.5 & 59.8 & 54.9 & 83.9 & 30.5 & 66.0 \\
Same dataset & 79.5 & 48.9 & 87.4 & 59.0 & 56.2 & 83.0 & 30.5 & 66.0 \\
\bottomrule
\end{tabular}
}
\resizebox{\textwidth}{!}{
\begin{tabular}{l|llllll|l}
\toprule
Dataset ($\rightarrow$) & MMLU & GSM8K & BBH & TyDi QA & HumanEval & Alpaca & Avg. \\
Setup ($\rightarrow$) & 0 FS & 8 FS, CoT & 3 FS, CoT & 1 FS, GP & 0 FS & 0 FS, 1.0 & \\
Metric ($\rightarrow$) & EM & EM & EM & F1 & pass@1 & \% Win & \\
\midrule
Any dataset & 56.1 & 43.5 & 53.1 & 46.6 & 33.5 & 72.3 & 50.9 \\
Same dataset & 55.0 & 45.0 & 54.4 & 49.3 & 29.6 & 73.4 & 51.1 \\
\bottomrule
\end{tabular}
}
\label{tab:appibn}
\end{table}

\begin{table}[htbp]
\centering
\caption{\textbf{Generative format ablations.}}
\resizebox{\textwidth}{!}{
\begin{tabular}{l|llllll|l}
\toprule
Dataset ($\rightarrow$) & MMLU & GSM8K & BBH & TyDi QA & HumanEval & Alpaca & Avg. \\
Setup ($\rightarrow$) & 0 FS & 8 FS, CoT & 3 FS, CoT & 1 FS, GP & 0 FS & 0 FS, 1.0 & \\
Metric ($\rightarrow$) & EM & EM & EM & F1 & pass@1 & \% Win & \\
\midrule
\tulu{} 2 format & 57.5 & 52.0 & 55.4 & 56.6 & 34.5 & 75.4 & 55.2 \\
Zephyr $\beta$ format & 57.3 & 53.5 & 52.7 & 59.1 & 0.0 & 71.2 & 49.0 \\
\bottomrule
\end{tabular}
}
\label{tab:appformat}
\end{table}

\begin{table}[htbp]
\centering
\caption{\textbf{Unified models max tokens ablations.} X:Y refers to ``maximum tokens allowed for embedding documents during training'':``maximum tokens allowed for queries and documents during embedding evaluation''. The sequence of Cs and Bs refers to the attention mechanism for (from left to right): Emb instruction, Emb sample, where C=Causal, B=Bidirectional, and Emb=Embedding.}
\resizebox{\textwidth}{!}{
\begin{tabular}{l|ccccccc|c}
\toprule
Task ($\rightarrow$) & CLF & Clust. & PairCLF & Rerank & Retrieval & STS & Summ. & Avg. \\
Metric ($\rightarrow$) & Acc. & V-Meas. & AP & MAP & nDCG & Spear. & Spear. & \\
Dataset \# ($\rightarrow$) & 12 & 11 & 3 & 4 & 15 & 10 & 1 & 56 \\
\midrule
MEDI 2048:512 & 77.9 & 47.9 & 81.5 & 59.0 & 49.4 & 80.3 & 29.4 & 62.8 \\
MEDI 2048:4096 & 77.9 & 47.9 & 81.5 & 59.0 & 49.4 & 80.2 & 31.3 & 62.8 \\
MEDI 4096:512 & 76.7 & 47.3 & 79.8 & 58.8 & 47.0 & 78.5 & 30.0 & 61.3 \\
MEDI 4096:4096 & 76.8 & 47.2 & 79.8 & 58.8 & 46.9 & 78.2 & 29.9 & 61.3 \\
\midrule
MEDI2 BBCC 2048:512 & 77.0 & 48.7 & 86.0 & 61.0 & 53.6 & 83.0 & 29.1 & 64.7 \\
MEDI2 BBCC 512:512 & 76.9 & 47.6 & 85.5 & 61.0 & 52.8 & 82.3 & 28.8 & 64.1\\
\bottomrule
\end{tabular}
}
\resizebox{\textwidth}{!}{
\begin{tabular}{l|llllll|l}
\toprule
Dataset ($\rightarrow$) & MMLU & GSM8K & BBH & TyDi QA & HumanEval & Alpaca & Avg. \\
Setup ($\rightarrow$) & 0 FS & 8 FS, CoT & 3 FS, CoT & 1 FS, GP & 0 FS & 0 FS, 1.0 & \\
Metric ($\rightarrow$) & EM & EM & EM & F1 & pass@1 & \% Win & \\
\midrule
MEDI 2048:512/4096 & 57.4 & 45.0 & 53.1 & 56.0 & 32.3 & 72.9 & 52.8 \\
MEDI 4096:512/4096 & 53.8 & 43.0 & 52.7 & 54.8 & 30.1 & - & - \\
\midrule
MEDI2 BBCC 2048:512 & 57.0 & 50.5 & 53.8 & 54.7 & 32.3 & 74.7 & 53.8 \\
MEDI2 BBCC 512:512  & 56.9 & 46.5 & 53.1 & 52.6 & 31.2 & 72.8 & 52.2 \\
\bottomrule
\end{tabular}
}
\label{tab:appsq}
\end{table}

\begin{table}[htbp]
\centering
\caption{\textbf{Loss ablations.} E.g. Mix (32~->~8) corresponds to token level loss across 32 samples and then sample level loss across 8 sub-batches for a total batch size of 256. E.g. 2.4 refers to the loss ratio of the 1st step: $\mathcal{L}_{\text{Emb}}/\mathcal{L}_{\text{Gen}}$.}
\resizebox{\textwidth}{!}{
\begin{tabular}{l|ccccccc|c}
\toprule
Task ($\rightarrow$) & CLF & Clust. & PairCLF & Rerank & Retrieval & STS & Summ. & Avg. \\
Metric ($\rightarrow$) & Acc. & V-Meas. & AP & MAP & nDCG & Spear. & Spear. & \\
Dataset \# ($\rightarrow$) & 12 & 11 & 3 & 4 & 15 & 10 & 1 & 56 \\
\midrule
E5S Token 2.4 & 79.5 & 50.1 & 86.5 & 60.0 & 55.6 & 83.2 & 30.3 & 66.1 \\
E5S Token 6.0 & 79.7 & 50.2 & 87.6 & 60.2 & 56.5 & 83.4 & 30.8 & 66.5 \\
E5S Mix (32~->~8) 4.1 & 79.4 & 50.5 & 87.2 & 60.5 & 57.4 & 83.4 & 30.4 & 66.7 \\
\bottomrule
\end{tabular}
}
\resizebox{\textwidth}{!}{
\begin{tabular}{l|llllll|l}
\toprule
Dataset ($\rightarrow$) & MMLU & GSM8K & BBH & TyDi QA & HumanEval & Alpaca & Avg. \\
Setup ($\rightarrow$) & 0 FS & 8 FS, CoT & 3 FS, CoT & 1 FS, GP & 0 FS & 0 FS, 1.0 & \\
Metric ($\rightarrow$) & EM & EM & EM & F1 & pass@1 & \% Win & \\
\midrule
E5S Token 2.4 & 57.9 & 48.5 & 53.5 & 56.5 & 35.2 & 75.0 & 54.4 \\
E5S Token 6.0 & 58.2 & 51.5 & 52.8 & 55.9 & 37.3 & 74.4 & 55.0 \\
E5S Mix (32~->~8) 4.1 & 57.6 & 57.0 & 54.8 & 55.4 & 32.8 & 74.8 & 55.4 \\
\midrule
MEDI2 Mix (4~->~64) 11.7 & 57.0 & 48.0 & 53.7 & 55.0 & 35.8 & 67.6 & 52.9 \\
MEDI2 Mix (32~->~8) 10.2 & 57.0 & 50.5 & 53.8 & 54.7 & 32.3 & 74.7 & 53.8 \\
\bottomrule
\end{tabular}
}
\label{tab:apploss}
\end{table}



\FloatBarrier

\section{\model{} MTEB Full Results}
\label{sec:mtebresults}

\begin{longtable}{l|cccc}
\caption{\textbf{MTEB full results from \autoref{tab:emb}.}} \\
\toprule
\multirow{2}{*}{Dataset} & \multirow{2}{*}{Gen-only} & \multirow{2}{*}{Emb-only} & \multicolumn{2}{c}{\model{}} \\
& & & \textsc{7B} & \textsc{8x7B} \\
\midrule
AmazonCounterfactualClassification & 70.06 & 82.55 & 81.18 & 80.48 \\
AmazonPolarityClassification & 74.74 & 96.19 & 96.52 & 96.32 \\
AmazonReviewsClassification & 38.63 & 57.28 & 57.81 & 57.18 \\
Banking77Classification & 71.25 & 88.73 & 88.47 & 87.46 \\
EmotionClassification & 36.61 & 51.83 & 52.81 & 50.06 \\
ImdbClassification & 73.94 & 94.58 & 95.00 & 94.32 \\
MassiveIntentClassification & 66.82 & 79.37 & 80.78 & 79.72 \\
MassiveScenarioClassification & 71.27 & 81.20 & 82.09 & 81.09 \\
MTOPDomainClassification & 85.40 & 96.72 & 96.16 & 95.29 \\
MTOPIntentClassification & 75.60 & 87.19 & 87.13 & 87.08 \\
ToxicConversationsClassification & 66.36 & 68.37 & 70.80 & 70.89 \\
TweetSentimentExtractionClassification & 54.61 & 61.91 & 64.78 & 62.48 \\
\midrule
ArxivClusteringP2P & 45.40 & 50.87 & 51.67 & 50.72 \\
ArxivClusteringS2S & 29.86 & 47.35 & 48.11 & 48.01 \\
BiorxivClusteringP2P & 33.45 & 40.18 & 40.87 & 41.41 \\
BiorxivClusteringS2S & 23.02 & 39.60 & 39.80 & 38.67 \\
MedrxivClusteringP2P & 27.49 & 36.61 & 36.52 & 36.54 \\
MedrxivClusteringS2S & 23.17 & 37.28 & 36.80 & 37.24 \\
RedditClustering & 23.28 & 63.52 & 61.30 & 63.01 \\
RedditClusteringP2P & 55.00 & 67.81 & 67.26 & 65.86 \\
StackExchangeClustering & 47.14 & 75.53 & 77.33 & 74.41 \\
StackExchangeClusteringP2P & 33.95 & 46.22 & 41.33 & 38.52 \\
TwentyNewsgroupsClustering & 18.15 & 56.8 & 55.70 & 57.16 \\
\midrule
SprintDuplicateQuestions & 51.57 & 93.37 & 93.00 & 91.24 \\
TwitterSemEval2015 & 50.60 & 80.61 & 81.08 & 77.21 \\
TwitterURLCorpus & 60.36 & 87.20 & 87.40 & 86.45 \\
\midrule
AskUbuntuDupQuestions & 49.02 & 68.13 & 67.34 & 65.60 \\
MindSmallReranking & 27.83 & 32.19 & 31.81 & 32.84 \\
SciDocsRR & 56.65 & 87.00 & 86.84 & 86.43 \\
StackOverflowDupQuestions & 38.42 & 55.48 & 55.96 & 54.33 \\
\midrule
ArguAna & 35.96 & 62.95 & 63.24 & 59.49 \\
ClimateFEVER & 8.96 & 31.09 & 30.91 & 28.69 \\
CQADupstackRetrieval & 7.20 & 50.83 & 49.42 & 47.63 \\
DBPedia & 2.15 & 47.06 & 46.60 & 46.54 \\
FEVER & 5.02 & 85.41 & 82.74 & 85.02 \\
FiQA2018 & 6.27 & 60.22 & 59.95 & 49.89 \\
HotpotQA & 6.67 & 79.15 & 79.40 & 73.83 \\
MSMARCO & 0.66 & 41.55 & 41.96 & 35.55 \\
NFCorpus & 3.74 & 41.69 & 40.89 & 39.05 \\
NQ & 2.14 & 69.46 & 70.30 & 63.87 \\
QuoraRetrieval & 64.42 & 89.08 & 89.47 & 87.70 \\
SCIDOCS & 2.32 & 24.86 & 24.41 & 23.06 \\
SciFact & 35.58 & 78.92 & 79.17 & 77.02 \\
Touche2020 & 3.06 & 24.30 & 27.93 & 27.97 \\
TRECCOVID & 20.92 & 75.29 & 74.8 & 81.07 \\
\midrule
BIOSSES & 70.87 & 86.20 & 86.35 & 87.34 \\
SICK-R & 58.95 & 83.03 & 83.13 & 80.56 \\
STS12 & 44.25 & 78.07 & 77.34 & 73.69 \\
STS13 & 64.22 & 85.98 & 85.04 & 85.82 \\
STS14 & 52.24 & 83.92 & 82.91 & 82.05 \\
STS15 & 64.53 & 89.18 & 88.13 & 88.8 \\
STS16 & 65.89 & 86.83 & 86.24 & 86.2 \\
STS17 & 69.64 & 89.7 & 90.13 & 91.46 \\
STS22 & 57.29 & 68.41 & 68.63 & 69.21 \\
STSBenchmark & 53.89 & 86.74 & 85.64 & 87.43 \\
\midrule
SummEval & 21.14 & 30.18 & 30.37 & 29.82 \\
\midrule
\textbf{Average} & \textbf{41.21} & \textbf{66.82} & \textbf{66.76} & \textbf{65.66} \\
\bottomrule
\end{longtable}

\FloatBarrier

\section{Reducing Embedding Training Memory}
\label{sec:embmem}

\begin{figure*}[htbp]
\centering
\begin{center}
\includegraphics[width=\textwidth]{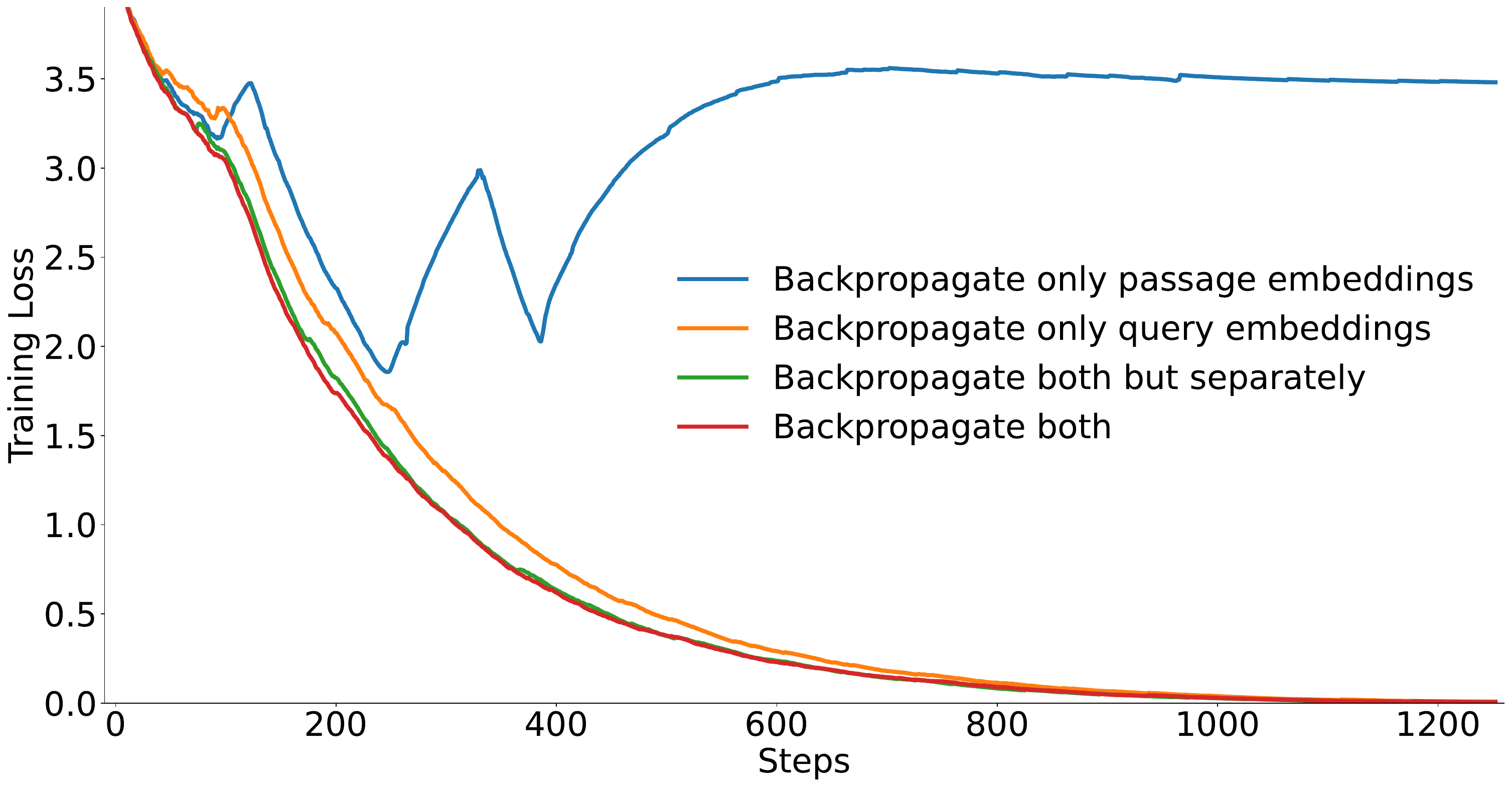}
\caption{\textbf{Embedding memory ablations.} Passage corresponds to both positive and document embeddings. Loss is smoothed with exponential moving average smoothing and a weight of 0.99.}
\label{fig:embmem}
\end{center}
\end{figure*}

\begin{lstlisting}[float,language=Python,
caption={\textbf{Splitting of the embedding pass to save memory, simplified.}\label{alg:embmem}},
basicstyle=\footnotesize\ttfamily,
morekeywords={self},              % Add keywords here
keywordstyle=\color{deepblue},
commentstyle=\color{blue},
emph={multi_chunk_cross_attention, multi_neighbour_cross_attention,cross_attention,relative_positional_encodings},          % Custom highlighting
emphstyle=\color{deepred},    % Custom highlighting style
stringstyle=\color{deepgreen},
frame=none,                         % Any extra options here
showstringspaces=false]

def distributed_contrastive_loss(q, p, n):
# Gather in-batch negatives across devices...
# Compute contrastive loss...

# Split triplet into three forward passes
pos_rep = model(pos)
with torch.no_grad():
q_rep = model(query)
neg_rep = model(neg)

# Only perform backward pass on positive documents
loss = distributed_contrastive_loss(q_rep, pos_rep, neg_rep)
loss.backward()

pos_rep = pos_rep.detach()
# Perform forward + backward on negatives & reuse rest
neg_rep = model(neg)
loss = distributed_contrastive_loss(q_rep, pos_rep, neg_rep)
loss.backward()

# Perform forward + backward on queries & reuse rest
neg_rep = neg_rep.detach()
q_rep = model(query)
loss = distributed_contrastive_loss(q_rep, pos_rep, neg_rep)
loss.backward()

\end{lstlisting}

Generative training only requires sufficient memory to perform a forward and backward pass on a single training sample of a given sequence length. Meanwhile, naive embedding training with in-batch negatives requires sufficient memory to accommodate a forward and a backward pass on $3 * bs$ samples. The $3$ corresponds to the need for passing a triplet of a query, a positive, and a negative document (\autoref{eq:rep}). The batch size ($bs$) factor corresponds to the need for forwarding all samples together as regular gradient accumulation does not work with in-batch negatives. Below we outline the strategies we employ to reduce these memory needs.

\paragraph{Triplet} As the full triplet is only required for loss calculation (\autoref{eq:rep}), it can be split across separate forward and backward passes. To avoid the memory requirements of gradients in PyTorch Autograd~\cite{paszke2019pytorch}, this requires two additional forward passes without gradients. Simplified code representing this procedure is depicted in \autoref{alg:embmem}. In our training, it was sufficient to only split the triplet into two parts: query and passages, where passages consist of both a positive and a negative document. Thus, we only incur the cost of one additional forward pass without gradients on the query. Alternatively, one could only backpropagate on a subset of the embeddings, however, we show in \autoref{fig:embmem} that this leads to worse performance.

\paragraph{In-batch negatives} There are two strategies to reduce the batch size memory requirement to that of a single batch while using nearly unlimited in-batch negatives. \textbf{(1) Distributed Training:} The best strategy is to distribute the training across up to $bs$ GPUs. The representations can then be gathered across GPUs to compute the contrastive loss with in-batch negatives. \textbf{(2) GradCache:} If enough GPUs are not available, GradCache~\cite{gao2021scaling} can be used. GradCache maintains in-batch negatives while allowing computation of gradients for each triplet at a time, thus effectively corresponding to gradient accumulation for contrastive loss. However, it comes at the cost of additional forward passes.

Across training runs, we make use of all three strategies (splitting, distributed training, GradCache).

\FloatBarrier

\section{Hyperparameters}
\label{sec:hps}

We finetune all parameters of our models for up to 1253 steps. Our learning rate is 2e-5, we use 3\% of steps for linear warm-up of the learning rate and decay it linearly to 0 over training. To save memory, we use PyTorch FSDP~\cite{zhao2023pytorch}, gradient checkpointing, BF16 mixed precision training, and strategies outlined in \autoref{sec:embmem}. During training, we use a sequence length of 2048 for generative samples, 256 for embedding queries, and 2048 for embedding documents unless otherwise specified. We finetune using the Adam optimizer~\cite{kingma2017adam} with beta1=0.9 and beta2=0.999 and no weight decay. We also use Flash-Attention 2~\cite{dao2022flashattention,dao2023flashattention} via PyTorch SDPA.

We evaluate models using the settings put forth by the creators of MTEB~\cite{muennighoff2023mteb}, \tulu~\cite{ivison2023camels,wang2024improving} and HumanEvalSynthesize~\cite{muennighoff2023octopack,zhuo2024astraios}. For MTEB, we evaluate using a maximum sequence length of 512 unless otherwise specified.

\FloatBarrier

\section{Embedding Instruction for Generative Models}

As prior instruction-tuned models have been trained without an embedding objective, it is unclear whether one should add an instruction when evaluating them on embedding tasks. We benchmark the Mistral 7B instruct model on MTEB with and without instruction in \autoref{tab:instruct}. We find that performance is around the same, however, adding instructions performs slightly better. Thus, we add an instruction for all instruction-tuned models when benchmarking their embedding performance.

\begin{table}[htbp]
\centering
\caption{\textbf{Benchmarking the benefit of an embedding instruction for generative instruction-tuned models.} When an instruction is used ("Mistral Instruct w/"), we use the default instructions from Instructor XL with the prompt template of the Mistral Instruct model. For no instruction ("Mistral Instruct w/o"), the procedure is the same as for the base model ("Mistral")
}
\resizebox{\textwidth}{!}{
\begin{tabular}{l|ccccccc|c}
\toprule
Task ($\rightarrow$) & CLF & Clust. & PairCLF & Rerank & Retrieval & STS & Summ. & Avg. \\
Metric ($\rightarrow$) & Acc. & V-Meas. & AP & MAP & nDCG & Spear. & Spear. & \\
Dataset \# ($\rightarrow$) & 12 & 11 & 3 & 4 & 15 & 10 & 1 & 56 \\
\midrule
Mistral & 63.5 & 34.6 & 53.5 & 43.2 & 13.2 & 57.4 & 19.7 & 40.5 \\
Mistral Instruct w/o & 65.4 & 35.6 & 60.2 & 44.6 & 16.8 & 61.1 & 25.9 & 43.3 \\
Mistral Instruct w/ & 67.1 & 34.6 & 59.6 & 44.8 & 16.3 & 63.4 & 25.9 & 43.7 \\
\bottomrule
\end{tabular}
\label{tab:instruct}
}
\end{table}

\FloatBarrier

\section{HumanEval Format}

\begin{wraptable}{r}{6cm}
\vspace{-4em}
\caption{\textbf{HumanEvalSynthesize with different formats using \tulu{} 2 7B.}}
\begin{tabular}{c | c c}
\toprule
& \multicolumn{2}{c}{\tulu{} 2 7B} \\
Format & No Chat & Chat \\
\midrule
Pass@1 & 23.4 & 24.5 \\
Pass@10 & 32.4 & 31.3 \\
\bottomrule
\end{tabular}
\label{tab:humaneval}
\vspace{-0.79em}
\end{wraptable}

In \tulu{} 2~\cite{ivison2023camels}, models are evaluated on HumanEval~\cite{chen2021evaluating} without the model's chat format. As this does not reflect the intended usage of the models, we instead use the appropriate chat format for evaluating HumanEval. To do so, we use the instructions and evaluation procedure from HumanEvalSynthesize~\cite{muennighoff2023octopack}. In \autoref{tab:humaneval} we benchmark the impact this has on performance for the \tulu{} 2 7B model~\cite{ivison2023camels}. We find that the performance is around equivalent and thus use the chat format for all evaluations of chat models. For non-chat models, we use the original HumanEval continuation format as proposed by~\citet{chen2021evaluating}

\FloatBarrier

\section{Embedding in FP32 vs BF16}
\label{sec:precision}

We perform all training and evaluations in BF16 (bfloat16) mixed precision to speed up computations. We verified that it performs comparably to FP32 (float32) on MTEB in \autoref{tab:appprecision2}. Note that pooling and subsequent similarity computations are still in FP32.

\begin{table}[htbp]
\centering
\caption{\textbf{Embeddings in FP32 vs BF16.} Benchmarking of the raw Mistral 7B model. ``FP32'' corresponds to doing all computations in float32 precision. ``BF16'' and ``BF16 Cache'' corresponds to doing most operations in bfloat16 except for operations that PyTorch auto casts to float32 (e.g. normalization), pooling and similarity computations. For ``BF16 Cache'', we cast the embeddings after pooling to BF16 and then back to FP32 before similarity computations. This corresponds to locally caching the embeddings in BF16 to save storage and then casting them to FP32 at inference.}
\resizebox{\textwidth}{!}{
\begin{tabular}{l|ccccccc|c}
\toprule
Task ($\rightarrow$) & CLF & Clust. & PairCLF & Rerank & Retrieval & STS & Summ. & Avg. \\
Metric ($\rightarrow$) & Acc. & V-Meas. & AP & MAP & nDCG & Spear. & Spear. & \\
Dataset \# ($\rightarrow$) & 12 & 11 & 3 & 4 & 15 & 10 & 1 & 56 \\
\midrule
FP32 & 63.46 & 34.62 & 53.56 & 43.24 & 13.26 & 57.38 & 19.87 & 40.51 \\
BF16 & 63.47 & 34.60 & 53.52 & 43.24 & 13.24 & 57.38 & 19.68 & 40.50 \\
BF16 Cache & 63.47 & 34.56 & 53.52 & 43.25 & 13.11 & 57.38 & 19.71 & 40.46 \\
\bottomrule
\end{tabular}
\label{tab:appprecision2}
}
\end{table}

\FloatBarrier

\section{Unreliability of MT-Bench}
\label{sec:mtbench}

\begin{wraptable}{r}{6.1cm}
\caption{\textbf{Using GPT-4 vs GPT-4 Turbo as a judge for MT-Bench.} Each evaluator is provided with the same generations of the same instruction-tuned model.}
\begin{tabular}{l|cc|c}
\toprule
& GPT-4 & GPT-4 Turbo & Drop \\
\midrule
Turn 1 & 4.08 & 3.05 & 25\% \\
Turn 2 & 2.64 & 1.88 & 29\% \\
\midrule
Avg. & 3.36 & 2.48 & 26\% \\
\bottomrule
\end{tabular}
\label{tab:mtbench}
\end{wraptable}

We experiment with using MT-Bench with its recommended absolute scores for our generative evaluation~\cite{zheng2023judging}. However, we find that as soon as we switch the LLM Evaluator from GPT-4 to GPT-4 Turbo, the scores change significantly (\autoref{tab:mtbench}). GPT-4 is a closed-source model with changes happening behind the scenes that users may not know about~\cite{chen2023chatgpts}. Thus, if OpenAI decides to change GPT-4, all existing MT-Bench absolute scores would essentially become obsolete. The same applies if the API is retired. To alleviate this, we also experiment with using Zephyr 7B $\beta$~\cite{tunstall2023zephyr} and \llama{} 2 70B Chat~\cite{touvron2023llama} as evaluators, however, we find them to often not provide any rating as they struggle to understand the prompt. While AlpacaEval~\cite{dubois2023alpacafarm,alpaca_eval}, which we use, shares some of these problems, its comparison-based evaluation is more stable. This is because comparing if one generation is better than another generally has an objective ground truth solution. Meanwhile, there is no objective solution as to whether an absolute score of a given generation should be 3 or 4 (MT-Bench has eleven levels from 0-10). This is up to the subjective value system of the evaluator.

\newpage

\section{Dataset Composition}
\label{sec:datasetcomp}

\begin{table}[!htb]
\begin{minipage}{.5\linewidth}
\centering
\caption{\textbf{E5S dataset composition.}}
\begin{tabular}{l|c}
\toprule
Dataset ($\downarrow$) & Num samples \\
\midrule
DuReader \cite{qiu2022dureaderretrieval} & 86,395 \\
ELI5 \cite{fan2019eli5} & 50293 \\
FEVER \cite{thorne2018fever} & 71,257 \\
GPT4 Bitext \cite{wang2024improving} & 89,324 \\
GPT4 P2P \cite{wang2024improving} & 16,842 \\
GPT4 P2S \cite{wang2024improving} & 121,878 \\
GPT4 Retrieval \cite{wang2024improving} & 166,602 \\
GPT4 S2S \cite{wang2024improving} & 13,481 \\
GPT4 STS \cite{wang2024improving} & 98,626 \\
HotpotQA \cite{yang2018hotpotqa} & 68,659 \\
NLI \cite{gao2022simcse} & 275,601 \\
MIRACL \cite{zhang2022making} & 40,203 \\
MSMARCO \cite{bajaj2018ms} & 244,582 \\
MSMARCO Doc \cite{bajaj2018ms} & 71,594 \\
Mr. TyDi \cite{zhang2021mr} & 48,729 \\
NQ \cite{kwiatkowski2019natural} & 71,408 \\
S2ORC \cite{lo2020s2orc} & 80,000 \\
SQuAD \cite{rajpurkar2016squad} & 87,599 \\
T2Ranking \cite{xie2023t2ranking} & 112,335 \\
TriviaQA \cite{karpukhin2020dense} & 60,296 \\
Quora \cite{quora-question-pairs} & 14,926 \\
\midrule
Total & 1,890,630 \\
\bottomrule
\end{tabular}
\end{minipage}
\begin{minipage}{.5\linewidth}

\centering
\caption{\textbf{MEDI2 dataset composition.}}
\begin{tabular}{l|c}
\toprule
MEDI Dataset ($\downarrow$) & Num samples \\
\midrule
AGNews \cite{zhang2016characterlevel} & 199,792 \\
Altlex \cite{hidey-mckeown-2016-identifying} & 112,602 \\
Amazon QA \cite{gupta2019amazonqa} & 199,180 \\
Amazon Review \cite{keung2020multilingual} & 198,298 \\
CC News \cite{Hamborg2017newspleaseA} & 190,503 \\
CNN/Dailymail \cite{fabbri2021summeval} & 189,407 \\
COCO Captions \cite{chen2015microsoft} & 82,783 \\
ELI5 \cite{fan2019eli5} & 196,572 \\
FEVER KILT \cite{thorne2018fever,petroni2021kilt} & 71,257 \\
Flickr 30k \cite{young2014image} & 31,783 \\
Gigaword \cite{rush2015neural,graff2003english} & 200,000 \\
GooAQ \cite{khashabi2021gooaq} & 199,981 \\
HotpotQA KILT \cite{yang2018hotpotqa,petroni2021kilt} & 65,351 \\
NLI \cite{gao2022simcse} & 277,195 \\
MSMARCO \cite{bajaj2018ms} & 491,980 \\
MedMCQA \cite{pal2022medmcqa} & 156,905 \\
Multi-LexSum \cite{shen2022multilexsum} & 2,771 \\
NPR \cite{NPR} & 193,399 \\
NQ \cite{kwiatkowski2019natural} & 73,226 \\
PAQ \cite{lewis2021paq} & 190,162 \\
PubMedQA \cite{jin2019pubmedqa} & 190,481 \\
Reddit \cite{Reddit} & 196,247 \\
S2ORC \cite{lo2020s2orc} & 193,458 \\
SQuAD \cite{rajpurkar2016squad} & 84,105 \\
SciTLDR \cite{cachola2020tldr} & 1,742 \\
SearchQA \cite{DBLP:journals/corr/DunnSHGCC17} & 114,520 \\
Sentence Compression \cite{Filippova2013OvercomingTL} & 179,996 \\
SimpleWiki \cite{Coster2011SimpleEW} & 102,035  \\
StackExchange \cite{StackExchangeDataset} & 201,050 \\
SuperNI (300 datasets) \cite{wang2022supernaturalinstructions} & 2,682,465 \\
SPECTER \cite{cohan2020specter} & 684,000 \\
T-REx KILT \cite{ElSahar2018TRExAL,petroni2021kilt} & 191,383 \\
Quora \cite{quora-question-pairs} & 101,762 \\
WikiAnswers \cite{Fader2014OpenQA} & 200,000 \\
WikiHow \cite{koupaee2018wikihow} & 128,542 \\
XSum \cite{narayan2018dont} & 190,427 \\
Yahoo \cite{zhang2016characterlevel} & 198,346 \\
Zeroshot KILT \cite{levy2017zeroshot,petroni2021kilt} & 124,547 \\
\midrule
Total & 9,084,806 \\
\bottomrule
\end{tabular}
\end{minipage}
\label{tab:dcomp}
\end{table}

\FloatBarrier
\newpage

\section{Dataset Samples}
\label{sec:datasetsamples}

\begin{figure}[htbp]
\hrulefill

\textbf{Query instruction:}

\hrulefill

Represent the sentence for retrieving supporting documents;

\hrulefill

\textbf{Query sample:}

\hrulefill

what two plates form the san andreas fault

\hrulefill

\textbf{Positive instruction:}

\hrulefill

Represent the document for retrieval;

\hrulefill

\textbf{Positive sample:}

\hrulefill

The San Andreas Fault marks the junction between the North American and Pacific Plates. The fault is 1300 km long, extends to at least 25 km in depth, and has a north west-south east trend. It is classified as a right lateral (dextral) strike-slip fault. Loading the player ...

\hrulefill

\textbf{Negative instruction:}

\hrulefill

Represent the document for retrieval;

\hrulefill

\textbf{Negative sample:}

\hrulefill

The San Andreas Fault is the sliding boundary between the Pacific Plate and the North American Plate. It slices California in two from Cape Mendocino to the Mexican border. San Diego, Los Angeles and Big Sur are on the Pacific Plate.

\hrulefill

\caption{\textbf{MEDI sample.}}
\label{fig:medi}
\end{figure}

\begin{figure}[htbp]
\hrulefill

\textbf{Query instruction:}

\hrulefill

Represent this question to retrieve a fitting Wikipedia passage (formal)

\hrulefill

\textbf{Query sample:}

\hrulefill

which two plates meet along the west coast of the USA

\hrulefill

\textbf{Positive instruction:}

\hrulefill

Represent this Wikipedia text in order to get a user query which it answers!

\hrulefill

\textbf{Positive sample:}

\hrulefill

on to a transitional deformation zone in the Chersky Range, then the Ulakhan Fault between it and the Okhotsk Plate, and finally the Aleutian Trench to the end of the Queen Charlotte Fault system.\\
The westerly boundary is the Queen Charlotte Fault running offshore along the coast of Alaska and the Cascadia subduction zone to the north, the San Andreas Fault through California, the East Pacific Rise in the Gulf of California, and the Middle America Trench to the south.\\
On its western edge, the Farallon Plate has been subducting

\hrulefill

\textbf{Negative instruction:}

\hrulefill

Represent this passage to easily find a natural-written user question that can be answered by it.

\hrulefill

\textbf{Negative sample:}

\hrulefill

the continental margin.\\
Types.\\
There are two types of continental margins: \"active\" and \"passive\" margins.\\
Active margins are typically associated with lithospheric plate boundaries. These active margins can be convergent or transform margins, and are also places of high tectonic activity, including volcanoes and earthquakes. The West Coast of North America and South America are active margins. Active continental margins are typically narrow from coast to shelf break, with steep descents into trenches. Convergent active margins occur where oceanic plates meet continental

\hrulefill

\caption{\textbf{MEDI2 sample.}}
\label{fig:medi2}
\end{figure}

\begin{figure}[htbp]
\hrulefill

\textbf{Query instruction:}

\hrulefill

Given a question, retrieve Wikipedia passages that answer the question

\hrulefill

\textbf{Query sample:}

\hrulefill

which two plates meet along the west coast of the USA

\hrulefill

\textbf{Positive sample:}

\hrulefill

North American Plate boundary is the Queen Charlotte Fault running offshore along the coast of Alaska and the Cascadia subduction zone to the north, the San Andreas Fault through California, the East Pacific Rise in the Gulf of California, and the Middle America Trench to the south. On its western edge, the Farallon Plate has been subducting under the North American Plate since the Jurassic Period. The Farallon Plate has almost completely subducted beneath the western portion of the North American Plate leaving that part of the North American Plate in contact with the Pacific Plate as the San Andreas Fault. The Juan

\hrulefill

\textbf{Negative sample:}

\hrulefill

Caribbean Plate Caribbean Plate The Caribbean Plate is a mostly oceanic tectonic plate underlying Central America and the Caribbean Sea off the north coast of South America. Roughly 3.2 million square kilometers (1.2 million square miles) in area, the Caribbean Plate borders the North American Plate, the South American Plate, the Nazca Plate and the Cocos Plate. These borders are regions of intense seismic activity, including frequent earthquakes, occasional tsunamis, and volcanic eruptions. The northern boundary with the North American plate is a transform or strike-slip boundary which runs from the border area of Belize, Guatemala (Motagua Fault), and Honduras in Central

\hrulefill

\caption{\textbf{E5 sample.} The E5 dataset does not use instructions for documents, thus the positive and negative samples do not have instructions.}
\label{fig:e5}
\end{figure}

\begin{figure}[htbp]
\hrulefill

\textbf{Instruction:}

\hrulefill

Q: Lloyd, Mark, and Michael have their Pokemon cards collection. Currently, Mark has thrice as many cards as Lloyd but has 10 fewer cards than Michael. If Michael has 100 cards now, how many more cards should they collect so that all three of them will have a total of 300 cards?

A: 80

Explain how we arrive at this answer: 

\hrulefill

\textbf{Sample:}

\hrulefill

Explanation: Mark has 10 fewer cards than Michael so Mark has 100 cards - 10 cards = 90 cards. So, Lloyd has 90 cards / 3 = 30 cards. All three of them have 90 cards + 30 cards + 100 cards = 220 cards. Thus, they need to collect 300 cards - 220 cards = 80 more cards.

\hrulefill

\caption{\textbf{\tulu{} 2 sample.}}
\label{fig:tulu2}
\end{figure}

\FloatBarrier

\section{Evaluation Prompts}
\label{sec:prompts}

\subsection{Embedding Prompts}
\label{sec:promptsemb}

\autoref{tab:e5instructions} contains the prompt for each MTEB dataset when training on the E5 dataset, which are the same instructions as used in \citet{wang2024improving}. \autoref{tab:medi2instructions} contains the MTEB prompts we use when training on MEDI2, which we wrote ourselves. For models trained on MEDI, we use the instructions for Instructor-XL from~\citet{su2023embedder}.

\setlength\extrarowheight{5pt}
\begin{longtable}{lp{9cm}}
\caption{\textbf{Instructions used for evaluation on the MTEB benchmark when training with the E5 dataset.} ``STS*'' indicates we use the same instructions for all the STS tasks. For retrieval datasets, we do not use an instruction for the document and only display the query instruction.} \\
\toprule
Task Name & Instruction \\
\midrule
AmazonCounterfactualClassif. & Classify a given Amazon customer review text as either counterfactual or not-counterfactual \\
AmazonPolarityClassification & Classify Amazon reviews into positive or negative sentiment \\
AmazonReviewsClassification & Classify the given Amazon review into its appropriate rating category \\
Banking77Classification & Given a online banking query, find the corresponding intents \\
EmotionClassification & Classify the emotion expressed in the given Twitter message into one of the six emotions: anger, fear, joy, love, sadness, and surprise \\
ImdbClassification & Classify the sentiment expressed in the given movie review text from the IMDB dataset \\
MassiveIntentClassification & Given a user utterance as query, find the user intents \\
MassiveScenarioClassification & Given a user utterance as query, find the user scenarios \\
MTOPDomainClassification & Classify the intent domain of the given utterance in task-oriented conversation \\
MTOPIntentClassification & Classify the intent of the given utterance in task-oriented conversation \\
ToxicConversationsClassif. & Classify the given comments as either toxic or not toxic \\
TweetSentimentClassification & Classify the sentiment of a given tweet as either positive, negative, or neutral \\
\midrule
ArxivClusteringP2P & Identify the main and secondary category of Arxiv papers based on the titles and abstracts \\
ArxivClusteringS2S & Identify the main and secondary category of Arxiv papers based on the titles \\
BiorxivClusteringP2P & Identify the main category of Biorxiv papers based on the titles and abstracts \\
BiorxivClusteringS2S & Identify the main category of Biorxiv papers based on the titles \\
MedrxivClusteringP2P & Identify the main category of Medrxiv papers based on the titles and abstracts \\
MedrxivClusteringS2S & Identify the main category of Medrxiv papers based on the titles \\
RedditClustering & Identify the topic or theme of Reddit posts based on the titles \\
RedditClusteringP2P & Identify the topic or theme of Reddit posts based on the titles and posts \\
StackExchangeClustering & Identify the topic or theme of StackExchange posts based on the titles \\
StackExchangeClusteringP2P & Identify the topic or theme of StackExchange posts based on the given paragraphs \\
TwentyNewsgroupsClustering & Identify the topic or theme of the given news articles \\
\midrule
SprintDuplicateQuestions & Retrieve duplicate questions from Sprint forum \\
TwitterSemEval2015 & Retrieve tweets that are semantically similar to the given tweet \\
TwitterURLCorpus & Retrieve tweets that are semantically similar to the given tweet \\
\midrule
AskUbuntuDupQuestions & Retrieve duplicate questions from AskUbuntu forum \\
MindSmallReranking & Retrieve relevant news articles based on user browsing history \\
SciDocsRR & Given a title of a scientific paper, retrieve the titles of other relevant papers \\
StackOverflowDupQuestions & Retrieve duplicate questions from StackOverflow forum \\
\midrule
ArguAna & Given a claim, find documents that refute the claim \\
ClimateFEVER & Given a claim about climate change, retrieve documents that support or refute the claim \\
CQADupstackRetrieval & Given a question, retrieve detailed question descriptions from Stackexchange that are duplicates to the given question \\
DBPedia & Given a query, retrieve relevant entity descriptions from DBPedia \\
FEVER & Given a claim, retrieve documents that support or refute the claim \\
FiQA2018 & Given a financial question, retrieve user replies that best answer the question \\
HotpotQA & Given a multi-hop question, retrieve documents that can help answer the question \\
MSMARCO & Given a web search query, retrieve relevant passages that answer the query \\
NFCorpus & Given a question, retrieve relevant documents that best answer the question \\
NQ & Given a question, retrieve Wikipedia passages that answer the question \\
QuoraRetrieval & Given a question, retrieve questions that are semantically equivalent to the given question \\
SCIDOCS & Given a scientific paper title, retrieve paper abstracts that are cited by the given paper \\
SciFact & Given a scientific claim, retrieve documents that support or refute the claim \\
Touche2020 & Given a question, retrieve detailed and persuasive arguments that answer the question \\
TRECCOVID & Given a query on COVID-19, retrieve documents that answer the query \\
\midrule
STS* & Retrieve semantically similar text. \\
\midrule
SummEval & Given a news summary, retrieve other semantically similar summaries \\
\bottomrule
\label{tab:e5instructions}
\end{longtable}

\setlength\extrarowheight{5pt}
\begin{longtable}{lp{9cm}}
\caption{\textbf{Instructions used for evaluation on the MTEB benchmark when training with the MEDI2 dataset.} For asymmetric datasets, Q refers to instructions for queries, while D refers to document instructions.} \\
\toprule
Task Name & Instruction \\
\midrule
AmazonCounterfactualClassification & Represent the text to find another sentence with the same counterfactuality, e.g. sentences with "would", "wish", etc. should match with other sentences of that kind. \\
AmazonPolarityClassification & Represent the review for finding another Amazon review with the same sentiment (positive / negative) \\
AmazonReviewsClassification & Represent the review for finding another Amazon review with the same rating \\
Banking77Classification & Represent the text for finding another one-sentence banking query with the same intent \\
EmotionClassification & Represent the text for finding another one-sentence text with the same emotion \\
ImdbClassification & Represent the text for finding another one-sentence movie review with the same sentiment \\
MassiveIntentClassification & Represent the text for finding another text of a few words with the same intent \\
MassiveScenarioClassification & Represent the text for finding another text of a few words about the same scenario \\
MTOPDomainClassification & Represent the text for finding another text of a few words about the same domain \\
MTOPIntentClassification & Represent the text for finding another text of a few words with the same intent \\
ToxicConversationsClassification & Represent the text for finding another comment of up to a passage in length with the same level of toxicity (either toxic or not toxic) \\
TweetSentimentExtractionClassification & Represent the tweet for finding another tweet with the same sentiment (positive / neutral / negative) \\
\midrule
ArxivClusteringP2P & Represent the text to find another arXiv title with abstract (concatenated) about the same topic \\
ArxivClusteringS2S & Represent the text to find another arXiv title about the same topic \\
BiorxivClusteringP2P & Represent the text to find another bioRxiv title with abstract (concatenated) about the same topic \\
BiorxivClusteringS2S & Represent the text to find another bioRxiv title about the same topic \\
MedrxivClusteringS2S & Represent the text to find another medRxiv title about the same topic \\
MedrxivClusteringP2P & Represent the text to find another medRxiv title with abstract (concatenated) about the same topic \\
RedditClustering & Represent the text to find another Reddit community title that stems from the same subreddit \\
RedditClusteringP2P & Represent the text to find another Reddit community title with post (concatenated) from the same subreddit \\
StackExchangeClustering & Represent the text to find another StackExchange title that stems from the same StackExchange \\
StackExchangeClusteringP2P & Represent the text to find another StackExchange title with post (concatenated) that stems from the same StackExchange \\
TwentyNewsgroupsClustering & Represent the title to find a similar news title from the same newsgroup \\
\midrule
SprintDuplicateQuestions & Represent the question to be matched with another duplicate user question from the Sprint community forum \\
TwitterSemEval2015 & Represent the tweet to find another tweet that is a paraphrase of it \\
TwitterURLCorpus & Represent the tweet to find another tweet that is a paraphrase of it \\
\midrule
ArguAna Q & Represent the passage to find a passage with a counter-argument about the same topic to it \\
ArguAna D & Represent the passage to find a passage with a counter-argument about the same topic to it \\
ClimateFEVER Q & Represent the climate-based claim to find a Wikipedia abstract to support it \\
ClimateFEVER D & Represent the Wikipedia abstract to find a climate-related claim that it supports \\
CQADupstackAndroidRetrieval Q & Represent the title of a user question to find a duplicate user question title with body from the Android StackExchange forum \\
CQADupstackAndroidRetrieval D & Represent the question title with body posted by a user to find a duplicate user question title from the Android StackExchange forum \\
CQADupstackEnglishRetrieval Q & Represent the title of a user question to find a duplicate user question title with body from the English StackExchange forum \\
CQADupstackEnglishRetrieval D & Represent the question title with body posted by a user to find a duplicate user question title from the English StackExchange forum \\
CQADupstackGamingRetrieval Q & Represent the title of a user question to find a duplicate user question title with body from the Gaming StackExchange forum \\
CQADupstackGamingRetrieval D & Represent the question title with body posted by a user to find a duplicate user question title from the Gaming StackExchange forum \\
CQADupstackGisRetrieval Q & Represent the title of a user question to find a duplicate user question title with body from the Gis StackExchange forum \\
CQADupstackGisRetrieval D & Represent the question title with body posted by a user to find a duplicate user question title from the Gis StackExchange forum \\
CQADupstackMathematicaRetrieval Q & Represent the title of a user question to find a duplicate user question title with body from the Mathematica StackExchange forum \\
CQADupstackMathematicaRetrieval D & Represent the question title with body posted by a user to find a duplicate user question title from the Mathematica StackExchange forum \\
CQADupstackPhysicsRetrieval Q & Represent the title of a user question to find a duplicate user question title with body from the Physics StackExchange forum \\
CQADupstackPhysicsRetrieval D & Represent the question title with body posted by a user to find a duplicate user question title from the Physics StackExchange forum \\
CQADupstackProgrammersRetrieval Q & Represent the title of a user question to find a duplicate user question title with body from the Programmers StackExchange forum \\
CQADupstackProgrammersRetrieval D & Represent the question title with body posted by a user to find a duplicate user question title from the Programmers StackExchange forum \\
CQADupstackStatsRetrieval Q & Represent the title of a user question to find a duplicate user question title with body from the Stats StackExchange forum \\
CQADupstackStatsRetrieval D & Represent the question title with body posted by a user to find a duplicate user question title from the Stats StackExchange forum \\
CQADupstackTexRetrieval Q & Represent the title of a user question to find a duplicate user question title with body from the Tex StackExchange forum \\
CQADupstackTexRetrieval D & Represent the question title with body posted by a user to find a duplicate user question title from the Tex StackExchange forum \\
CQADupstackUnixRetrieval Q & Represent the title of a user question to find a duplicate user question title with body from the Unix StackExchange forum \\
CQADupstackUnixRetrieval D & Represent the question title with body posted by a user to find a duplicate user question title from the Unix StackExchange forum \\
CQADupstackWebmastersRetrieval Q & Represent the title of a user question to find a duplicate user question title with body from the Webmasters StackExchange forum \\
CQADupstackWebmastersRetrieval D & Represent the question title with body posted by a user to find a duplicate user question title from the Webmasters StackExchange forum \\
CQADupstackWordpressRetrieval Q & Represent the title of a user question to find a duplicate user question title with body from the Wordpress StackExchange forum \\
CQADupstackWordpressRetrieval D & Represent the question title with body posted by a user to find a duplicate user question title from the Wordpress StackExchange forum \\
DBPedia Q & Represent the entity to find a title with abstract about this entity from the DBPedia corpus \\
DBPedia D & Represent the title with abstract of a DBPedia corpus entry to find the entity of a few words it is about \\
FEVER Q & Represent the claim to find a Wikipedia abstract to support it \\
FEVER D & Represent the Wikipedia abstract to find a claim that it supports \\
FiQA2018 Q & Represent the StackExchange user query to find a StackExchange post from the Investment topic that answers it \\
FiQA2018 D & Represent the StackExchange post from the Investment topic to find a StackExchange user query that it answers \\
HotpotQA Q & Represent the multi-hop question to find a Wikipedia passage that answers it \\
HotpotQA D & Represent the Wikipedia passage to find a multi-hop question that it answers \\
MSMARCO Q & Represent the Bing user search query to find a passage that adequately addresses it \\
MSMARCO D & Represent the passage for finding a Bing user search query about it \\
NFCorpus Q & Represent the query from NutritionFacts to find a title with text of a medical document from PubMed about it \\
NFCorpus D & Represent this text of a medical document from PubMed to find a query someone may enter at NutritionFacts that it answers \\
NQ Q & Represent the Google search query to find an answer span from a Wikipedia article that addresses it \\
NQ D & Represent the Wikipedia article span to find a Google search query that would be addressed by it \\
SCIDOCS Q & Represent the scientific paper title to find the title with abstract of a scientific paper on PubMed that it has likely cited \\
SCIDOCS D & Represent the title with abstract of this scientific paper to find the title of another scientific paper on PubMed that likely cites this article \\
SciFact Q & Represent the scientific claim to find a scientific paper abstract from PubMed to support it \\
SciFact D & Represent the scientific paper abstract from PubMed to find a scientific claim that it supports \\
TRECCOVID Q & Represent the search query to find a scientific article about COVID-19 that adequately addresses the query \\
TRECCOVID D & Represent the scientific article about COVID-19 to find a user query that it adequately addresses \\
Touche2020 Q & Represent the question to find a title with passage of an argument from args.me that takes a stance about it \\
Touche2020 D & Represent the title with passage of an argument from args.me to find a question that it takes a stance about \\
QuoraRetrieval Q & Represent the Quora question to find another short duplicate question on Quora \\
QuoraRetrieval D & Represent the Quora question to find another short duplicate question on Quora \\
\midrule
AskUbuntuDupQuestions Q & Represent the query to find a duplicate query on the AskUbuntu community forum \\
AskUbuntuDupQuestions D & Represent the query to find a duplicate query on the AskUbuntu community forum \\
MindSmallReranking Q & Represent the news headline to find another news headline that the same reader would enjoy \\
MindSmallReranking D & Represent the news headline to find another news headline that the same reader would enjoy \\
SciDocsRR Q & Represent the title to find a similar scientific paper title \\
SciDocsRR D & Represent the title to find a similar scientific paper title \\
StackOverflowDupQuestions Q & Represent the query to find a duplicate query on the StackOverflow Java/JavaScript/Python community forums \\
StackOverflowDupQuestions D & Represent the query to find a duplicate query on the StackOverflow Java/JavaScript/Python community forums \\
\midrule
BIOSSES & Represent the text to find another biological statement with the same meaning \\
SICK-R & Represent the sentence to find another sentence with the same meaning \\
STS12 & Represent the sentence to find another sentence with the same meaning \\
STS13 & Represent the sentence to find another sentence with the same meaning \\
STS14 & Represent the sentence to find another sentence with the same meaning \\
STS15 & Represent the sentence to find another sentence with the same meaning \\
STS16 & Represent the sentence to find another sentence with the same meaning \\
STS17 & Represent the sentence to find another sentence with the same meaning \\
STS22 & Represent the sentence to find another sentence with the same meaning \\
STSBenchmark & Represent the sentence to find another sentence with the same meaning \\
\midrule
SummEval Q & Represent the human-written summary to find a high-quality machine-written summary of the same news article \\
SummEval D & Represent the machine-written summary to find a human-written summary with similar quality of the same news article \\
\bottomrule
\label{tab:medi2instructions}
\end{longtable}

\subsection{Embedding Few-Shot Prompts}
\label{sec:promptsembfewshot}

\setlength\extrarowheight{5pt}
\begin{longtable}{lp{9cm}}
\caption{\textbf{1-shot example for the model trained on E5S.} The example is appended to the respective instruction in \autoref{tab:e5instructions} separated by two newlines.}\\
\toprule
Task Name & Instruction \\
\midrule
Banking77Classification & For example given "I am still waiting on my card?", it would match with "card\_arrival"\\
EmotionClassification & For example given "ive been feeling a little burdened lately wasnt sure why that was", it would match with "sadness"\\
ImdbClassification & For example given "If only to avoid making this type of film in the future. This film is interesting as an experiment but tells no cogent story.<br /><br />One might feel virtuous for sitting thru it because it touches on so many IMPORTANT issues but it does so without any discernable motive. The viewer comes away with no new perspectives (unless one comes up with one while one's mind wanders, as it will invariably do during this pointless film).<br /><br />One might better spend one's time staring out a window at a tree growing.<br /><br />", it would match with "negative"\\
\midrule
BiorxivClusteringP2P & For example given "Association of CDH11 with ASD revealed by matched-gene co-expression analysis and mouse behavioral studies", it would match with "neuroscience" \\
\midrule
TwitterSemEval2015 & For example given "The Ending to 8 Mile is my fav part of the whole movie", it would match with "Those last 3 battles in 8 Mile are THE shit" \\
TwitterURLCorpus & For example given "Liberals , dont let Donald Trump tarnish L.L. Beans sterling brand reputation ", it would match with "Liberals, Don\&rsquo;t Let Donald Trump Tarnish L.L. Bean\&rsquo;s Sterling Brand Reputation" \\
SprintDuplicateQuestions & For example given "Why is it impossible for me to find a easy way to send a picture with text on my Kyocera DuraCore ?", it would match with "Send or receive a picture with text - Kyocera DuraCore" \\
\midrule
AskUbuntuDupQuestions & For example given "what is a short cut i can use to switch applications ?", you should retrieve "keyboard short cut for switching between two or more instances of the same application ?" \\
\midrule
ArguAna & For example given "People will die if we don’t do animal testing Every year, 23 new drugs are introduced in the UK alone.[13] Almost all will be tested on animals. A new drug will be used for a long time. Think of all the people saved by the use of penicillin. If drugs cost more to test, that means drug companies will develop less. This means more people suffering and dying", you should retrieve "animals science science general ban animal testing junior Many of these drugs are “me too” drugs – ones with a slight change that doesn’t make much difference to an existing drug. [14] So often the benefits from animal testing are marginal, and even if there was a slight increase in human suffering, it would be worth it based on the animal suffering saved."\\
SCIDOCS & For example given "A Direct Search Method to solve Economic Dispatch Problem with Valve-Point Effect", you should retrieve "A Hybrid EP and SQP for Dynamic Economic Dispatch with Nonsmooth Fuel Cost Function Dynamic economic dispatch (DED) is one of the main functions of power generation operation and control. It determines the optimal settings of generator units with predicted load demand over a certain period of time. The objective is to operate an electric power system most economically while the system is operating within its security limits. This paper proposes a new hybrid methodology for solving DED. The proposed method is developed in such a way that a simple evolutionary programming (EP) is applied as a based level search, which can give a good direction to the optimal global region, and a local search sequential quadratic programming (SQP) is used as a fine tuning to determine the optimal solution at the final. Ten units test system with nonsmooth fuel cost function is used to illustrate the effectiveness of the proposed method compared with those obtained from EP and SQP alone."\\
\midrule
STS12 & For example given "Counties with population declines will be Vermillion, Posey and Madison.", it would match with "Vermillion, Posey and Madison County populations will decline."\\
\midrule
SummEval & The provided query could be "Mexican restaurant has decided to tap into \$70 billion food delivery market. Fast-casual chain will work with the Postmates app to allow mobile orders. App works in similar way to Uber, using hired drivers to deliver the food. But the chain will add a 9\% service charge - on top of Postmates\' \$5 rate." and the positive "chipotle has decided to tap into the \$ 70 billion food delivery market by teaming up with an app to bring burritos straight to customers \' doors . the fast-casual chain will work with the postmates app to begin offering delivery for online and mobile orders in 67 cities . the restaurant plans to add a nine per cent service charge - with the delivery fees for postmates beginning at \$ 5 and up depending on distance and demand ."\\
\bottomrule
\label{tab:e5fewshots}
\end{longtable}

\setlength\extrarowheight{5pt}
\begin{longtable}{lp{9cm}}
\caption{\textbf{1-shot example for the model trained on MEDI2.} The example is appended to the respective instruction in \autoref{tab:medi2instructions} separated by two newlines.}\\
\toprule
Task Name & Instruction \\
\midrule
Banking77Classification & The provided query could be "I am still waiting on my card?" and the positive "What can I
do if my card still hasn't arrived after 2 weeks?"\\
EmotionClassification & The provided query could be "ive been feeling a little burdened lately wasnt sure why that was" and the positive "i feel like i have to make the suffering i m seeing mean something"\\
ImdbClassification & The provided query could be "If only to avoid making this type of film in the future. This film is interesting as an experiment but tells no cogent story.<br /><br />One might feel virtuous for sitting thru it because it touches on so many IMPORTANT issues but it does so without any discernable motive. The viewer comes away with no new perspectives (unless one comes up with one while one's mind wanders, as it will invariably do during this pointless film).<br /><br />One might better spend one's time staring out a window at a tree growing.<br /><br />" and the positive "The silent one-panel cartoon Henry comes to Fleischer Studios, billed as "The world's funniest human" in this dull little cartoon. Betty, long past her prime, thanks to the Production Code, is running a pet shop and leaves Henry in charge for far too long -- five minutes. A bore." \\
\midrule
SprintDuplicateQuestions & The provided query could be "Why is it impossible for me to find a easy way to send a picture with text on my Kyocera DuraCore ?" and the positive "Send or receive a picture with text - Kyocera DuraCore" \\
TwitterSemEval2015 & For example given "The Ending to 8 Mile is my fav part of the whole movie", it would match with "Those last 3 battles in 8 Mile are THE shit" \\
TwitterURLCorpus & For example given "Liberals , dont let Donald Trump tarnish L.L. Beans sterling brand reputation ", it would match with "Liberals, Don\&rsquo;t Let Donald Trump Tarnish L.L. Bean\&rsquo;s Sterling Brand Reputation" \\
\midrule
AskUbuntuDupQuestions & The provided query could be "what is a short cut i can use to switch applications ?" and the positive "keyboard short cut for switching between two or more instances of the same application ?" \\
ArguAna & The provided query could be "People will die if we don’t do animal testing Every year, 23 new drugs are introduced in the UK alone.[13] Almost all will be tested on animals. A new drug will be used for a long time. Think of all the people saved by the use of penicillin. If drugs cost more to test, that means drug companies will develop less. This means more people suffering and dying" and the positive "animals science science general ban animal testing junior Many of these drugs are “me too” drugs – ones with a slight change that doesn’t make much difference to an existing drug. [14] So often the benefits from animal testing are marginal, and even if there was a slight increase in human suffering, it would be worth it based on the animal suffering saved." \\
SCIDOCS & The provided query could be "A Direct Search Method to solve Economic Dispatch Problem with Valve-Point Effect" and the positive "A Hybrid EP and SQP for Dynamic Economic Dispatch with Nonsmooth Fuel Cost Function Dynamic economic dispatch (DED) is one of the main functions of power generation operation and control. It determines the optimal settings of generator units with predicted load demand over a certain period of time. The objective is to operate an electric power system most economically while the system is operating within its security limits. This paper proposes a new hybrid methodology for solving DED. The proposed method is developed in such a way that a simple evolutionary programming (EP) is applied as a based level search, which can give a good direction to the optimal global region, and a local search sequential quadratic programming (SQP) is used as a fine tuning to determine the optimal solution at the final. Ten units test system with nonsmooth fuel cost function is used to illustrate the effectiveness of the proposed method compared with those obtained from EP and SQP alone." \\
\midrule
STS12 & The provided query could be "Counties with population declines will be Vermillion, Posey and Madison." and the positive "Vermillion, Posey and Madison County populations will decline." \\
\midrule
SummEval & The provided query could be "Mexican restaurant has decided to tap into \$70 billion food delivery market. Fast-casual chain will work with the Postmates app to allow mobile orders. App works in similar way to Uber, using hired drivers to deliver the food. But the chain will add a 9\% service charge - on top of Postmates\' \$5 rate." and the positive "chipotle has decided to tap into the \$ 70 billion food delivery market by teaming up with an app to bring burritos straight to customers \' doors . the fast-casual chain will work with the postmates app to begin offering delivery for online and mobile orders in 67 cities . the restaurant plans to add a nine per cent service charge - with the delivery fees for postmates beginning at \$ 5 and up depending on distance and demand ." \\
\bottomrule
\label{tab:medi2fewshots}
\end{longtable}

\subsection{Generative Prompts}
\label{sec:promptsgen}

\autoref{fig:mmlu} until \autoref{fig:alpacaeval} contain the prompts with examples used for our generative tasks.

\begin{figure}[htbp]
\hrulefill

\textbf{Input:}

\hrulefill

<s><|user|>

The following are multiple choice questions (with answers) about  abstract algebra. \\

Find the degree for the given field extension Q(sqrt(2), sqrt(3), sqrt(18)) over Q.

A. 0

B. 4

C. 2

D. 6

Answer:

<|assistant|>

The answer is:

\hrulefill

\textbf{Correct completion:}

\hrulefill

~B

\hrulefill

\caption{\textbf{MMLU prompt example.}}
\label{fig:mmlu}
\end{figure}

\begin{figure}[htbp]
\hrulefill

\textbf{Input:}

\hrulefill

<s><|user|>\\
Answer the following questions.\\
Question: There are 15 trees in the grove. Grove workers will plant trees in the grove today. After they are done, there will be 21 trees. How many trees did the grove workers plant today?\\
Answer: There are 15 trees originally. Then there were 21 trees after some more were planted. So there must have been 21 - 15 = 6. So the answer is 6.\\

Question: If there are 3 cars in the parking lot and 2 more cars arrive, how many cars are in the parking lot?\\
Answer: There are originally 3 cars. 2 more cars arrive. 3 + 2 = 5. So the answer is 5.\\

Question: Leah had 32 chocolates and her sister had 42. If they ate 35, how many pieces do they have left in total?\\
Answer: Originally, Leah had 32 chocolates. Her sister had 42. So in total they had 32 + 42 = 74. After eating 35, they had 74 - 35 = 39. So the answer is 39.\\

Question: Jason had 20 lollipops. He gave Denny some lollipops. Now Jason has 12 lollipops. How many lollipops did Jason give to Denny?\\
Answer: Jason started with 20 lollipops. Then he had 12 after giving some to Denny. So he gave Denny 20 - 12 = 8. So the answer is 8.\\

Question: Shawn has five toys. For Christmas, he got two toys each from his mom and dad. How many toys does he have now?\\
Answer: Shawn started with 5 toys. If he got 2 toys each from his mom and dad, then that is 4 more toys. 5 + 4 = 9. So the answer is 9.\\

Question: There were nine computers in the server room. Five more computers were installed each day, from monday to thursday. How many computers are now in the server room?\\
Answer: There were originally 9 computers. For each of 4 days, 5 more computers were added. So 5 * 4 = 20 computers were added. 9 + 20 is 29. So the answer is 29.\\

Question: Michael had 58 golf balls. On tuesday, he lost 23 golf balls. On wednesday, he lost 2 more. How many golf balls did he have at the end of wednesday?\\
Answer: Michael started with 58 golf balls. After losing 23 on tuesday, he had 58 - 23 = 35. After losing 2 more, he had 35 - 2 = 33 golf balls. So the answer is 33.\\

Question: Olivia has \$23. She bought five bagels for \$3 each. How much money does she have left?\\
Answer: Olivia had 23 dollars. 5 bagels for 3 dollars each will be 5 x 3 = 15 dollars. So she has 23 - 15 dollars left. 23 - 15 is 8. So the answer is 8.\\

Question: The girls are trying to raise money for a carnival. Kim raises \$320 more than Alexandra, who raises \$430, and Maryam raises \$400 more than Sarah, who raises \$300. How much money, in dollars, did they all raise in total?\\
<|assistant|>\\
Answer:

\hrulefill

\textbf{Correct completion:}

\hrulefill

~Kim raises 320+430=750 dollars. Maryam raises 400+300=700 dollars. They raise 750+430+400+700=2280 dollars. So the answer is 2280.

\hrulefill

\caption{\textbf{GSM8K prompt example.}}
\label{fig:gsm}
\end{figure}

\begin{figure}[htbp]
\hrulefill

\textbf{Input:}

\hrulefill

<s><|user|>\\
Questions that involve enumerating objects and asking the model to count them.\\\\  
Q: I have a blackberry, a clarinet, a nectarine, a plum, a strawberry, a banana, a flute, an orange, and a violin. How many fruits do I have?\\
A: Let's think step by step.\\
We first identify the fruits on the list and include their quantity in parentheses:\\
- blackberry (1)\\
- nectarine (1)\\
- plum (1)\\
- strawberry (1)\\
- banana (1)\\
- orange (1)\\
Now, let's add the numbers in parentheses: 1 + 1 + 1 + 1 + 1 + 1 = 6. So the answer is 6.\\

Q: I have an orange, a raspberry, two peaches, a blackberry, an apple, a grape, a nectarine, and three plums. How many fruits do I have?\\
A: Let's think step by step.\\
We first identify the fruits on the list and include their quantity in parentheses:\\
- orange (1)\\
- raspberry (1)\\
- peaches (2)\\
- blackberry (1)\\
- apple (1)\\
- grape (1)\\
- nectarine (1)\\
- plums (3)\\
Now, let's add the numbers in parentheses: 1 + 1 + 2 + 1 + 1 + 1 + 1 + 3 = 11. So the answer is 11.\\

Q: I have a lettuce head, a head of broccoli, an onion, a stalk of celery, two carrots, a garlic, and a yam. How many vegetables do I have?\\
A: Let's think step by step.\\
We first identify the vegetables on the list and include their quantity in parentheses:\\
- lettuce (1)\\
- broccoli (1)\\
- onion (1)\\
- celery (1)\\
- carrots (2)\\
- garlic (1)\\
- yam (1)\\
Now, let's add the numbers in parentheses: 1 + 1 + 1 + 1 + 2 + 1 + 1 = 8. So the answer is 8.\\

Q: I have a banana, four strawberries, an apple, two peaches, a plum, a blackberry, and two raspberries. How many fruits do I have?\\
<|assistant|>

\hrulefill

\textbf{Correct completion:}

\hrulefill

~12

\hrulefill

\caption{\textbf{BBH prompt example.}}
\label{fig:bbh}
\end{figure}

\begin{figure}[htbp]
\hrulefill

\textbf{Input:}

\hrulefill

<s><|user|>\\
Jawab pertanyaan berikut berdasarkan informasi di bagian yang diberikan.\\

Bagian: Mula-mula pada pelukis seorang pelukis pemandangan Wahdi Sumanta, Abdullah Suriosubroto (ayah Basuki Abdullah). Kemudian bertemu dan berkenalan dengan Affandi, Sudarso, dan Barli. Mereka lalu membentuk kelompok Lima serangkai. Di rumah tempat tinggal Affandi mereka mengadakan latihan melukis bersama dengan tekun dan mendalam. Dari Wahdi, ia banyak menggali pengetahuan tentang melukis. Kegiatannya bukan hanya melukis semata, tetapi pada waktu senggang ia menceburkan diri pada kelompok sandiwara Sunda sebagai pelukis dekor. Dari pengalaman itulah, ia mengasah kemampuannya.\\
Pertanyaan: dari manakah Hendra Gunawan belajar melukis?\\
Jawaban: kelompok Lima serangkai\\

Bagian: Empat Sehat Lima Sempurna adalah kampanye yang dilakukan pemerintah sejak tahun 1955 untuk membuat masyarakat memahami pola makan yang benar.[1]. Dalam konsep 4 sehat 5 sempurna, makanan dibagi atas empat sumber nutrisi penting, yaitu makanan pokok, lauk pauk, sayur-mayur, buah-buahan, dan disempurnakan dengan susu bila mampu, menjadi lima sempurna[2] Konsep ini menekankan pentingnya empat golongan makanan berupa sumber kalori untuk tenaga, protein untuk pembangun, sayur dan buah sumber vitamin dan mineral untuk pemeliharaan.[1]\\
Pertanyaan: siapakah yang mencptakan Ide 4 sehat 5 sempurna pertama kali?\\

<|assistant|>\\
Jawaban:

\hrulefill

\textbf{Correct completion:}

\hrulefill

~pemerintah

\hrulefill

\caption{\textbf{TyDi QA prompt example from Indonesian.}}
\label{fig:tydiqa}
\end{figure}

\begin{figure}[htbp]
\hrulefill

\textbf{Input:}

\hrulefill

<|user|>

Write a Python function `has\_close\_elements(numbers: List[float], threshold: float) ->
bool` to solve the following problem:
Check if in given list of numbers, are any two numbers closer to each other than       
given threshold.
\begin{lstlisting}[language=Python,
basicstyle=\footnotesize\ttfamily,
morekeywords={self},              % Add keywords here
keywordstyle=\color{deepblue},
commentstyle=\color{blue},
emph={multi_chunk_cross_attention, multi_neighbour_cross_attention,cross_attention,relative_positional_encodings},          % Custom highlighting
emphstyle=\color{deepred},    % Custom highlighting style
stringstyle=\color{deepgreen},
frame=none,                         % Any extra options here
showstringspaces=false]
>>> has_close_elements([1.0, 2.0, 3.0], 0.5)
False
>>> has_close_elements([1.0, 2.8, 3.0, 4.0, 5.0, 2.0], 0.3)
True
\end{lstlisting}
<|assistant|>
\begin{lstlisting}[language=Python,
basicstyle=\footnotesize\ttfamily,
morekeywords={self},              % Add keywords here
keywordstyle=\color{deepblue},
commentstyle=\color{blue},
emph={multi_chunk_cross_attention, multi_neighbour_cross_attention,cross_attention,relative_positional_encodings},          % Custom highlighting
emphstyle=\color{deepred},    % Custom highlighting style
stringstyle=\color{deepgreen},
frame=none,                         % Any extra options here
showstringspaces=false]
from typing import List
def has_close_elements(numbers: List[float], threshold: float) -> bool:
""" Check if in given list of numbers, are any two numbers closer to each other than
given threshold.
>>> has_close_elements([1.0, 2.0, 3.0], 0.5)
False
>>> has_close_elements([1.0, 2.8, 3.0, 4.0, 5.0, 2.0], 0.3)
True
"""
\end{lstlisting}

\hrulefill

\textbf{Correct completion:}

\hrulefill

\begin{lstlisting}[language=Python,
basicstyle=\footnotesize\ttfamily,
morekeywords={self},              % Add keywords here
keywordstyle=\color{deepblue},
commentstyle=\color{blue},
emph={multi_chunk_cross_attention, multi_neighbour_cross_attention,cross_attention,relative_positional_encodings},          % Custom highlighting
emphstyle=\color{deepred},    % Custom highlighting style
stringstyle=\color{deepgreen},
frame=none,                         % Any extra options here
showstringspaces=false]
for idx, elem in enumerate(numbers):
for idx2, elem2 in enumerate(numbers):
    if idx != idx2:
        distance = abs(elem - elem2)
        if distance < threshold:
            return True

return False
\end{lstlisting}

\hrulefill

\caption{\textbf{HumanEvalSynthesize prompt example.}}
\label{fig:humaneval}
\end{figure}

\begin{figure}[htbp]
\hrulefill

\textbf{Input:}

\hrulefill

<s><|user|>\\
If you were a Shakespearean character, how would you declare your love for someone in a soliloquy?

\hrulefill

\textbf{Completion by GPT-4 Turbo:}

\hrulefill

As an AI, I lack personal emotions and experiences, but I can craft a soliloquy in the style of a Shakespearean character declaring love. Here's an example:\\
\\
(Imagine a dimly lit stage, a single spotlight shining down upon the forlorn lover, who steps forward to share the depths of their heart.)\\
\\
O, speak, thou lustrous beacon of the night,\\
With stars enkindled in thine eyes so bright.\\
Thou art the daybreak to my endless gloom,\\
The rose that in my barren heart doth bloom.\\
\\
Forsooth, this love, a tempest full of grace,\\
Hath swept upon my soul without a trace.\\
My heart, once caged, now beats beneath thy gaze,\\
And yearns to laud thy beauty all its days.\\
\\
What art thou, Love? A tyrant or a muse?\\
That rends my sense and all my thoughts confuse?\\
Thou art the riddle none can e'er resolve,\\
Yet in thy puzzle, my life doth revolve.\\
\\
With every breath, I thee do magnify,\\
And to the world, my passion testify.\\
Thine essence hath become my sole refrain,\\
A melody that in my mind doth reign.\\
\\
Shall I compare thee to a summer's day?\\
Thou art more lovely and more temperate.\\
Rough winds do shake the darling buds of May,\\
And summer's lease hath all too short a date.\\
\\
But thy eternal summer shall not fade,\\
Nor lose possession of that fair thou ow'st.\\
Nor shall Death brag thou wander'st in his shade,\\
When in eternal lines to time thou grow'st.\\
\\
So long as men can breathe, or eyes can see,\\
So long lives this, and this gives life to thee.\\
\\
(With a heavy sigh, the lover steps back, their heart laid bare, the words hanging in the air like the fragrance of an unseen bloom, their love declared, though the object of their affection remains unaware, the audience enthralled by the vulnerability of true, unspoken love.)

\hrulefill

\caption{\textbf{AlpacaEval prompt example.}}
\label{fig:alpacaeval}
\end{figure}

\FloatBarrier

\subsection{RAG Prompts}

\begin{figure}[htbp]
\hrulefill

\textbf{Generative input:}

\hrulefill

<s><|user|>\\
what was the name of darth vader star destroyer\\
<|assistant|>\\
The answer is

\hrulefill

\textbf{Generation:}

\hrulefill

~not clear. Darth Vader had a star destroyer, but it was

\hrulefill
\caption{\textbf{No RAG example and generation by \modelbase{}.}}
\label{fig:norag}
\end{figure}

\begin{figure}[htbp]
\hrulefill

\textbf{Generative input:}

\hrulefill

<s><|user|>\\
what was the name of darth vader star destroyer\\\\
Star Destroyer The iconic Star Destroyer first appears in the opening scene of Star Wars, as Darth Vader's flagship, the Devastator, chases the Tantive IV above Tatooine. This appearance shows the Imperial ship's massive size in comparison to the Tantive IV.\\\\
Optionally using the prior context answer the query prior to it\\
<|assistant|>\\
The answer is

\hrulefill

\textbf{Generation:}

\hrulefill

~the Devastator.</s>

\hrulefill
\caption{\textbf{RAG query then document example and generation by \modelbase{}.}}
\label{fig:rag1}
\end{figure}

\begin{figure}[htbp]
\hrulefill

\textbf{Generative input:}

\hrulefill

<s><|user|>\\
Star Destroyer The iconic Star Destroyer first appears in the opening scene of Star Wars, as Darth Vader's flagship, the Devastator, chases the Tantive IV above Tatooine. This appearance shows the Imperial ship's massive size in comparison to the Tantive IV.\\\\
what was the name of darth vader star destroyer\\\\
Answer the prior query while optionally using the context prior to it\\
<|assistant|>\\
The answer is

\hrulefill

\textbf{Generation:}

\hrulefill

~the Devastator.</s>

\hrulefill
\caption{\textbf{RAG document then query example and generation by \modelbase{}.}}
\label{fig:rag2}
\end{figure}

\begin{figure}[htbp]
\hrulefill

\textbf{Embedding input (query) passed via key-value states:}

\hrulefill

<s><|embed|>\\
what was the name of darth vader star destroyer

\hrulefill

\textbf{Generative input:}

\hrulefill

$\backslash$n<|user|>\\
Star Destroyer The iconic Star Destroyer first appears in the opening scene of Star Wars, as Darth Vader's flagship, the Devastator, chases the Tantive IV above Tatooine. This appearance shows the Imperial ship's massive size in comparison to the Tantive IV.\\\\
Optionally using the prior context answer the query prior to it\\
<|assistant|>\\
The answer is

\hrulefill

\textbf{Generation:}

\hrulefill

~Star Destroyer.</s>

\hrulefill

\caption{\textbf{\method{} Query Caching example and generation by \modelbase{}.}}
\label{fig:ragq}
\end{figure}

\begin{figure}[htbp]
\hrulefill

\textbf{Embedding input (doc) passed via key-value states and cached in the index:}

\hrulefill

<s><|embed|>\\
Star Destroyer The iconic Star Destroyer first appears in the opening scene of Star Wars, as Darth Vader's flagship, the Devastator, chases the Tantive IV above Tatooine. This appearance shows the Imperial ship's massive size in comparison to the Tantive IV.

\hrulefill

\textbf{Generative input:}

\hrulefill

$\backslash$n<|user|>\\
what was the name of darth vader star destroyer\\\\
Answer the prior query while optionally using the context prior to it\\
<|assistant|>\\
The answer is

\hrulefill

\textbf{Generation:}

\hrulefill

~Devastator. The iconic Star Destroyer first appears in the opening

\hrulefill

\caption{\textbf{\method{} Doc Caching example and generation by \modelbase{}.}}
\label{fig:ragd}
\end{figure}

\begin{figure}[htbp]
\hrulefill

\textbf{Embedding input (doc) passed via key-value states and cached in the index:}

\hrulefill

<s><|embed|>\\
Star Destroyer The iconic Star Destroyer first appears in the opening scene of Star Wars, as Darth Vader's flagship, the Devastator, chases the Tantive IV above Tatooine. This appearance shows the Imperial ship's massive size in comparison to the Tantive IV.

\hrulefill

\textbf{Embedding input (query) passed via key-value states:}

\hrulefill

<s><|embed|>\\
what was the name of darth vader star destroyer

\hrulefill

\textbf{Generative input:}

\hrulefill

$\backslash$n<|user|>\\
Answer the prior query while optionally using the context prior to it\\
<|assistant|>\\
The answer is

\hrulefill

\textbf{Generation:}

\hrulefill

~the Star Destroyer. The Star Destroyer is a massive spacecraft

\hrulefill

\caption{\textbf{\method{} Doc-Query Caching example and generation by \modelbase{}.} Unlike for Doc Caching, we prepend the bos token (``<s>'') to both query and document, which improved the match score from 14.13 to 18.39.}
\label{fig:ragdq}
\end{figure}

\FloatBarrier

\begin{figure}[htbp]
\hrulefill

\textbf{Embedding input (query) passed via key-value states:}

\hrulefill

<s><|embed|>\\
what was the name of darth vader star destroyer

\hrulefill

\textbf{Embedding input (doc) passed via key-value states and cached in the index:}

\hrulefill

<|embed|>\\
Star Destroyer The iconic Star Destroyer first appears in the opening scene of Star Wars, as Darth Vader's flagship, the Devastator, chases the Tantive IV above Tatooine. This appearance shows the Imperial ship's massive size in comparison to the Tantive IV.

\hrulefill

\textbf{Generative Input:}

\hrulefill

$\backslash$n<|user|>\\
Optionally using the prior context answer the query prior to it\\
<|assistant|>\\
The answer is

\hrulefill

\textbf{Generation:}

\hrulefill

~the Star Destroyer.

\hrulefill

\caption{\textbf{\method{} Query-Doc Caching example and generation by \modelbase{}.}}
\label{fig:ragqd}
\end{figure}

\section{Hardware}
\label{sec:hardware}


For the training of \modelbase{}, we used 8 nodes with 8 NVIDIA A100 80GB GPUs each for 48 hours corresponding to 3,072 GPU hours. Meanwhile for \modelbig{}, we used 32 nodes with 8 NVIDIA H100 80GB GPUs each for 80 hours corresponding to 20,480 GPU hours. As we train both models for 1253 steps, this corresponds to several minutes per step. This slow training time is mainly due to (a) a large batch size per step, (b) large models and our associated strategies to make them fit into memory at the cost of speed (\autoref{sec:embmem}, \autoref{sec:hps}), and (c) a cluster with slow inter-node communication. The Gen.-only and Emb.-only models in \autoref{tab:emb} used 72 and 1760 H100 80GB GPU hours, respectively. Adding up all ablations and evaluations, we likely used somewhere around 100,000 GPU hours.

\section{Limitations and Future Work}
\label{sec:limits}

\paragraph{\model{} Agents} Future work may consider using the embedding capability to let the generative model initiate a search over an index when it deems necessary. Currently, this is often accomplished via external retrieval plugins. Such plugins are no longer necessary if the model can retrieve on its own. Teaching the model to invoke its own embedding capability likely requires additional finetuning (just like teaching it to invoke an external plugin~\cite{schick2023toolformer}). A sample could look something like:\\``\verb@<|user|>\nWhat is the capital of Japan?\n<|internal|>\nI am not@ \verb@sure I know this. Let me produce an embedding for it and search@ \verb@for the answer. Retrieve answers for this query.\n<|embed|>\nWhat@ \verb@is the capital of Japan?\n<|output|>\nTokyo, Japan’s busy capital,@ \verb@mixes the ultramodern and the traditional..\n<|assistant|>\n@ \verb@The capital of Japan is Tokyo.\n</s>@''

\paragraph{Pretraining} For our experiments we take an off-the-shelf pretrained language model. However, it should also be possible to use the \method{} approach to pretrain from scratch. As labeled embedding data is likely too scarce for pretraining, one could either rely on unsupervised approaches for the embedding objective, such as RetroMAE~\cite{xiao2022retromae,xiao2022retromae2}, or use methods like data augmentation~\cite{dhole2022nlaugmenter}, pruning~\cite{xia2023sheared} or multi-epoch training to deal with the data constraint~\cite{muennighoff2023scaling,luukkonen2023fingpt}.

\paragraph{Format Efficiency} Our format in \autoref{fig:format} is inefficient, as encoding the embedding format, \verb@<s><|user|>\n<|embed|>\n@,  requires 13 tokens and encoding the generative format, \verb@<s><|user|>\n<|assistant|>\n</s>@, requires 15 tokens. Using special tokens could simplify this and thus make training and inference slightly cheaper.

\paragraph{Training efficiency: Packing and Reusing} It is common to pack samples during generative instruction tuning to maximize efficiency~\citep{chung2022scaling,muennighoff2023crosslingual}. Packing embedding samples during training should also be possible by ensuring attention is only paid to each respective sample. Going even further is it possible to pack generative and embedding training data into the same sample and reuse the same sample for both tasks? This could look similar to the example provided in ``\model{} Agents'' with the generative loss applied over the assistant response and the contrastive loss applied to the representation of the text following ``\verb@<|embed|>@''. By reusing samples it may be possible to significantly decrease the resources needed for \method{}.

\section{Version Control}
\label{sec:vc}

\textbf{V2 → V3 (2025-03):}
\begin{itemize}
\item Fixed a figure reference in \autoref{sec:rag}
\item Elaborated more on training discussion and potential avenues for improvement in \autoref{sec:limits}
\item Added \autoref{sec:addcaching}
\item Rephrased the end of \autoref{sec:intro} to better motivate that GritLM requires less compute than separate generative and embedding models when considering pretraining
\item Added more discussion on the potential speed-performance trade-off of using a smaller and faster embedding model by using the embedding from intermediate layers of GritLM in \autoref{sec:ablations} after our embedding head ablation
\item Added resources used by Gen.-only and Emb.-only baselines in \autoref{sec:hardware}
\end{itemize}

\textbf{V1 → V2 (2024-04):}
\begin{itemize}
\item Added KTO experiments in \autoref{sec:main}
\item Fixed sample in \autoref{fig:tulu2}
\end{itemize}

\end{document}